\pdfoutput=1

\documentclass[final]{cvpr}

\usepackage{times}
\usepackage{epsfig}
\usepackage{graphicx}
\usepackage{amsmath}
\usepackage{amssymb}
\usepackage{hyperref}
\usepackage{url}
\usepackage[ruled,vlined]{algorithm2e}
\usepackage{graphicx}
\usepackage{comment}
\usepackage{pbox}
\usepackage{rotating}
\usepackage{array,multirow,graphicx}
\usepackage{float}
\usepackage[normalem]{ulem}

\usepackage[utf8]{inputenc}
\usepackage[english]{babel}

\usepackage[dvipsnames,table,xcdraw]{xcolor}

\definecolor{blue}{rgb}{0,0,1}
\definecolor{red}{rgb}{1,0,0}
\definecolor{green}{rgb}{0,.5,0}
\definecolor{mygreen}{rgb}{0,.5,0.1}
\definecolor{orange}{rgb}{0.75, 0.4, 0}

\newcommand{\wu}[1]{{\color{blue}\textbf{}#1}}

\newcommand{\apsec}[1]{Appendix~\ref{#1}}
\newcommand{\apfig}[1]{app.~Figure~\ref{#1}}
\newcommand{\aptab}[1]{app.~Table~\ref{#1}}

\def\cvprPaperID{5532} 
\def\confYear{CVPR 2021}

\begin{document}

\title{StyleSpace Analysis: Disentangled Controls for StyleGAN Image Generation}

\author{Zongze Wu\\
Hebrew University\\
{\tt\small zongze.wu@mail.huji.ac.il}
\and
Dani Lischinski\\
Hebrew University\\
{\tt\small danix@cs.huji.ac.il}
\and
Eli Shechtman\\
Adobe Research\\
{\tt\small elishe@adobe.com}
}

\maketitle

\begin{abstract}

We explore and analyze the latent style space of StyleGAN2, a state-of-the-art architecture for image generation, using models pretrained on several different datasets.
We first show that StyleSpace, the space of channel-wise style parameters, is significantly more disentangled than the other intermediate latent spaces explored by previous works.
Next, we describe a method for discovering a large collection of style channels, each of which is shown to control a distinct visual attribute in a highly localized and disentangled manner.
Third, we propose a simple method for identifying style channels that control a specific attribute, using a pretrained classifier or a small number of example images. 
Manipulation of visual attributes via these StyleSpace controls is shown to be better disentangled than via those proposed in previous works. To show this, we make use of a newly proposed Attribute Dependency metric.
Finally, we demonstrate the applicability of StyleSpace controls to the manipulation of real images.  
Our findings pave the way to semantically meaningful and well-disentangled image manipulations via simple and intuitive interfaces.

\end{abstract}
\section{Introduction}

\newcommand{\und}{\rule{1ex}{.4pt}}
\begin{figure*}
	\setlength{\tabcolsep}{1pt}
	\begin{center}
	\begin{tabular}{cccccccc}
		& {\small Amount of hair (6\und{}364)} & & & {\small Pillow presence (8\und{}119)} & & & {\small Hubcap style (12\und{}113)} \\
    	\rotatebox{90}{\small \hspace{-1.5cm} Hair greyness (11\und{}286)} &
    	\renewcommand{\arraystretch}{0.5}
    	\begin{tabular}{ccc}
		\includegraphics[width=15mm]{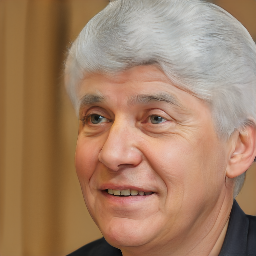} &
		\includegraphics[width=15mm]{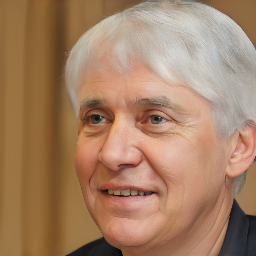} &
		\includegraphics[width=15mm]{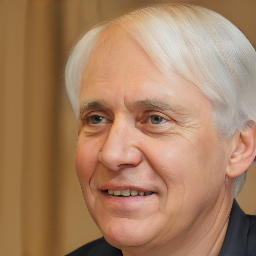} \\
		
		\includegraphics[width=15mm]{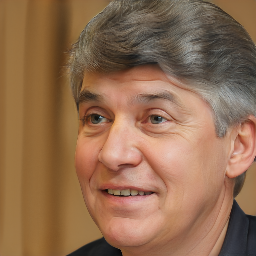} &
		\includegraphics[width=15mm]{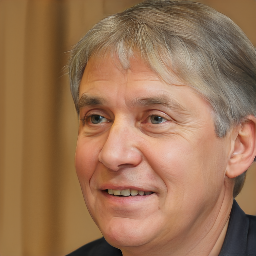} &
		\includegraphics[width=15mm]{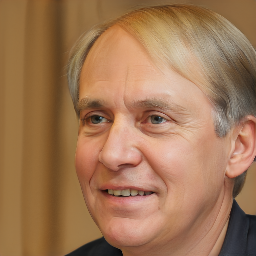} \\
		
		\includegraphics[width=15mm]{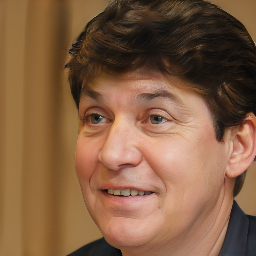} &
		\includegraphics[width=15mm]{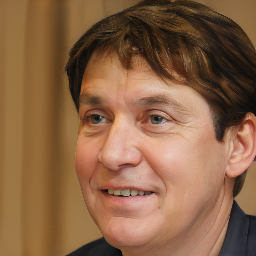} &
		\includegraphics[width=15mm]{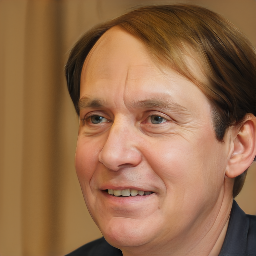} \\
	\end{tabular} & \hspace{2mm} & \rotatebox{90}{\small \hspace{-1.20cm} Cover style (6\und{}420)} &
    	\renewcommand{\arraystretch}{0.5}
		\begin{tabular}{ccc}
		\includegraphics[width=15mm]{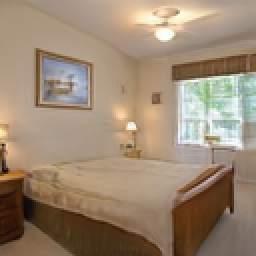} &
		\includegraphics[width=15mm]{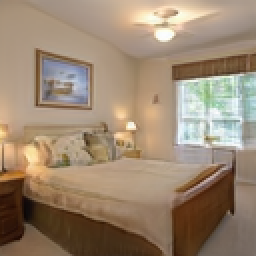} &
		\includegraphics[width=15mm]{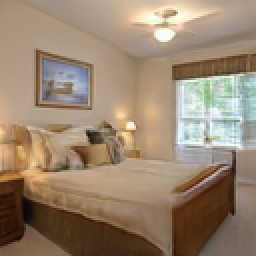} \\
		
		\includegraphics[width=15mm]{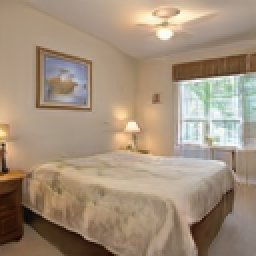} &
		\includegraphics[width=15mm]{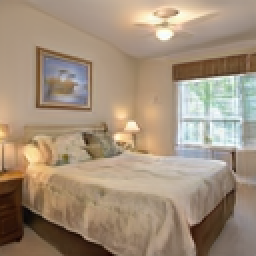} &
		\includegraphics[width=15mm]{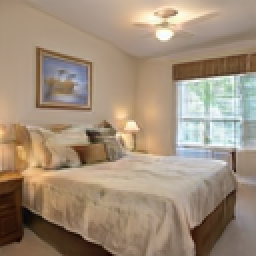} \\
		
		\includegraphics[width=15mm]{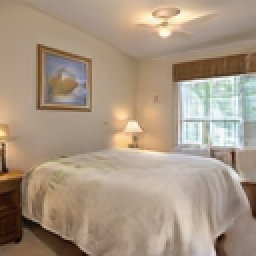} &
		\includegraphics[width=15mm]{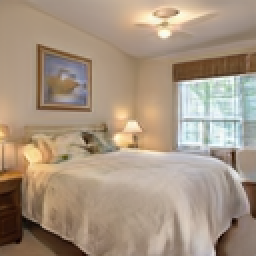} &
		\includegraphics[width=15mm]{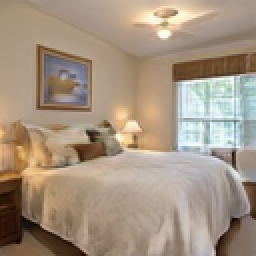} \\
	\end{tabular} & \hspace{2mm} & \rotatebox{90}{\small \hspace{-1.18cm} Car color (12\und{}142)} &
    	\renewcommand{\arraystretch}{0.5}
		\begin{tabular}{ccc}
		\includegraphics[width=15mm]{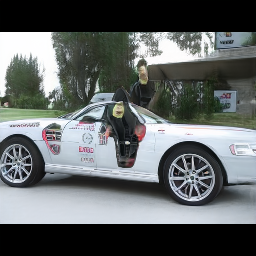} &
		\includegraphics[width=15mm]{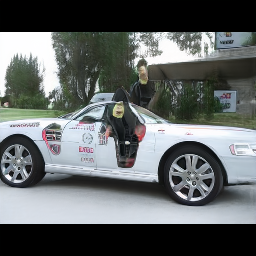} &
		\includegraphics[width=15mm]{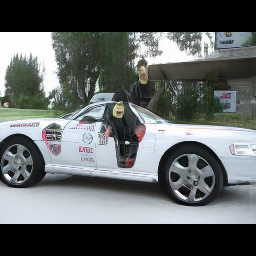} \\
		
		\includegraphics[width=15mm]{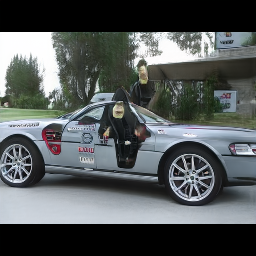} &
		\includegraphics[width=15mm]{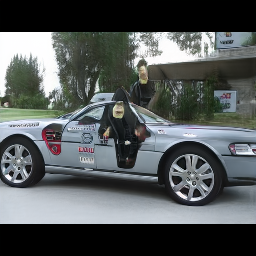} &
		\includegraphics[width=15mm]{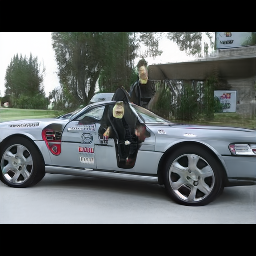} \\
		
		\includegraphics[width=15mm]{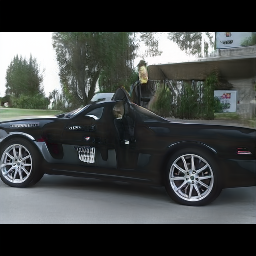} &
		\includegraphics[width=15mm]{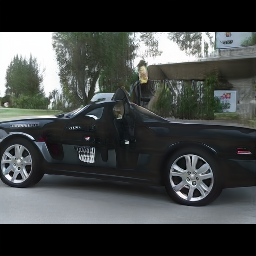} &
		\includegraphics[width=15mm]{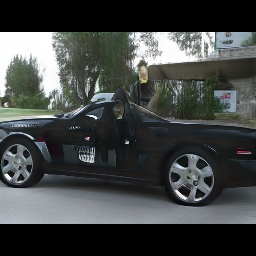} \\
		\end{tabular}
	\end{tabular}
	\end{center}
	\caption{Disentanglement in style space, demonstrated using three different datasets. Each of the three groups above shows two manipulations that occur independently inside the same semantic region (hair, bed, and car, from left to right). The indices of the manipulated layer and channel are indicated in parentheses.
}
	\label{fig:disentanglement}
\end{figure*}

Modern Generative Adversarial Networks (GANs) are able to produce a wide variety of highly realistic synthetic images.
The phenomenal success of these generative models underscores the need for a better
understanding of ``what makes them tick'' and what kinds of control these models offer over the generated data.
Of particular practical importance are controls that are interpretable and disentangled, as they suggest intuitive image manipulation interfaces.

In traditional GAN architectures, such as DCGAN~\cite{radford2015unsupervised} and Progressive GAN~\cite{karras2017progressive}, the generator starts with a random latent vector, drawn from a simple distribution, and transforms it into a realistic image via a sequence of convolutional layers. 
Recently, style-based designs have become increasingly popular, where the random latent vector is first transformed into an intermediate latent code via a mapping function. 
This code is then used to modify the channel-wise activation statistics at each of the generator's convolution layers. BigGAN~\cite{brock2018large} uses class-conditional BatchNorm \cite{ioffe2015batch}, while StyleGAN~\cite{karras2019style} uses AdaIN~\cite{huang2017arbitrary} to modulate channel-wise means and variances. StyleGAN2~\cite{karras2020analyzing} controls channel-wise variances by modulating the weights of the convolution kernels.
It has been shown that the intermediate latent space is more disentangled than the initial one \cite{karras2019style}. Additionally, Shen \etal~\cite{shen2020interpreting} show that the latent space of StyleGAN~\cite{karras2019style,karras2020analyzing} is more disentangled than that of Progressive GAN~\cite{karras2017progressive}.

Some control over the generated results may be obtained via conditioning~\cite{mirza2014conditional}, which requires training the model with annotated data. In contrast, style-based design enables discovering a variety of interpretable generator controls after training the generator. However, current methods require either a pretrained classifier~\cite{goetschalckx2019ganalyze,shen2020interpreting,shen2020interfacegan,yang2019semantic}, a large set of paired examples~\cite{jahanian2019steerability}, or manual examination of many candidate control directions~\cite{harkonen2020ganspace}, which limits the versatility of these approaches. Furthermore, the individual controls discovered by these methods are typically entangled, affecting multiple attributes, and are often non-local.

In this work, our goal is to understand to what degree disentanglement is inherent in style-based generator architectures. Perhaps an even more important question is to how to find these disentangled controls? In particular, can this be done in an unsupervised manner, or with only a small amount of supervision? In this paper we report several findings with respect to these questions.

Recent studies of disentangled representations~\cite{eastwood2018framework,ridgeway2018learning} consider a latent representation to be perfectly disentangled if each latent dimension controls a single visual attribute (\emph{disentanglement}), and each attribute is controlled by a single dimension (\emph{completeness}).
Following this terminology, we explore the latent space of StyleGAN2~\cite{karras2020analyzing}. Unlike other works that analyze the (intermediate) latent space $\mathcal{W}$ or $\mathcal{W+}$~\cite{abdal2019image2stylegan}, we examine \emph{StyleSpace}, the space spanned by the channel-wise style parameters, denoted $\mathcal{S}$. In Section~\ref{sec:disent_level} we  measure and compare the disentanglement and completeness of these spaces using the metrics proposed for this purpose~\cite{eastwood2018framework}. To our knowledge we are the first to apply this quantitative framework to models trained on real data. Our experiments reveal that $\mathcal{S}$ is significantly better disentangled than $\mathcal{W}$ or $\mathcal{W}+$. 

In Section~\ref{sec:localized} we propose a simple method for detecting StyleSpace channels that control the appearance of local semantic regions in the image. By computing the gradient maps of generated images with respect to different style parameters, we identify those channels that are consistently active in specific semantic regions, such as hair or mouth, in the case of portraits. We demonstrate the effectiveness of this approach across three different datasets (FFHQ \cite{karras2019style}, LSUN Bedroom, and LSUN Car \cite{yu2015lsun}).
The StyleSpace channels that we detect are highly localized, affecting only a specific area without any visible impact of other regions. They are also surprisingly well disentangled from each other, as demonstrated in Figure~\ref{fig:disentanglement}.

Our next goal is to identify style channels that control a specific target attribute. To achieve this goal we require a set of exemplar images that exhibit the attribute of interest. The basic idea is to compare the average style vector across the exemplar set to the population average, thereby detecting dimensions that deviate the most. Our experiments indicate that such dimensions usually indeed control the target attribute, and reveal that a single attribute is typically controlled by only a few different StyleSpace channels.

To our knowledge, there is no metric to compare the disentanglement of different image manipulation controls. In Section~\ref{sec:ADcomparison} we propose Attribute Dependency (AD) as a measure for how manipulating a target attribute affects other attributes.
Comparing manipulations performed in StyleSpace to those in $\mathcal{W}$ and $\mathcal{W+}$ spaces \cite{harkonen2020ganspace,shen2020interfacegan}, shows that our controls exhibit significantly lower AD. 

Finally, we share our insights about the pros and cons of two major image inversion methods, latent optimization \cite{karras2020analyzing,abdal2019image2stylegan,abdal2020image2stylegan++} and encoders~\cite{zhu2020domain}. We show that a combination of the two may be used in order to apply our StyleSpace controls to disentangled manipulation of real images.

\section{Related Work}




Understanding the latent representations of pretrained generators has attracted considerable attention, since it contributes to better GAN architecture design and facilitates controllable manipulation. Bau \etal~\cite{bau2019gandissect,bau2019semantic} utilized semantic segmentation to analyze Progressive GAN~\cite{karras2017progressive} and detect causal units that control the presence of certain objects through ablation.
Shen \etal~\cite{shen2020interfacegan} and Yang \etal~\cite{yang2019semantic} 
utilize classifiers to analyze StyleGAN \cite{karras2019style} and show that a linear manipulation in $\mathcal{W}$ space can control a specific target attribute. They further show that in $\mathcal{W+}$ space, early layers control layout, middle layers control the presence of objects, and late layers control final rendering.
Collins \etal~\cite{collins2020editing} transfer the appearance of a specific object part
from a reference image to a target image, through swapping between style codes.
Concurrent work by Xu \etal~\cite{xu2020generative} shows that style space can be used for a variety of discriminative and generative tasks.

By utilizing the weights of pretrained generators, several works \cite{abdal2019image2stylegan,abdal2020image2stylegan++,gabbay2019style,gu2020image,pan2020exploiting} design different latent optimization methods to do inpainting, style transfer, morphing, colorization, denoising and super resolution. Instead of latent optimization, Nitzan \etal~\cite{nitzan2020disentangling} use the generator as a fixed decoder, and facilitate disentanglement by training an encoder for identity and another encoder for pose. Richardson \etal~\cite{richardson2020encoding} do image translation by training encoders from sketches or semantic maps into StyleGAN's $\mathcal{W}$ space.

To facilitate attribute manipulations in an unsupervised manner, H{\"a}rk{\"o}nen \etal~\cite{harkonen2020ganspace} detect interpretable controls based on PCA applied either to the latent space of StyleGAN \cite{karras2019style} or to the feature space of BigGAN~\cite{brock2018large}. Layerwise perturbations along the principle directions give rise to a variety of useful controls.
Similarly, Shen \etal~\cite{shen2020closed} do eigenvector decomposition in the affine transformation layer between $\mathcal{W}$ and $\mathcal{S}$ spaces, and use eigenvectors with the highest eigenvalues as manipulation directions. 
Peebles \etal~\cite{peebles2020hessian} identify interpretable controls by minimizing a Hessian loss. However, in unsupervised settings, users must examine many different manipulation directions and manually
identify meaningful controls.

In contrast, we discover a large amount of localized controls using semantic maps (Section~\ref{sec:localized}). The controls are ranked making it easier to detect meaningful localized manipulations in each semantic region. Furthermore, our controls are surprisingly well disentangled and fine-grained. We also detect attribute-specific controls using a small number of examples (Section~\ref{sec:attributes}).

\section{Disentanglement of StyleGAN latent spaces}
\label{sec:disent_level}

The StyleGAN/StyleGAN2 generation process involves a number of latent spaces. The first latent space, $\mathcal{Z}$, is typically normally distributed. Random noise vectors $z \in \mathcal{Z}$ are transformed into an \emph{intermediate} latent space $\mathcal{W}$ via a sequence of fully connected layers. The $\mathcal{W}$ space is claimed to better reflect the disentangled nature of the learned distribution~\cite{karras2019style}. Each $w \in \mathcal{W}$ is further transformed to channel-wise style parameters $s$, using a different learned affine transformation for each layer of the generator. We refer to the space spanned by these style parameters as \emph{StyleSpace}, or $\mathcal{S}$.
Some works make use of another latent space, $\mathcal{W+}$, 
where a different intermediate latent vector $w$ is fed to each of the generator's layers. $\mathcal{W+}$ is mainly used for style mixing~\cite{karras2019style} and for image inversion~\cite{abdal2019image2stylegan,karras2020analyzing,zhu2020domain}.


In StyleGAN2~\cite{karras2020analyzing}, there is a single style parameter per channel, which controls the feature map variances by modulating the convolution kernel weights. Additional style parameters are used by the tRGB blocks that transform feature maps to RGB images at each resolution \cite{karras2020analyzing}.
Thus, in a 1024 $\times$ 1024 StyleGAN2 with 18 layers, $\mathcal{W}$ has 512 dimensions, $\mathcal{W+}$ has 9216 dimensions, and $\mathcal{S}$ has 9088 dimensions in total, consisting of 6048 dimensions applied to feature maps, and 3040 additional dimensions for tRGB blocks. See \apsec{supp-sec:structureS} for more detail. Below we refer to individual \emph{dimensions} of $\mathcal{S}$ as \emph{StyleSpace channels}. 



Our first goal is to determine which of these latent spaces offers the most disentangled representation. To this end, we use the recently proposed  DCI  (disentanglement / completeness / informativeness) metrics~\cite{eastwood2018framework}, which are suitable for comparing latent representations with different dimensions. The DCI metrics employ regressors trained using a set of latent vectors paired with corresponding attribute vectors (split into training and testing sets). \emph{Disentanglement} measures the degree to which each latent dimension captures at most one attribute, \emph{completeness} measures the degree to which each attribute is controlled by at most one latent dimension, while \emph{informativeness} measures the classification accuracy of the attributes, given the latent representation.

Rather than analyzing the degree of disentanglement using a synthetically generated dataset, where the factors of variations are few and known~\cite{eastwood2018framework}, we analyze StyleGAN2 trained on a real dataset, specifically FFHQ. To generate the training data for the DCI regressors, we employ 40 binary classifiers pretrained on the CelebA attributes \cite{karras2019style}. The classifiers are trained to detect common features in portraits such as gray hair, smiling, and lipstick, and their logit outcome is converted to a binary one via a sigmoid activation.


We first randomly sample 500K latent vectors $z \in \mathcal{Z}$ and record their corresponding $w$ and $s$ vectors, as well as the generated images. Each image is then annotated by each of the 40 classifiers, where we record the logit, rather than just the binary outcome. Since not all attributes are well represented in the generated images (for example, there are very few portraits with a necktie), we only consider 31 attributes for which there are more than 5\% positive and 5\% negative outcomes. Similarly to Shen \etal~\cite{shen2020interfacegan}, we reduce classifer uncertainty by using only the most positive 2\% and most negative 2\% examples, for each attribute, and split the examples equally into training and testing sets. 

\setlength{\tabcolsep}{3pt}
\begin{table}[bt]
\scriptsize
\begin{center}
\begin{tabular}{lccc} 
	& \multicolumn{3}{c}{\textbf{Comparison w/ $\mathcal{Z}$ and $\mathcal{W}$}} \\
  & \textbf{Disent.} & \textbf{Compl.}  &\textbf{Inform.}\\ 
      \hline \\
$\mathcal{Z}$ & 0.31 & 0.21 & 0.72 \\ 
$\mathcal{W}$ & 0.54 & 0.57 & 0.97\\ 
$\mathcal{S}$ & \textbf{0.75} & \textbf{0.87} & \textbf{0.99}\\ 
\end{tabular}
\hspace{0.15cm}
\vline
\hspace{0.15cm}
\begin{tabular}{lccc} 
	& \multicolumn{3}{c}{\textbf{Comparison with $\mathcal{W+}$}} \\
  & \textbf{Disent.} & \textbf{Compl.}  &\textbf{Inform.}\\ 
      \hline \\
      & & \\
$\mathcal{W+}$ & 0.54 & 0.64 & 0.94 \\ 
$\mathcal{S}$ & \textbf{0.63} & \textbf{0.81} &\textbf{0.98} \\ 
\end{tabular}
\end{center}
\caption{\label{tab:disent} Disentanglement, completeness and informativeness for different latent spaces (larger is better, maximum is 1). The two comparisons are performed using different sets of images; thus, the scores are not comparable between the two tables.} 
\end{table}

Finally, we compute the DCI metrics \cite{eastwood2018framework} to compare the latent spaces $\mathcal{Z}$, $\mathcal{W}$ and $\mathcal{S}$.  
As shown in Table~\ref{tab:disent} (left), while the informativeness of both $\mathcal{W}$ and $\mathcal{S}$ is high and comparable, $\mathcal{S}$ scores much higher in terms of disentanglement and completeness. This indicates that each dimension of $\mathcal{S}$ is more likely to control a single attribute and vice versa.

Since $\mathcal{W+}$ is often used for StyleGAN inversion~\cite{abdal2019image2stylegan,karras2020analyzing}, we also perform a separate experiment to compare between $\mathcal{W+}$ and $\mathcal{S}$. Specifically, we first randomly sample 500K intermediate latent codes $w \in \mathcal{W}$, and construct each $w+$ by concatenating $n_l$ random $w$ codes ($n_l = 18$ for a $1024 \times 1024$ StyleGAN2). The resulting images are somewhat less natural than those obtained in the standard manner, resulting in a smaller number of considered attributes (25 instead of 31), which we use to evaluate $\mathcal{W+}$ and $\mathcal{S}$ as before. Table~\ref{tab:disent} (right) shows, again, that $\mathcal{S}$ scores higher than $\mathcal{W+}$. 




To our knowledge, we are the first to perform a quantitative evaluation of latent space disentanglement for a GAN model trained on real data. Since our analysis indicates that the style space $\mathcal{S}$ is more disentangled than the other latent spaces of StyleGAN2, we proceed to further analyze $\mathcal{S}$ in the remainder of this paper.

\section{Detecting locally-active style channels}
\label{sec:localized}

\begin{figure}[tb]
\begin{center}
\setlength{\tabcolsep}{1pt}

\begin{tabular}{c}
\scriptsize{Grad.~map} \vspace{-1.5mm} \\
\scriptsize $G_{(11\rule{1ex}{.4pt}286)}$ \\
\includegraphics[scale=0.12]{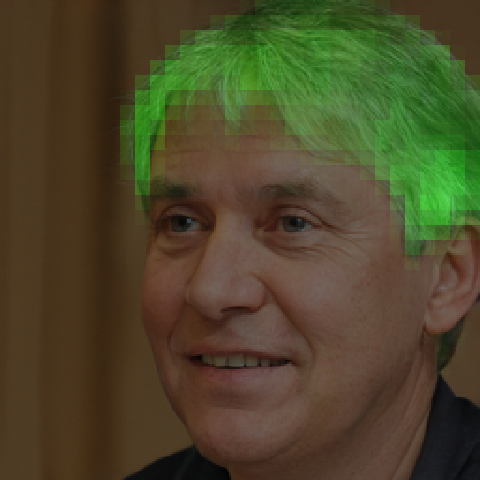}\\
\end{tabular}
\begin{tabular}{ccc}
 & {\scriptsize $G \!>\! t_{u,c}$} & {\scriptsize $M_c$} \\
 \rotatebox{90}{\scriptsize \phantom{kk}hair} &
\includegraphics[scale=0.08]{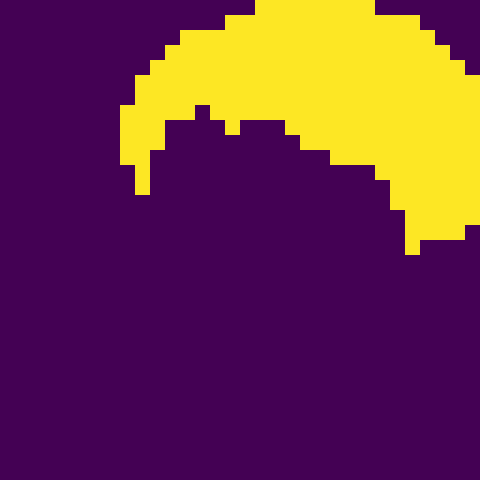} &
\includegraphics[scale=0.08]{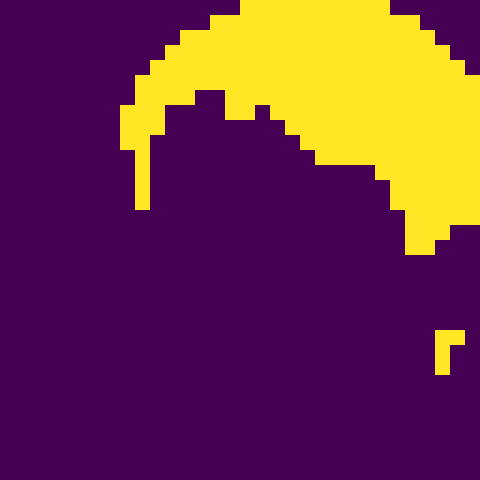} \\
\rotatebox{90}{\scriptsize \phantom{k}mouth} &
\includegraphics[scale=0.08]{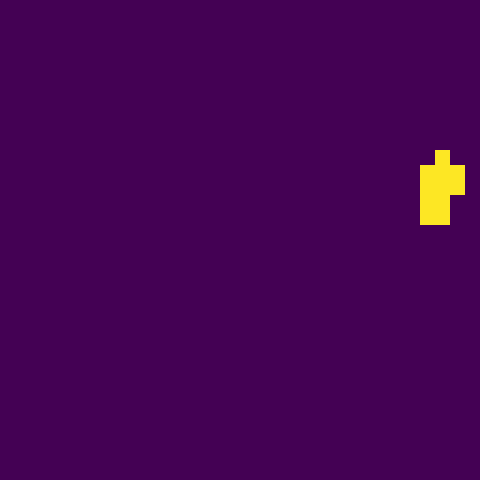} &
\includegraphics[scale=0.08]{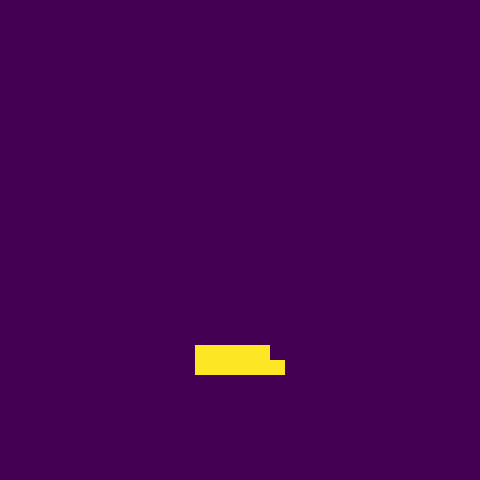} \\
\end{tabular}
\vline
\begin{tabular}{c}
\scriptsize{Grad.~map} \vspace{-1.5mm} \\
\scriptsize $G_{(6\rule{1ex}{.4pt}202)}$ \\
\includegraphics[scale=0.12]{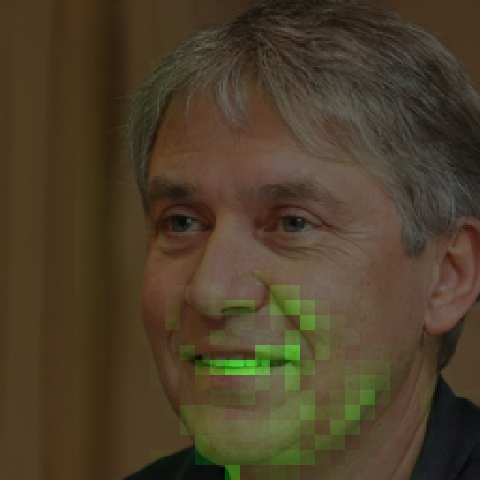}\\
\end{tabular}
\begin{tabular}{ccc}
 & {\scriptsize $G \!>\! t_{u,c}$} & {\scriptsize $M_c$} \\
\rotatebox{90}{\scriptsize \phantom{kk}hair} &
\includegraphics[scale=0.08]{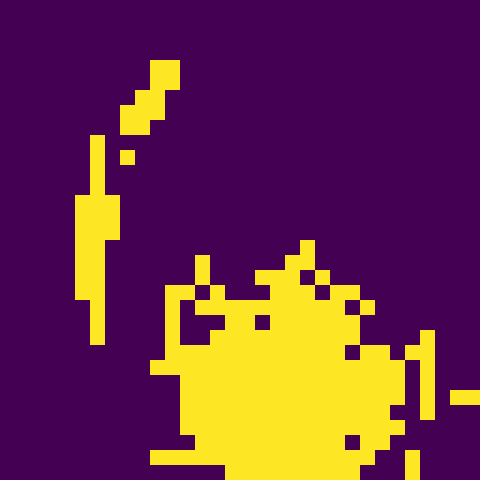} &
\includegraphics[scale=0.08]{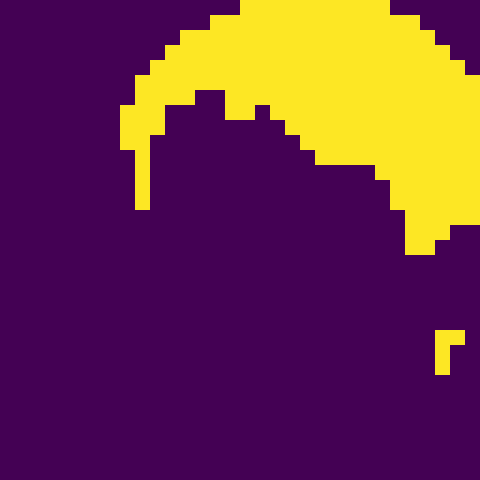} \\
\rotatebox{90}{\scriptsize \phantom{k}mouth} &
\includegraphics[scale=0.08]{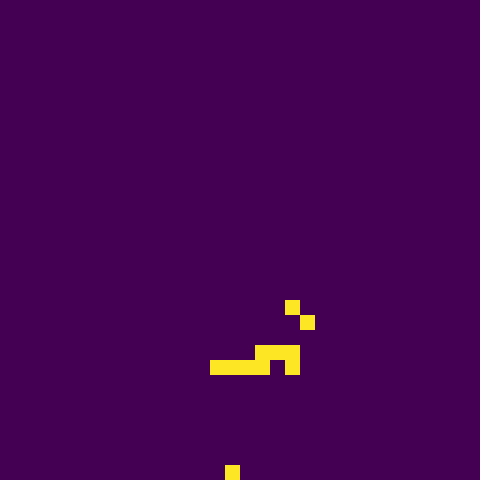} &
\includegraphics[scale=0.08]{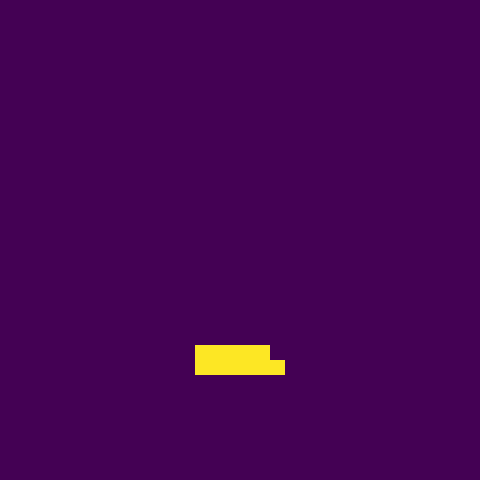} \\
\end{tabular}
\end{center}
   \caption{
   	 A gradient map with respect to each style channel $u$, e.g., (11\rule{1ex}{.4pt}286), channel 286 of generator level 11, is thresholded against a category-specific threshold, chosen such that the resulting mask has the same size as the semantic mask $M_{c}$. The gradient mask of (11\rule{1ex}{.4pt}286) has large overlap with the mask for hair, and no overlap with the mouth, while that of (6\rule{1ex}{.4pt}202) has large overlap with the mask for mouth and almost none with the hair.}
\label{fig:gradient}
\end{figure}




In this section we describe a simple method for detecting StyleSpace channels that control the visual appearance of local semantic regions.
The intuition behind our approach is that by examining the gradient maps of generated images with respect to different channels, and measuring their overlap with specific semantic regions, we can identify those channels that are consistently active in each region. This is demonstrated in Figure~\ref{fig:gradient} using two
gradient maps for two different channels. If the overlap is consistent over a large number of images, these channels will be identified as locally-active for the overlapped semantic regions.


Specifically, for each image generated with style code $s \in \mathcal{S}$, we apply back-propagation to compute the gradient map of the image with respect to each channel of $s$. To save computation, the gradient maps are computed at a reduced spatial resolution $r \times r$ ($r=32$ in our experiments).
Next, a pretrained image segmentation network is used to obtain the semantic map $M^s$ of the generated image. The map is resized to $r \times r$ by using the most abundant semantic category inside each bin as its
semantic label. For each semantic category $c$ and each channel $u$, we measure the overlap between the semantic region $M^s_c$ and the gradient map $G^s_u$:  
\begin{equation}\label{OC}
OC^s_{u,c}=\frac{|(G^s_{u} > t^s_{u,c}) \cap M^s_{c}|}{|M^s_{c}|^{d}}.
\end{equation}
Here $t^s_{u,c}$ is a threshold chosen such that gradient mask $(G^s_{u} > t^s_{u,c})$ has the same size as $M^s_{c}$ (see Figure \ref{fig:gradient}). The correction factor $d$ gives more weight to small areas, since a large overlap between two small masks indicates precise localization. In practice, $d=2$ gives us good balance between large and small areas.

\begin{figure*}
	\textbf{Hair:}\\
	\includegraphics[scale=0.23]{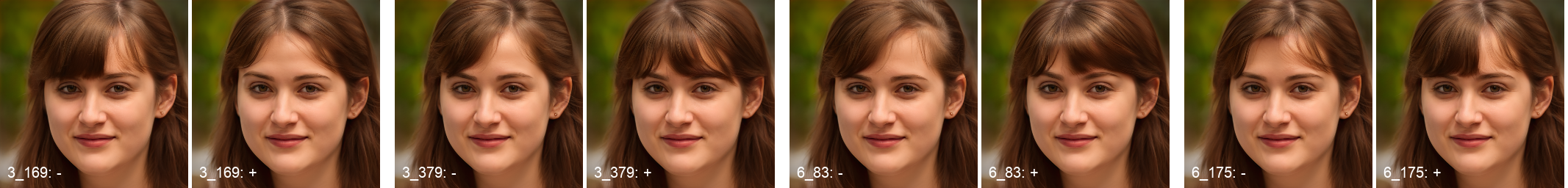}
	\includegraphics[scale=0.23]{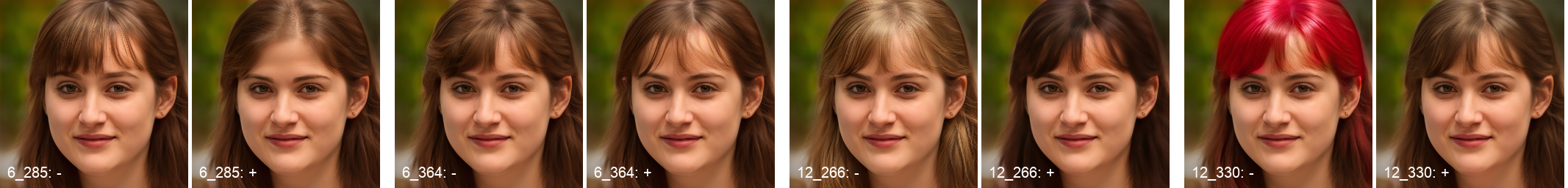}\\
	\textbf{Mouth:}\\
	\includegraphics[scale=0.23]{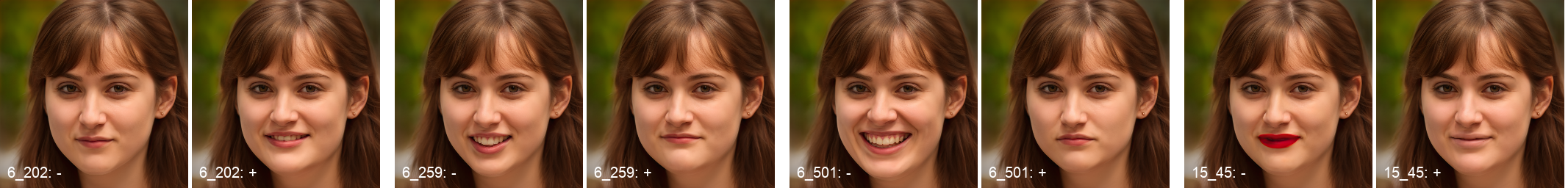}\\
	\textbf{Eyes:}\\
	\includegraphics[scale=0.23]{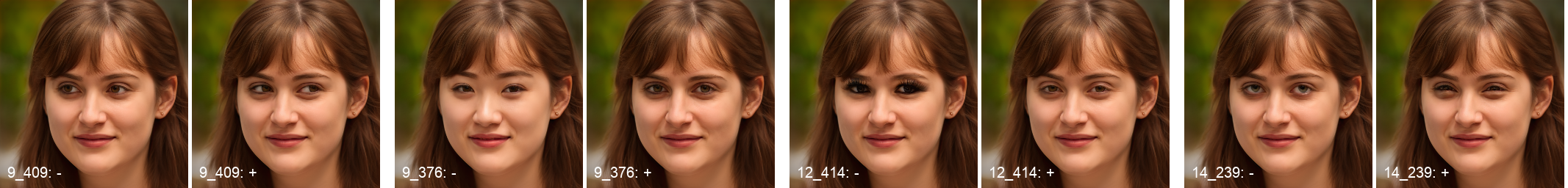}\\
	\textbf{Eyebrows:}\\
	\includegraphics[scale=0.23]{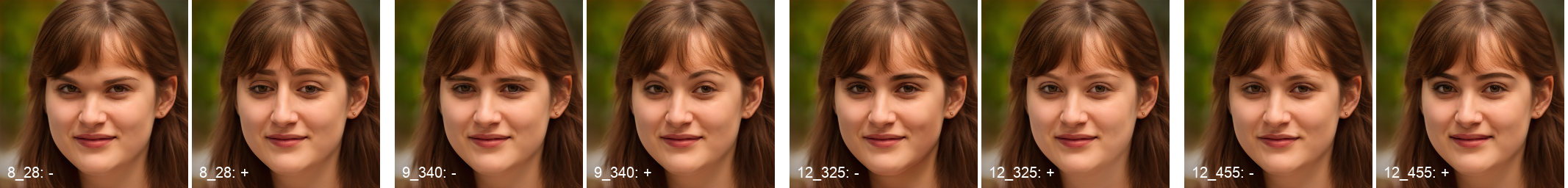}\\
	\textbf{Ears:}\\
	\includegraphics[scale=0.23]{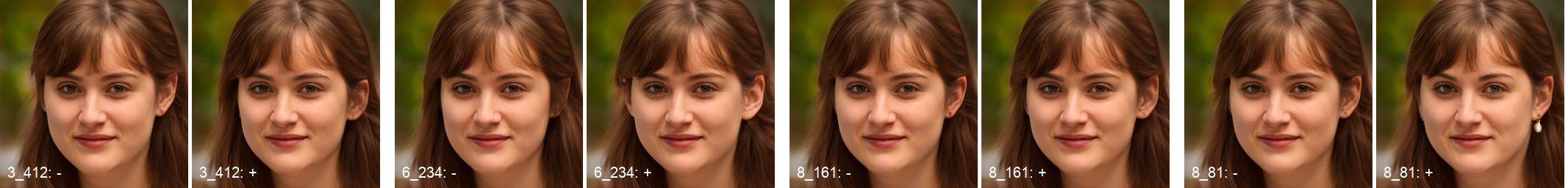}
	
	\caption{Examples of manipulations, each controlled by a single style channel. Each pair of images shows the result of manipulation by decreasing (-) and increasing (+) the value of the style parameter (the original image is omitted). The layer index, channel index, and the direction of change is overlayed in the bottom left corner.
	}
	\label{fig:localized_dimenion}
\end{figure*}

\begin{figure*}

	\textbf{Cars:}\\
	\includegraphics[width=\textwidth]{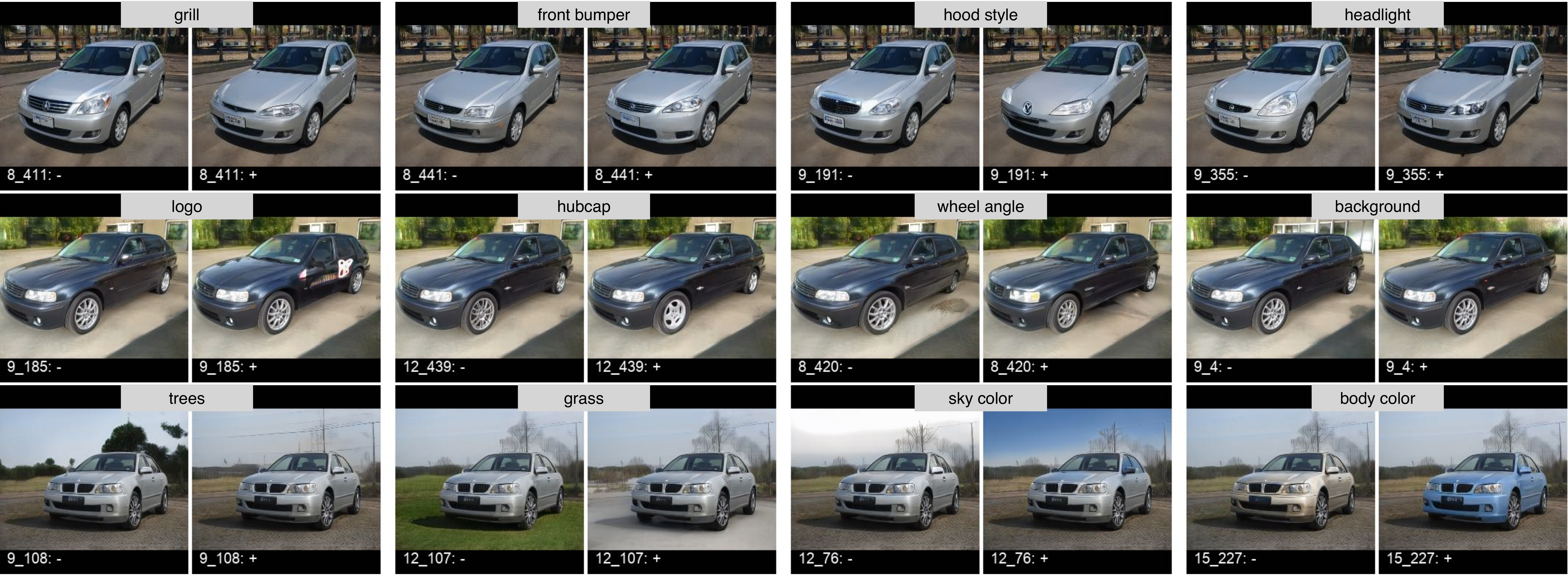}
	\textbf{Bedrooms:}\\
	\includegraphics[width=\textwidth]{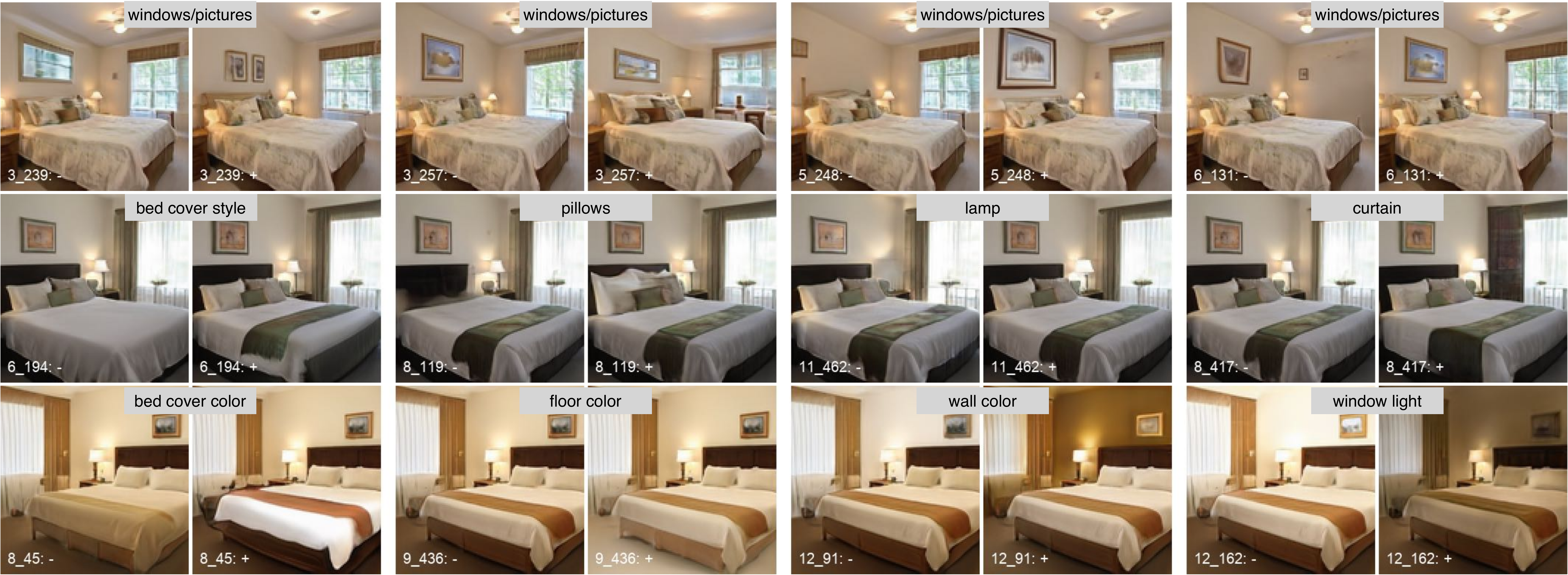}
	\vspace{-4mm}	
	\caption{\label{fig:localized_dimenion2} 
	Examples of manipulations, each controlled by a single style dimension. Each pair of images shows the result of manipulation by decreasing (-) and increasing (+) the value of the style parameter (the original image is omitted). The layer index, channel index, and the direction of change is overlayed in the bottom left corner.
	}
\end{figure*}

To ensure consistency across a variety of images, we sample 1K different style codes, and compute for each code $s$ and each channel $u$ the semantic category with the highest overlap coefficient: $c_{s,u}^*= \arg\max OC^s_{u,c}$.
Our goal is to detect channels for which the highest overlap category 
is the same for the majority of the sampled images. Furthermore, we require that the overlap with the second most commonly affected category is twice as rare.

\subsection{Experiments} 

We analyze StyleGAN2~\cite{karras2020analyzing} pretrained on FFHQ 1024x1024, LSUN Car 512x384, and LSUN Bedroom 128x128~\cite{yu2015lsun}.
To obtain semantic maps, we use a BiSeNet model \cite{yu2018bisenet} pretrained on CelebAMask-HQ~\cite{lee2020maskgan}, and a unified parsing network~\cite{xiao2018unified} pretrained on Broden+~\cite{bau2017network}.

As explained in Section \ref{sec:disent_level} and \apsec{supp-sec:structureS}, 3040 channels of $\mathcal{S}$ are used to control the tRGB blocks.
None of these channels were found to have a localized effect. Rather, these channels have a global effect on the generated image, as shown in \apfig{fig:PG}.  

Among the remaining 6048 channels, 1871 were found to be locally-active (in the model trained on FFHQ).
Most of the detected channels control clothes ($34.9\%$) or hair ($21\%$). For the model trained on LSUN bedroom, we found 421 locally-active channels, most of which control the bed region (27.6\%).
For StyleGAN2 pretrained on LSUN car, we found 913 locally-active channels, most of which control window (33.1\%) and wheel (27.3\%) regions.
Most of the detected channels are spread among several middle layers, with barely any channels found in early or late layers. 
A detailed summary of the detected locally-active channels and their breakdown by different semantic regions is included in \apsec{ap:Local}.

Figures~\ref{fig:localized_dimenion} and \ref{fig:localized_dimenion2} demonstrate some of the localized manipulations obtained by modifying the values of the channels we detected. Surprisingly, each channel appears to only control a single attribute, and even channels affecting the same local region are well disentangled, as demonstrated in Figure~\ref{fig:disentanglement} and \apfig{fig:disentanglement2}.
Unlike most controls detected by previous methods, these SpaceStyle channels provide an extremely fine-grained level of control. For example, the four channels for the ear region (last row of Figure~\ref{fig:localized_dimenion}), provide separate controls for the visibility of the ear, its shape, and the presence of an earring.
A variety of fine-grained controls are also detected in the Car and Bedroom models (Figure~\ref{fig:localized_dimenion2}). 
It should be noted that finding such interpretable disentangled local controls is very easy with our method: out of the top 10 most localized channels for each semantic area, we observe that 4--10 dimensions control (subjectively) meaningful visual attributes. A detailed breakdown by semantic category is reported in \aptab{tab:meaningful}.

In contrast, individual channels of $\mathcal{W}$ or $\mathcal{W+}$ space are usually entangled, with each channel affecting multiple attributes, as predicted by the 
DCI-based analysis from the previous section. We attribute this to the fact that each channel of $\mathcal{W+}$ affects the style parameters of an entire generation layer (via an affine transformation), rather than those of a single feature map channel.





\section{Detecting attribute-specific channels}
\label{sec:attributes}


In this section we propose a method for identifying StyleSpace channels which control a specific target attribute, specified by a set of examples. For example, given a collection of portraits of grey-haired persons, our goal is to find individual channels that control hair greyness. In contrast to InterFaceGAN~\cite{shen2020interfacegan}, where around 10K positive and 10K negative examples are required, our approach typically requires only 10--30 positive exemplars.
This is an important advantage, since for many attributes, negative examples can be highly varied. For example, while it is easy to find positive examples for blond hair, negative examples should ideally include all non-blond hair colors.

Our approach is based on the simple idea that the differences between the mean style vector of the positive examples (exemplar mean) and that of the entire generated distribution (population mean) reveal which channels are the most relevant for the target attribute.

Specifically, let $\mu^p$ and $\sigma^p$ denote the mean and the standard deviation of the style vectors over the generated distribution.
Given the style vector $s^e$ of a specific positive example, we compute its normalized difference from the population mean: $\delta^{e}=\frac{s^e-\mu^p}{\sigma^p}$.
Next, let $\mu^{e}$ and $\sigma^{e}$ denote the mean and the standard deviation of the differences $\delta^e$ over the exemplar set.
For each style channel $u$, the magnitude of the corresponding component $\mu^{e}_u$ indicates the extent to which $u$ deviates from the population mean.
Thus, we measure the relevance of $u$ with respect to the target attribute as the ratio $\theta_u=\frac{|\mu^{e}_u|}{\sigma^{e}_u}$.
Due to the high disentanglement of $\mathcal{S}$ (Section \ref{sec:disent_level}), a style channel $u$ with a high $\theta_u$ value may be assumed to control the target attribute.

\subsection{Experiments}




We first use a large number (1K) of positive examples to verify that the simple method described above is indeed able to identify a set of attribute-specific control channels. Next, we demonstrate that as few as 10--30 positive examples are sufficient to detect most of these channels.



We first use the set of pretrained classifiers that were used in Section~\ref{sec:disent_level}, to identify 1K highly positive examples for each of selected 26 attributes (see \apsec{ap:annotation_celebA} and \aptab{tab:remove_attributes}). For each attribute, we rank all the style channels (except the 3040 tRGB ones)
by their relevance $\theta_u$, and manually examine the top 30 channels to verify that they indeed control the target attribute.


Our examination reveals that 16 out of the 26 attributes may be controlled by at least one single style channel (see supp.~Table~\ref*{tab:remove_attributes}). The channels detected for each attribute and their ranks are reported in supp.~Table~\ref*{tab:attribute}.
Interestingly, for well-defined visual attributes, such as gender, black hair, or gray hair, our method was able to find only one controlling channel.
In contrast, for less specific attributes, especially those related to hair styles (bangs, receding hairline), we identified multiple controlling channels. We observe that these controls are not redundant, each controlling a unique hair style. The remaining 10 attributes are typically entangled (e.g., high cheekbones, young, or chubby), and thus no disentangled single-channel controls were detected for them. See \apsec{ap:annotation_celebA} for further discussion.

Most of the detected attribute-specific control channels were highly ranked by our proposed importance score 
$\theta_u$. For example, for 14 out of 16 attributes, the top-ranked channel was verified to indeed control the attribute (see \aptab{tab:attribute} for the ranks of all the attribute-specific channels that we detected). This suggests that a small number of positive examples provided by a user might be sufficient for identifying such channels.

\begin{figure}[t]
\begin{center}
\includegraphics[trim={ 0cm 0cm 0cm 0.8cm},clip,scale=0.4]{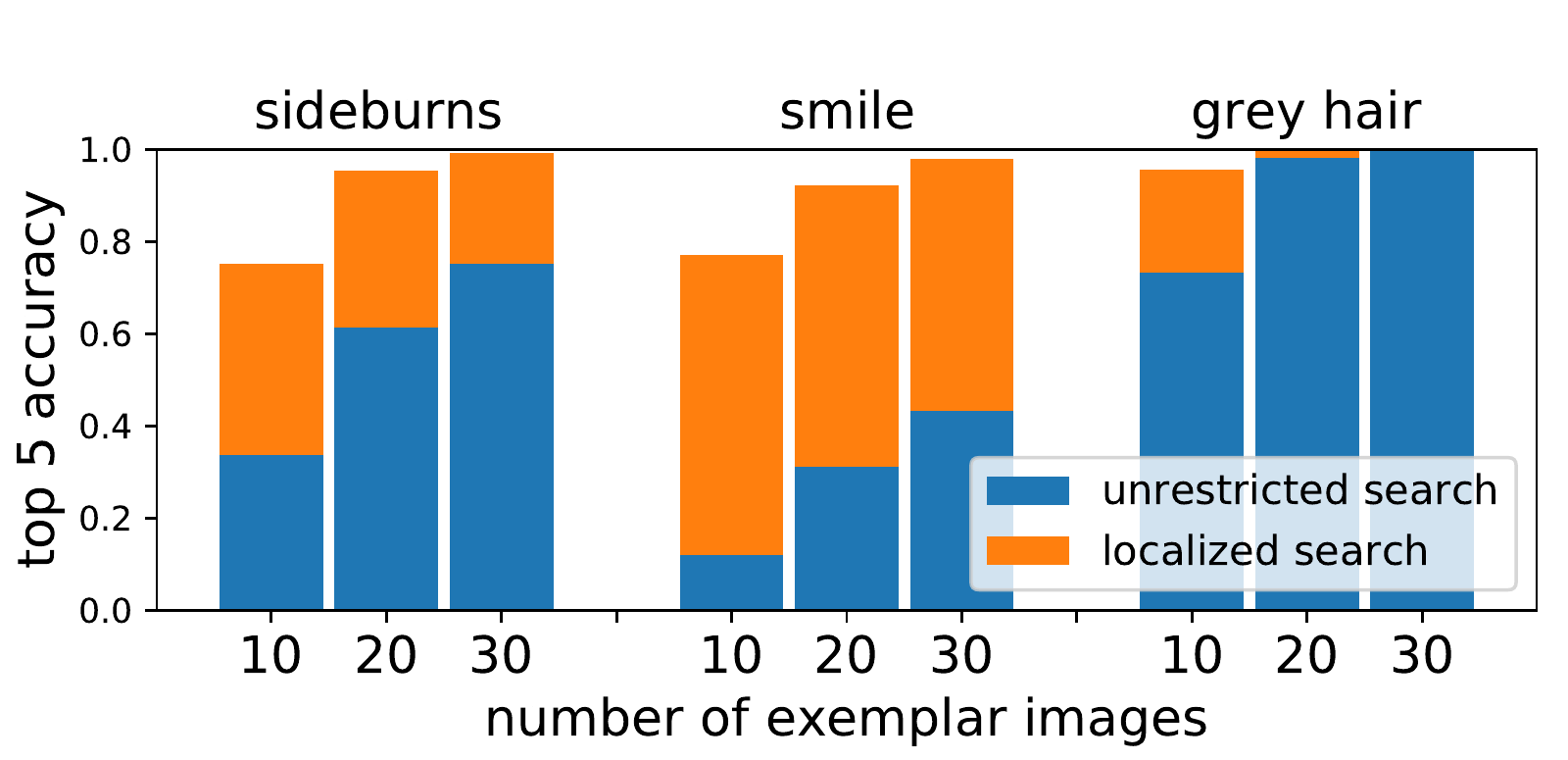} \vspace{-2mm}
\end{center}
   \caption{\label{fig:few examples}
   	Top-5 detection accuracy for attribute-specific controls (for three target attributes) using 10, 20, or 30 positive examples. 
	}
\end{figure}

To verify the above conjecture, we randomly select sets of 10, 20, and 30 positive examples for each of three attributes (sideburns, smile, gray hair) and identify the top 5 channels for each of these small exemplar sets. If the top 5 channels include any of the verified control channels (determined using 1K images), this is considered a success. The results are reported in Figure~\ref{fig:few examples}.

As shown in Figure~\ref{fig:few examples}, increasing the number of positive examples improves the detection accuracy. The accuracy may be further improved by only considering locally-active channels (found as described in Section~\ref{sec:localized}) in areas related to the target attribute. For example, if smile is the target attribute, considering only channels that are active in the mouth area, greatly improves the chances of detection.
As shown by the orange bars in Figure~\ref{fig:few examples}, the top-5 detection rate exceeds 92\% using as few as 20 examples, if the search is restricted to channels locally-active in the target area.

In summary, our approach requires only 10--30 positive examples, and detects single StyleSpace control channels.
In contrast, GANSpace~\cite{harkonen2020ganspace} identifies manipulation controls via a manual examination of a large number of different manipulation directions, which typically involve all of the channels of one or several layers. InterFaceGAN~\cite{shen2020interfacegan} requires more than 10K positive and 10K negative examples for each manipulation direction, which is defined in $\mathcal{W}$ space, and thus affects all layers. Furthermore, the controls detected by these two approaches are more entangled that our control channels, as shown in the next section.




\section{Disentangled attribute manipulation}
\label{sec:ADcomparison}

In this section we compare the ability of our approach to achieve disentangled manipulation of visual attributes to that of two state-of-the-art methods, specifically GANSpace~\cite{harkonen2020ganspace} and InterFaceGAN~\cite{shen2020interfacegan}.

\begin{figure}[tb]
	\setlength{\tabcolsep}{1pt}	
	\begin{tabular}{ccccc}
		& {\footnotesize Original} & {\footnotesize GANSpace} & {\footnotesize InterfaceGAN} & {\footnotesize Ours} \\
		
		\rotatebox{90}{\footnotesize \phantom{kk}Gender} &
		\includegraphics[scale=0.2]{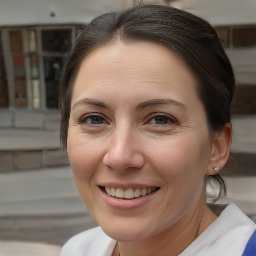} &
		\includegraphics[scale=0.2]{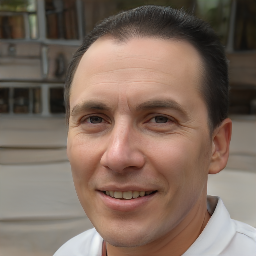} &
		\includegraphics[scale=0.2]{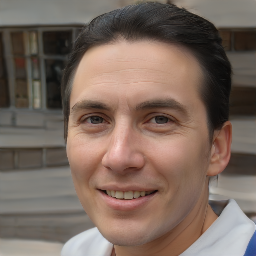} &
		\includegraphics[scale=0.2]{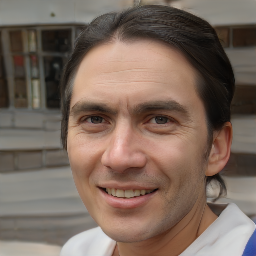} 
		\\
		
		\rotatebox{90}{\footnotesize \phantom{k}Gray hair} &
		\includegraphics[scale=0.2]{./fig4/original.png} &
		\includegraphics[scale=0.2]{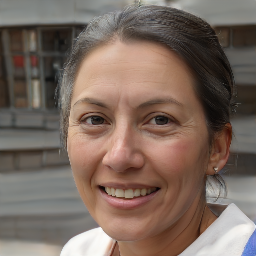} &
		\includegraphics[scale=0.2]{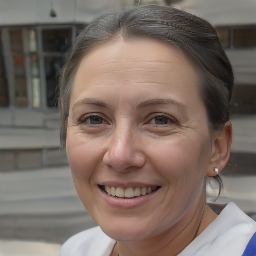} &
		\includegraphics[scale=0.2]{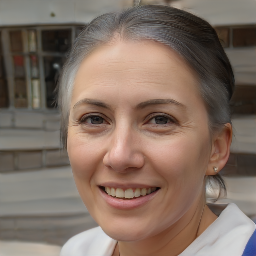} 
		\\
		
		\rotatebox{90}{\footnotesize \phantom{kk}Lipstick} &
		\includegraphics[scale=0.2]{./fig4/original.png} &
		\includegraphics[scale=0.2]{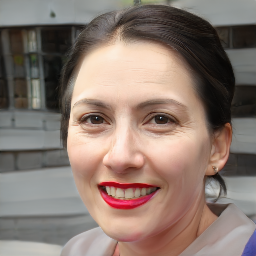} &
		\includegraphics[scale=0.2]{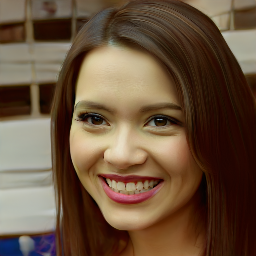} &
		\includegraphics[scale=0.2]{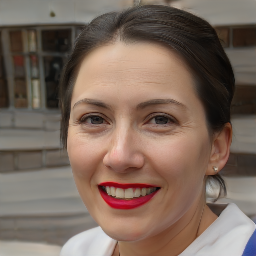} 
	\end{tabular}
	
	\caption{Comparison with state-of-the-art methods using the same amount of manipulation $\Delta l_t = 1.5\sigma(l_t)$. 
		\vspace{-3mm}
	}
	\label{fig:single-vs-all}
\end{figure}

Figure~\ref{fig:single-vs-all} and \apfig{fig:single-vs-all2} show a qualitative comparison between the three methods, showing the manipulation of three attributes for which the direction of manipulation is identified by all three approaches: \textit{Gender}, \textit{Gray hair}, and \textit{Lipstick}. The step size along the manipulation direction is chosen such that it induces the same amount of change in the logit value $l_t$ of the corresponding classifiers (pretrained on CelebA).
Note that InterFaceGAN manipulations sometimes significantly change the identity of the person (esp.~in the \textit{Lipstick} manipulation), and some other attributes as well (added wrinkles in the \textit{Gray hair} manipulation). GANSpace manipulations also exhibit some entanglement (\textit{Lipstick} affects face lightness, \textit{Gray hair} ages the rest of the face). In contrast, our approach appears to affect only the target attribute. Our \textit{Gender} manipulation, for example, does not affect the hair style, and minimally changes the face, yet the gender unmistakably changes. 

To perform a more comprehensive and quantitative comparison between the three methods, we propose a general disentanglement metric for real images, which we refer to as \textit{Attribute Dependency} (AD). Attribute Dependency measures the degree to which manipulation along a certain direction induces changes in other attributes, as measured by classifiers for those attributes (see \apsec{sec:insightAD} for additional details). The use of classifiers here is necessary in order to cope with real images, where the exact factors of variation are not known, and we have no means to measure them. Intuitively, disentangled manipulations should induce smaller changes in other attributes.

\begin{figure*}[tb]
	\setlength{\tabcolsep}{1.8pt}
	
	{\footnotesize
		\begin{tabular}{ccccccccc}
			Original & Inverted & Smile & Lipstick &  Gaze & Eye Shape & Frown Eyebrows & Goatee & Bulbous Nose \\
			\includegraphics[trim={ 0cm 0cm 0cm 0cm},scale=0.2]{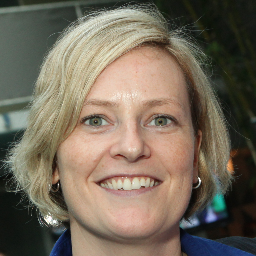} &
			\includegraphics[trim={ 0cm 0cm 0cm 0cm},scale=0.2]{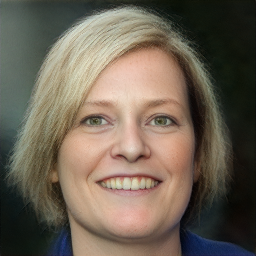} &
			\includegraphics[trim={ 0cm 0cm 0cm 0cm},scale=0.2]{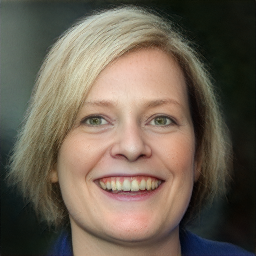} &
			\includegraphics[trim={ 0cm 0cm 0cm 0cm},scale=0.2]{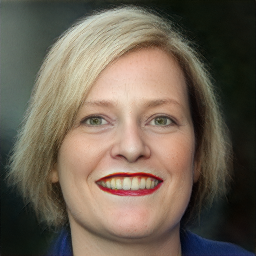} &
			\includegraphics[trim={ 0cm 0cm 0cm 0cm},scale=0.2]{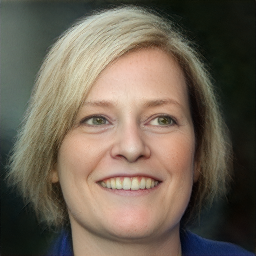} &
			\includegraphics[trim={ 0cm 0cm 0cm 0cm},scale=0.2]{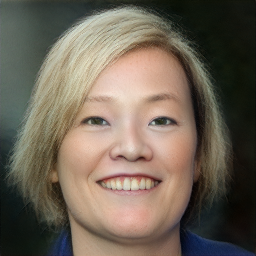} &
			\includegraphics[trim={ 0cm 0cm 0cm 0cm},scale=0.2]{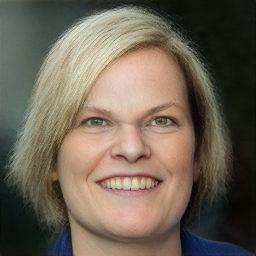}&
			\includegraphics[trim={ 0cm 0cm 0cm 0cm},scale=0.2]{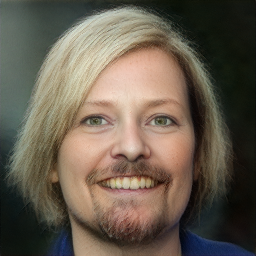}&
			\includegraphics[trim={ 0cm 0cm 0cm 0cm},scale=0.2]{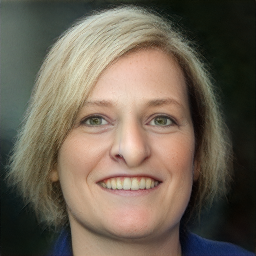}
			\\
			\includegraphics[trim={ 0cm 0cm 0cm 0cm},scale=0.2]{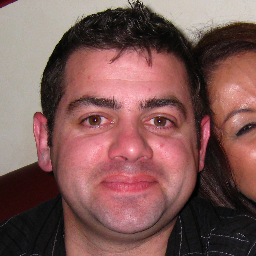} &
			\includegraphics[trim={ 0cm 0cm 0cm 0cm},scale=0.2]{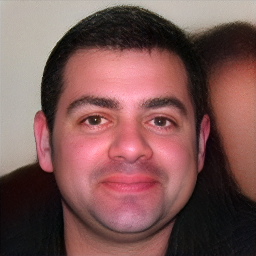} &
			\includegraphics[trim={ 0cm 0cm 0cm 0cm},scale=0.2]{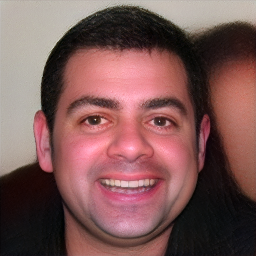} &
			\includegraphics[trim={ 0cm 0cm 0cm 0cm},scale=0.2]{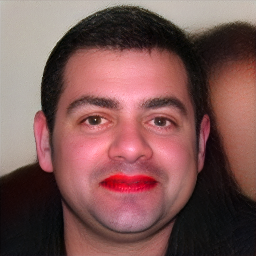} &
			\includegraphics[trim={ 0cm 0cm 0cm 0cm},scale=0.2]{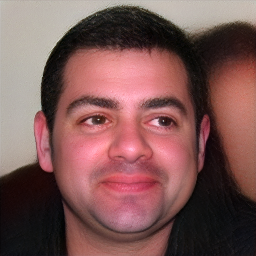} &
			\includegraphics[trim={ 0cm 0cm 0cm 0cm},scale=0.2]{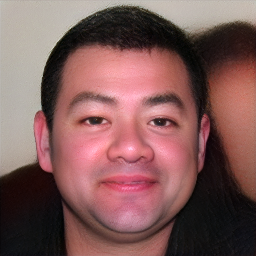} &
			\includegraphics[trim={ 0cm 0cm 0cm 0cm},scale=0.2]{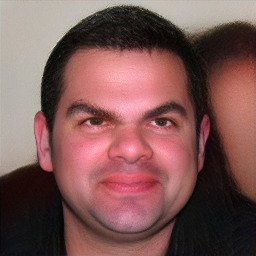}&
			\includegraphics[trim={ 0cm 0cm 0cm 0cm},scale=0.2]{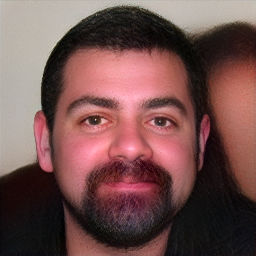}&
			\includegraphics[trim={ 0cm 0cm 0cm 0cm},scale=0.2]{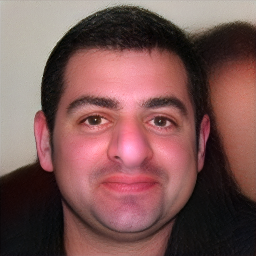}
			\\
			\includegraphics[trim={ 0cm 0cm 0cm 0cm},scale=0.2]{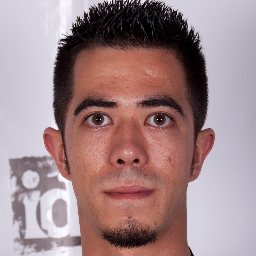} &
			\includegraphics[trim={ 0cm 0cm 0cm 0cm},scale=0.2]{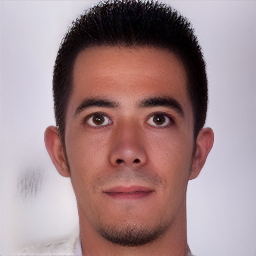} &
			\includegraphics[trim={ 0cm 0cm 0cm 0cm},scale=0.2]{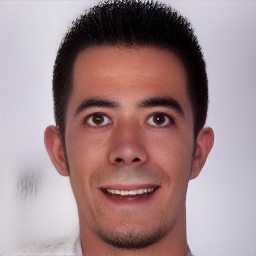} &
			\includegraphics[trim={ 0cm 0cm 0cm 0cm},scale=0.2]{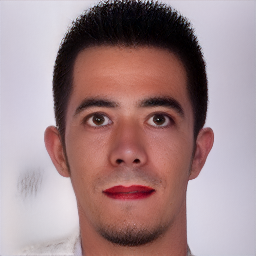} &
			\includegraphics[trim={ 0cm 0cm 0cm 0cm},scale=0.2]{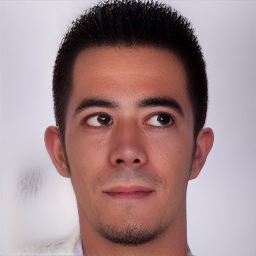} &
			\includegraphics[trim={ 0cm 0cm 0cm 0cm},scale=0.2]{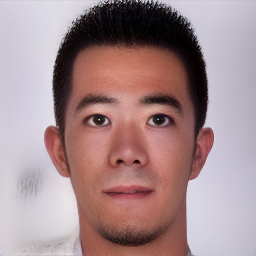} &
			\includegraphics[trim={ 0cm 0cm 0cm 0cm},scale=0.2]{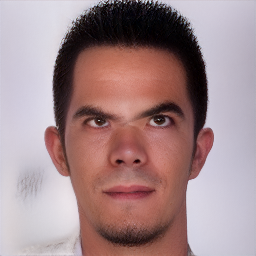}&
			\includegraphics[trim={ 0cm 0cm 0cm 0cm},scale=0.2]{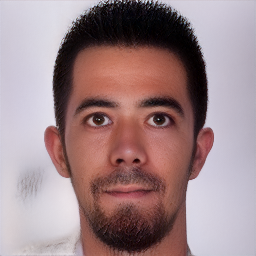}&
			\includegraphics[trim={ 0cm 0cm 0cm 0cm},scale=0.2]{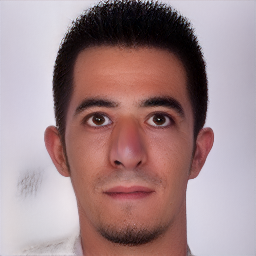}
			\\
		\end{tabular}
	}
	\caption{Manipulation of real images using encoder-based inversion.
		Original images are from FFHQ, and were not part of the encoder's training set. More results can be found in \apfig{fig:real2}. 
	\vspace{-3mm}
	}
	\label{fig:real}
\end{figure*}

\begin{figure}[bt]
	\begin{center}
		\includegraphics[trim={0cm 0cm 0cm 0cm},scale=0.335]{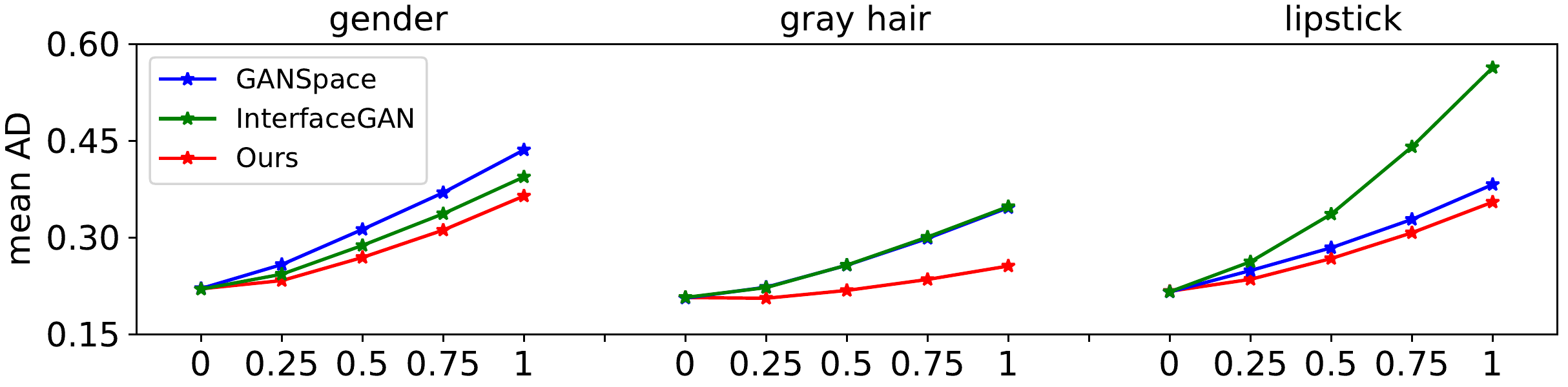} \vspace{-6mm}
	\end{center}
	\caption{Mean-AD vs.~the degree of target attribute manipulation ($\Delta l_t/\sigma(l_t)$). Lower mean-AD indicates better disentanglement.
	\vspace{-3mm}
	}
	\label{fig:meanAD}
\end{figure}

To perform the comparison, we sample a set of images without the target attribute $t$ (e.g., without gray hair), and manipulate them towards the target attribute, by a certain amount measured by the change in the logit outcome $\Delta l_t$ of a classifier pretrained to detect attribute $t$.
Next, we measure the change of logit $\Delta l_i$ between the original images and the manipulated ones for other attributes $\forall i \in \mathcal{A}\backslash{t}$, where $\mathcal{A}$ is the set of all attributes. Each change is normalized by $\sigma(l_i)$, the standard deviation of the logit value for attribute $i$ over a large set of generated images.
We measure mean-AD, defined as $E({1 \over k}\sum_{i \in \mathcal{A}\backslash{t}} (\frac{\Delta l_i}{\sigma(l_i)}))$, where $k = |\mathcal{A}| - 1$.
Similarly, we measure max-AD, defined as $E(\max_{i \in \mathcal{A}\backslash{t}} (\frac{\Delta l_i}{\sigma(l_i)}))$. 

Figure~\ref{fig:meanAD} 
plots the mean-AD of the three methods (GANSpace, InterFaceGAN, and ours) for a range of maniplations of the \textit{Gender}, \textit{Gray hair}, and \textit{Lipstick} attributes.
It may be seen that our method (in red) exhibits a smaller mean-AD, compared to the other two methods, for each of these three attributes and across the entire manipulation range. This is consistent with our qualitative visual observations, as demonstrated in Figure~\ref{fig:single-vs-all}. Our method also achieves lower max-AD scores, as reported in the supplementary material.

\section{Manipulation of Real Images}
\label{sec:real-images}

To manipulate real images, it is necessary to first invert them into latent codes.
This may be done via latent optimization~\cite{abdal2019image2stylegan,abdal2020image2stylegan++} or by training an encoder~\cite{zhu2016generative,zhu2020domain} based on reconstruction loss (LPIPS~\cite{zhang2018unreasonable} or L2). We adapt the latent optimization algorithm of Karras \etal~\cite{karras2020analyzing} to invert real images into $\mathcal{W}$, $\mathcal{W+}$, and $\mathcal{S}$ separately.
Latent optimization in $\mathcal{W+}$ and $\mathcal{S}$ spaces has more flexibility than in $\mathcal{W}$, enabling a closer reconstruction of the input image.
Indeed, we find that the visual accuracy of the reconstruction is the highest when optimizing in $\mathcal{S}$, followed by $\mathcal{W+}$, and is the lowest for $\mathcal{W}$ (see \apfig{fig:invert}).
Unfortunately, the extra flexibility may result in latent codes that do not lie on the generated image manifold, and attempting to manipulate such codes typically results in unnatural artifacts.
Thus, conversely to reconstruction accuracy, we find that manipulation naturalness is best when the latent optimization is done in $\mathcal{W}$, followed by $\mathcal{W+}$, and the worst for $\mathcal{S}$ (see \apfig{fig:invert_m}).


In order to achieve a satisfactory compromise between reconstruction accuracy and artifact-free manipulation, we train an encoder to $\mathcal{S}$ space following the training strategy of~\cite{zhu2020domain} using only reconstruction loss (LPIPS). The encoder's structure follows that of StyleALAE~\cite{pidhorskyi2020adversarial}.
Due to limited computational resources, the encoder is trained on real images from FFHQ whose resolution was reduced to $128 \times 128$. The reconstructed images bear good similarity to the input images, but exhibit some compression artifacts. The encoder's result serves as the starting point for latent optimization~\wu{\cite{karras2020analyzing}}
in $\mathcal{S}$ space, which proceeds for a small number of iterations (50 rather than a few thousands).
We find that this process can efficiently remove compression artifacts, and the resulting inversions enable artifact-free manipulation, as demonstrated in Figure~\ref{fig:real}. We believe this is because the encoder learns to embed the input real images closer to the generated image manifold, and the few optimization iterations only fine-tune the embedding.


\section{Conclusion}
\vspace{-1mm}
We have shown that StyleSpace is highly disentangled, and proposed simple methods for detecting meaningful manipulation controls in this space. Future work should focus on finding meaningful control directions that involve multiple style channels. We also plan to develop inversion techniques that can deliver both high reconstruction accuracy and manipulability.


{\small
\bibliographystyle{ieee_fullname}
\bibliography{egbib}
}

\newpage
\phantom{k}
\newpage
\onecolumn
\appendix


\section{Structure of StyleGAN2 StyleSpace}
\label{supp-sec:structureS}

\begin{figure*}[b]
	\includegraphics[trim={ 0cm 0cm 0cm 0cm},width=\textwidth]{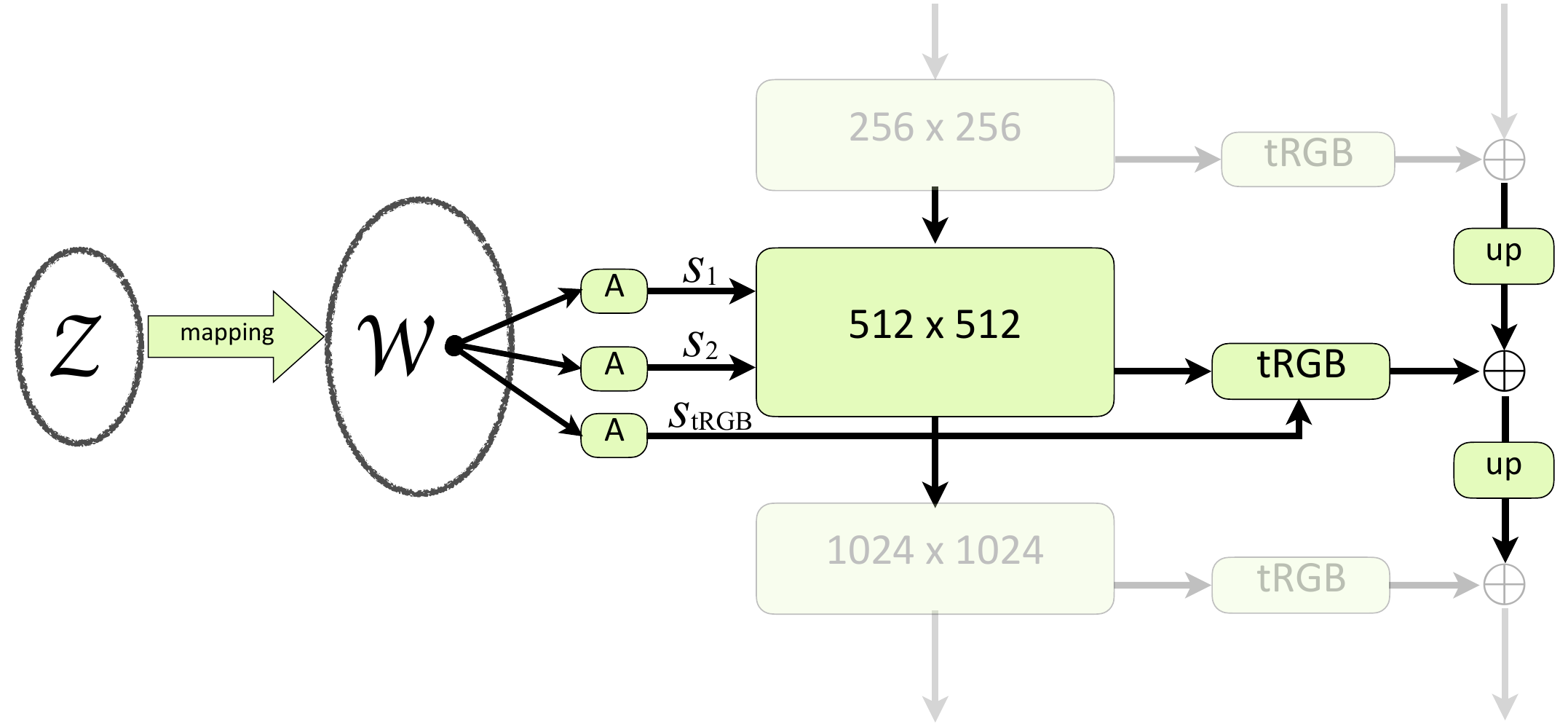}
	\caption{The internal structure of StyleSpace $\mathcal{S}$, shown for the $512\times 512$ generator resolution.}
	\label{fig:structure_s}
\end{figure*}

To supplement the description of the different StyleGAN2 latent spaces in Section~\ref{sec:disent_level}, here we describe the structure of the StyleSpace $\mathcal{S}$ in more detail.
Every major layer (every resolution) of the StyleGAN2 generator (synthesis network) consists of two convolution layers for feature map synthesis and a single convolution layer that converts the second feature map into an RGB image (referred to as tRGB), as shown in Figure~\ref{fig:structure_s}. Each of these three convolution layers is modulated by a vector of style parameters. We denote the three different vectors of style parameters as $s_1$, $s_2$, and $s_{tRGB}$. These are obtained from the intermediate latent vectors $w \in \mathcal{W}$ via three affine transformations, $w_1 \rightarrow s_1$, $w_2 \rightarrow s_2$, $w_2 \rightarrow s_{tRGB}$. In $\mathcal{W}$ space, $w_1$ and $w_2$ are the same vector, and it is the same vector for all layers. In $\mathcal{W+}$ space, $w_1$ and $w_2$ are two different vectors, and every major layer has its own pair $(w_1, w_2)$. The length of all the $w$ vectors is 512. The numbers of style parameters used by the different layers are listed in Table~\ref{tab:structure_s}. Note that in 4x4 resolution, there is only $s_1$ and $s_{tRGB}$. The length of $s$ is 512 from the early layers until layer 14. After that layer, the length decreases from 256 to 32.
In total, for a 1024x1024 generator, there are 6048 style channels that control feature maps, and 3040 additional channels that control the tRGB blocks.

\begin{table*}[t]
\begin{center}
\begin{tabular}{cccllc}
$\mathcal{W+}$ layer index & $\mathcal{S}$ layer index & resolution & layer name & type & \# channels \\
\rowcolor[HTML]{C0C0C0} 
0 & 0 & 4$\times$4 & Conv & $s_1$ & 512 \\
\rowcolor[HTML]{C0C0C0} 
1 & 1 & 4$\times$4 & ToRGB & $s_{tRGB}$ & 512 \\
2 & 2 & 8$\times$8 & Conv0\_up & $s_1$ & 512 \\
3 & 3 & 8$\times$8 & Conv1 & $s_2$ & 512 \\
3 & 4 & 8$\times$8 & ToRGB & $s_{tRGB}$ & 512 \\
\rowcolor[HTML]{C0C0C0} 
4 & 5 & 16$\times$16 & Conv0\_up & $s_1$ & 512 \\
\rowcolor[HTML]{C0C0C0} 
5 & 6 & 16$\times$16 & Conv1 & $s_2$ & 512 \\
\rowcolor[HTML]{C0C0C0} 
5 & 7 & 16$\times$16 & ToRGB & $s_{tRGB}$& 512 \\
6 & 8 & 32$\times$32 & Conv0\_up & $s_1$ & 512 \\
7 & 9 & 32$\times$32 & Conv1 & $s_2$ & 512 \\
7 & 10 & 32$\times$32 & ToRGB & $s_{tRGB}$ & 512 \\
\rowcolor[HTML]{C0C0C0} 
8 & 11 & 64$\times$64 & Conv0\_up & $s_1$ & 512 \\
\rowcolor[HTML]{C0C0C0} 
9 & 12 & 64$\times$64 & Conv1 & $s_2$ & 512 \\
\rowcolor[HTML]{C0C0C0} 
9 & 13 & 64$\times$64 & ToRGB & $s_{tRGB}$ & 512 \\
10 & 14 & 128$\times$128 & Conv0\_up & $s_1$ & 512 \\
11 & 15 & 128$\times$128 & Conv1 & $s_2$ & 256 \\
11 & 16 & 128$\times$128 & ToRGB & $s_{tRGB}$ & 256 \\
\rowcolor[HTML]{C0C0C0} 
12 & 17 & 256$\times$256 & Conv0\_up & $s_1$ & 256 \\
\rowcolor[HTML]{C0C0C0} 
13 & 18 & 256$\times$256 & Conv1 & $s_2$ & 128 \\
\rowcolor[HTML]{C0C0C0} 
13 & 19 & 256$\times$256 & ToRGB & $s_{tRGB}$& 128 \\
14 & 20 & 512$\times$512 & Conv0\_up & $s_1$ & 128 \\
15 & 21 & 512$\times$512 & Conv1 & $s_2$ & 64 \\
15 & 22 & 512$\times$512 & ToRGB & $s_{tRGB}$ & 64 \\
\rowcolor[HTML]{C0C0C0} 
16 & 23 & 1024$\times$1024 & Conv0\_up & $s_1$ & 64 \\
\rowcolor[HTML]{C0C0C0} 
17 & 24 & 1024$\times$1024 & Conv1 & $s_2$ & 32 \\
\rowcolor[HTML]{C0C0C0} 
17 & 25 & 1024$\times$1024 & ToRGB & $s_{tRGB}$ & 32
\end{tabular}
\end{center}
\caption{\label{tab:structure_s} Breakdown of StyleSpace channels by generator layers. 
} 
\end{table*}

\newpage
\phantom{k}
\newpage
\section{Effect of style parameters in tRGB layers}
\label{ap:tRGB}

To examine the function of style parameters that control the tRGB layers, we randomly generate a set of 500K style vectors $s \in \mathcal{S}$, and perturb their $s_{tRGB}$ channels to manipulate the tRGB layers, $s_{new}=s_{original}+n\sigma(s)$. $\sigma(s)$ is the standard deviation of each channel of $s$ over the generated set, used to normalize the amount of perturbation across different channels~\cite{harkonen2020ganspace}. $n$ is a vector of Gaussian noise, with mean 0 and standard deviation $\sigma(n)$, which indicates the manipulation strength. Below, we use $\sigma(n)=15$. 

\begin{figure}[h]
\setlength{\tabcolsep}{1.8pt}
\begin{center}
\begin{tabular}{ccccccccccc}
\footnotesize Original & \footnotesize Early & \footnotesize Mid & \footnotesize Late & \footnotesize All& \phantom{kk}&\footnotesize Original & \footnotesize Early & \footnotesize Mid & \footnotesize Late & \footnotesize All
\\
\includegraphics[trim={ 0cm 0cm 0cm 0cm},scale=0.17]{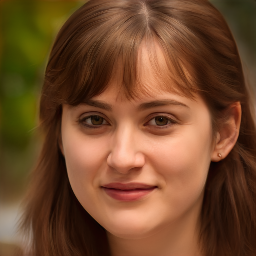}&
\includegraphics[trim={ 0cm 0cm 0cm 0cm},scale=0.17]{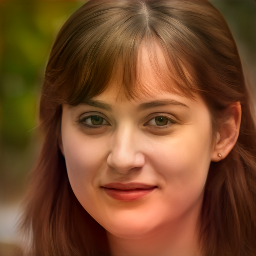} &
\includegraphics[trim={ 0cm 0cm 0cm 0cm},scale=0.17]{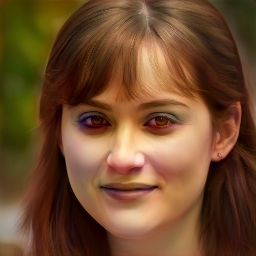} &
\includegraphics[trim={ 0cm 0cm 0cm 0cm},scale=0.17]{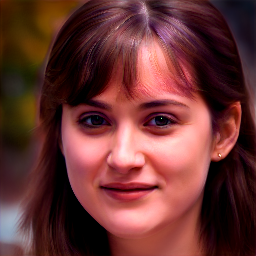} &
\includegraphics[trim={ 0cm 0cm 0cm 0cm},scale=0.17]{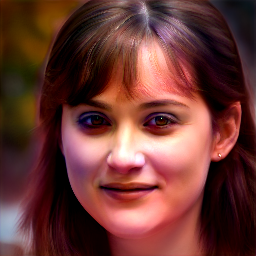} &
&
\includegraphics[trim={ 0cm 0cm 0cm 0cm},scale=0.17]{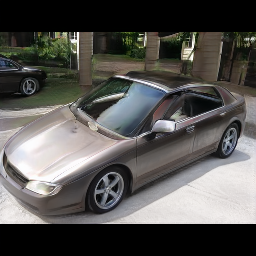}&
\includegraphics[trim={ 0cm 0cm 0cm 0cm},scale=0.17]{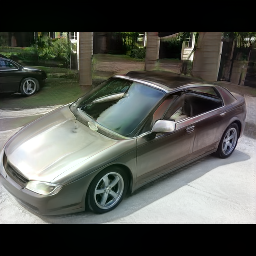} &
\includegraphics[trim={ 0cm 0cm 0cm 0cm},scale=0.17]{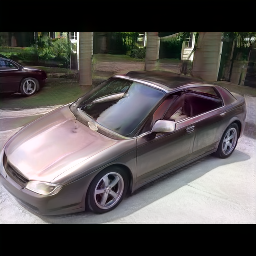} &
\includegraphics[trim={ 0cm 0cm 0cm 0cm},scale=0.17]{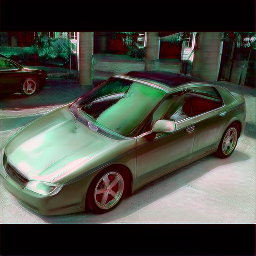} &
\includegraphics[trim={ 0cm 0cm 0cm 0cm},scale=0.17]{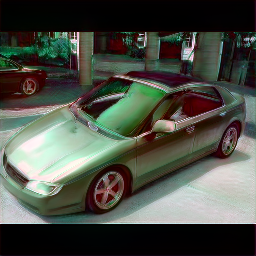} 
\\
\includegraphics[trim={ 0cm 0cm 0cm 0cm},scale=0.17]{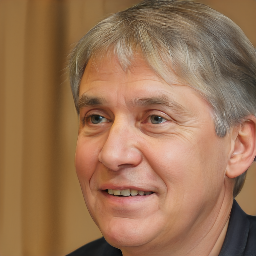}&
\includegraphics[trim={ 0cm 0cm 0cm 0cm},scale=0.17]{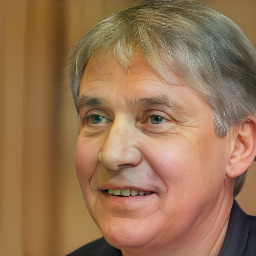} &
\includegraphics[trim={ 0cm 0cm 0cm 0cm},scale=0.17]{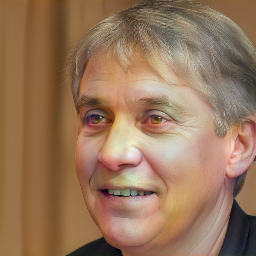} &
\includegraphics[trim={ 0cm 0cm 0cm 0cm},scale=0.17]{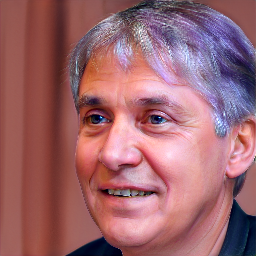} &
\includegraphics[trim={ 0cm 0cm 0cm 0cm},scale=0.17]{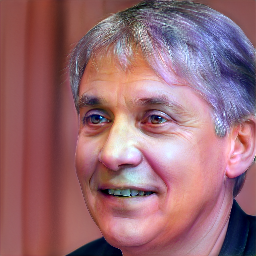} &
&
\includegraphics[trim={ 0cm 0cm 0cm 0cm},scale=0.17]{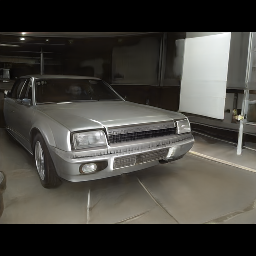}&
\includegraphics[trim={ 0cm 0cm 0cm 0cm},scale=0.17]{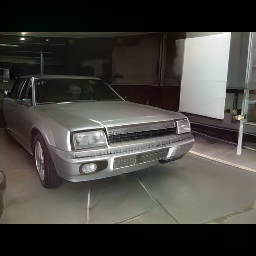} &
\includegraphics[trim={ 0cm 0cm 0cm 0cm},scale=0.17]{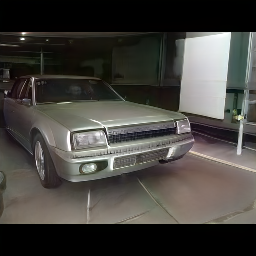} &
\includegraphics[trim={ 0cm 0cm 0cm 0cm},scale=0.17]{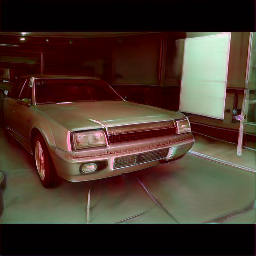} &
\includegraphics[trim={ 0cm 0cm 0cm 0cm},scale=0.17]{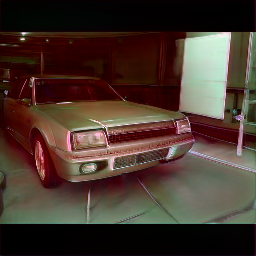} 
\\
\includegraphics[trim={ 0cm 0cm 0cm 0cm},scale=0.17]{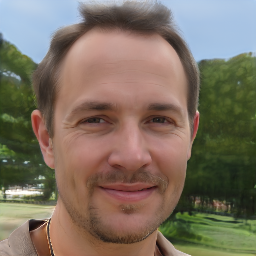}&
\includegraphics[trim={ 0cm 0cm 0cm 0cm},scale=0.17]{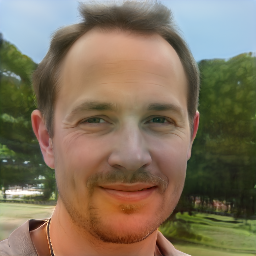} &
\includegraphics[trim={ 0cm 0cm 0cm 0cm},scale=0.17]{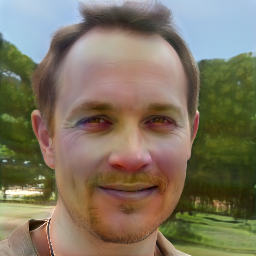} &
\includegraphics[trim={ 0cm 0cm 0cm 0cm},scale=0.17]{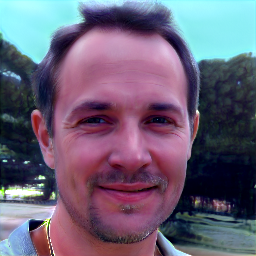} &
\includegraphics[trim={ 0cm 0cm 0cm 0cm},scale=0.17]{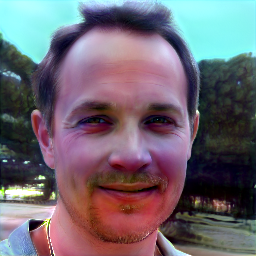} &
&
\includegraphics[trim={ 0cm 0cm 0cm 0cm},scale=0.17]{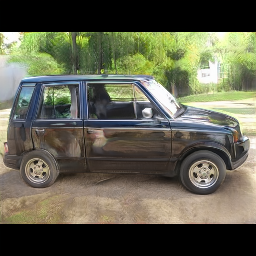}&
\includegraphics[trim={ 0cm 0cm 0cm 0cm},scale=0.17]{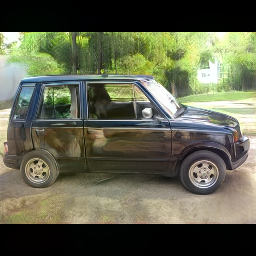} &
\includegraphics[trim={ 0cm 0cm 0cm 0cm},scale=0.17]{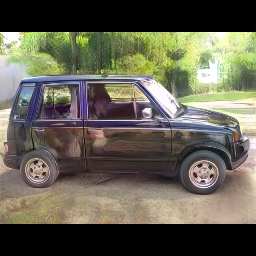} &
\includegraphics[trim={ 0cm 0cm 0cm 0cm},scale=0.17]{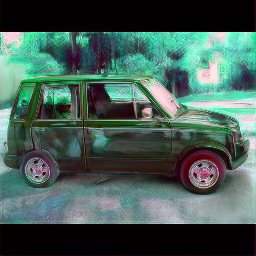} &
\includegraphics[trim={ 0cm 0cm 0cm 0cm},scale=0.17]{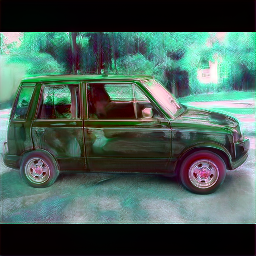} 
\\
\footnotesize Original & \footnotesize Early & \footnotesize Mid & \footnotesize Late & \footnotesize All&&\footnotesize Original & \footnotesize Early & \footnotesize Mid & \footnotesize Late & \footnotesize All
\\
\includegraphics[trim={ 0cm 0cm 0cm 0cm},scale=0.17]{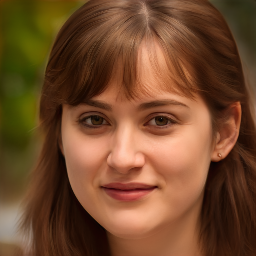}&
\includegraphics[trim={ 0cm 0cm 0cm 0cm},scale=0.17]{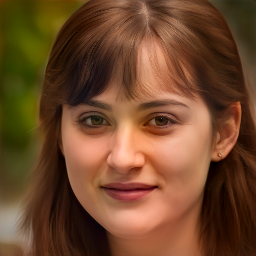} &
\includegraphics[trim={ 0cm 0cm 0cm 0cm},scale=0.17]{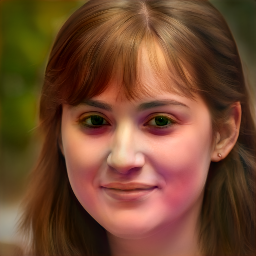} &
\includegraphics[trim={ 0cm 0cm 0cm 0cm},scale=0.17]{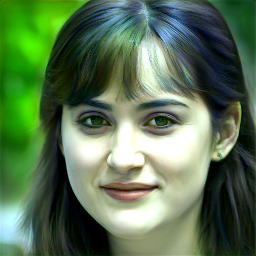} &
\includegraphics[trim={ 0cm 0cm 0cm 0cm},scale=0.17]{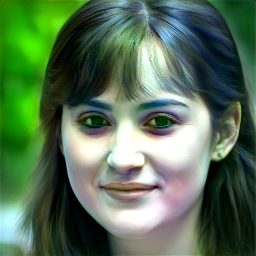} &
&
\includegraphics[trim={ 0cm 0cm 0cm 0cm},scale=0.17]{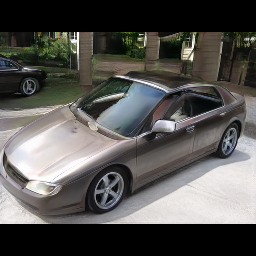}&
\includegraphics[trim={ 0cm 0cm 0cm 0cm},scale=0.17]{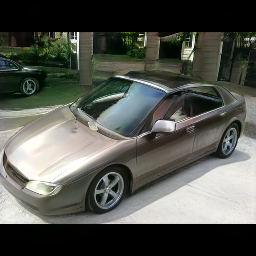} &
\includegraphics[trim={ 0cm 0cm 0cm 0cm},scale=0.17]{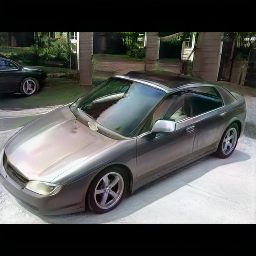} &
\includegraphics[trim={ 0cm 0cm 0cm 0cm},scale=0.17]{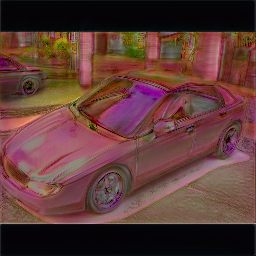} &
\includegraphics[trim={ 0cm 0cm 0cm 0cm},scale=0.17]{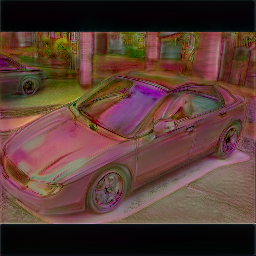} 
\\
\includegraphics[trim={ 0cm 0cm 0cm 0cm},scale=0.17]{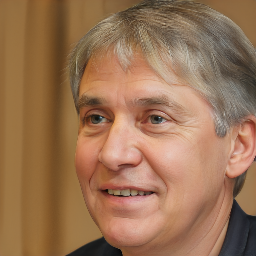}&
\includegraphics[trim={ 0cm 0cm 0cm 0cm},scale=0.17]{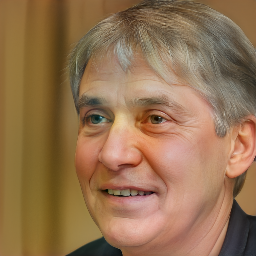} &
\includegraphics[trim={ 0cm 0cm 0cm 0cm},scale=0.17]{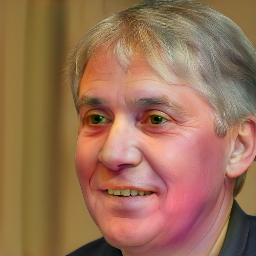} &
\includegraphics[trim={ 0cm 0cm 0cm 0cm},scale=0.17]{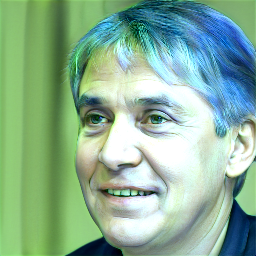} &
\includegraphics[trim={ 0cm 0cm 0cm 0cm},scale=0.17]{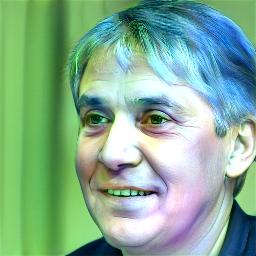} &
&
\includegraphics[trim={ 0cm 0cm 0cm 0cm},scale=0.17]{./PG2/4/0_0.png}&
\includegraphics[trim={ 0cm 0cm 0cm 0cm},scale=0.17]{./PG2/4/0_1.png} &
\includegraphics[trim={ 0cm 0cm 0cm 0cm},scale=0.17]{./PG2/4/0_2.png} &
\includegraphics[trim={ 0cm 0cm 0cm 0cm},scale=0.17]{./PG2/4/0_3.png} &
\includegraphics[trim={ 0cm 0cm 0cm 0cm},scale=0.17]{./PG2/4/0_4.png} 
\\
\includegraphics[trim={ 0cm 0cm 0cm 0cm},scale=0.17]{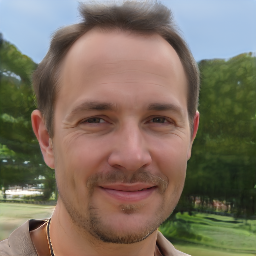}&
\includegraphics[trim={ 0cm 0cm 0cm 0cm},scale=0.17]{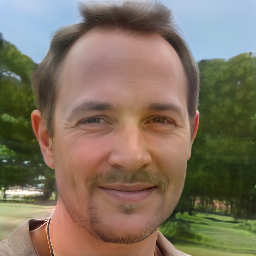} &
\includegraphics[trim={ 0cm 0cm 0cm 0cm},scale=0.17]{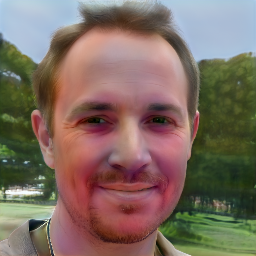} &
\includegraphics[trim={ 0cm 0cm 0cm 0cm},scale=0.17]{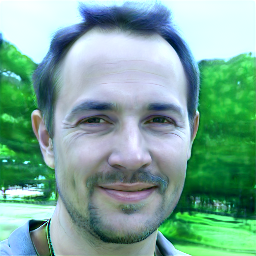} &
\includegraphics[trim={ 0cm 0cm 0cm 0cm},scale=0.17]{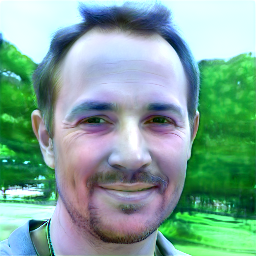} &
&
\includegraphics[trim={ 0cm 0cm 0cm 0cm},scale=0.17]{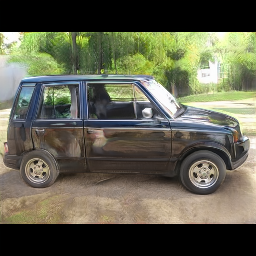}&
\includegraphics[trim={ 0cm 0cm 0cm 0cm},scale=0.17]{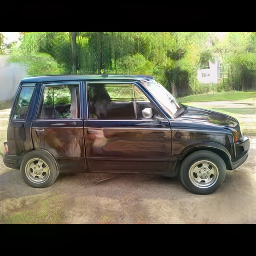} &
\includegraphics[trim={ 0cm 0cm 0cm 0cm},scale=0.17]{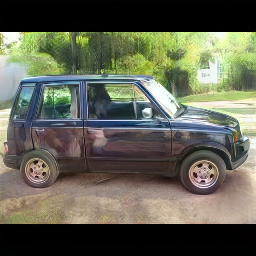} &
\includegraphics[trim={ 0cm 0cm 0cm 0cm},scale=0.17]{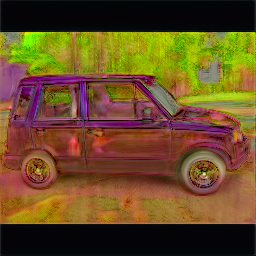} &
\includegraphics[trim={ 0cm 0cm 0cm 0cm},scale=0.17]{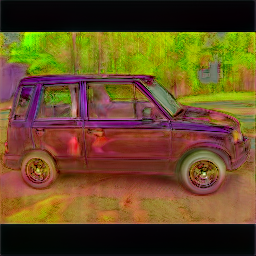} 
\end{tabular}
\end{center}
   \caption{Manipulation by perturbing the $s_{tRGB}$ channels in early resolution layers (0,1,2), middle layers (3,4,5) and late layers (6,7,8). Each of the four blocks above uses the same noise vector $n$. }
\label{fig:PG}
\end{figure}

As shown in Figure~\ref{fig:PG}, manipulating the early (coarse) resolutions (0,1,2) mainly affects the center of the target object (better visible in faces than in cars), manipulating the middle resolutions (3,4,5) typically affects the entire target object, and manipulating the late (fine) resolution layers (6,7,8) affects the entire image. (The LSUN Car model reaches only up to 512$\times$512, thus the fine resolution layers are (6,7)). The effect of the late (fine) resolution layers on the image is significantly stronger than that of the early and middle layers. The manipulations only affect color, without modifying shape or specific face or car related attributes. 

\newpage
\section{Locally-active style channels} \label{ap:Local}

\begin{figure*}[h]
\begin{center}
\begin{tabular}{ccc}
\includegraphics[trim={ 2.5cm 0cm 1.5cm 0cm},scale=0.4]{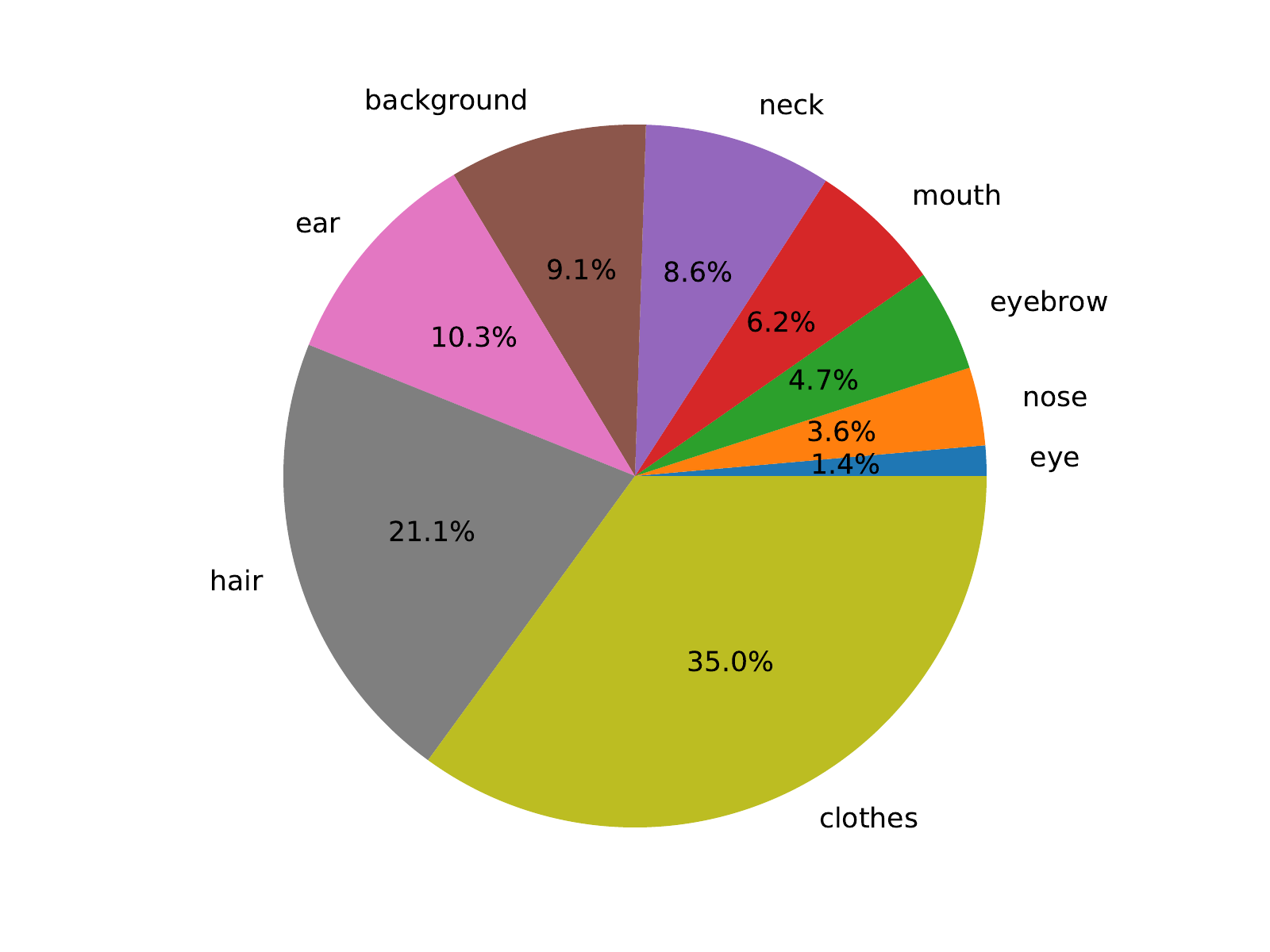}\hspace{3mm} &
\includegraphics[trim={ 2cm 0cm 2cm 0cm},scale=0.4]{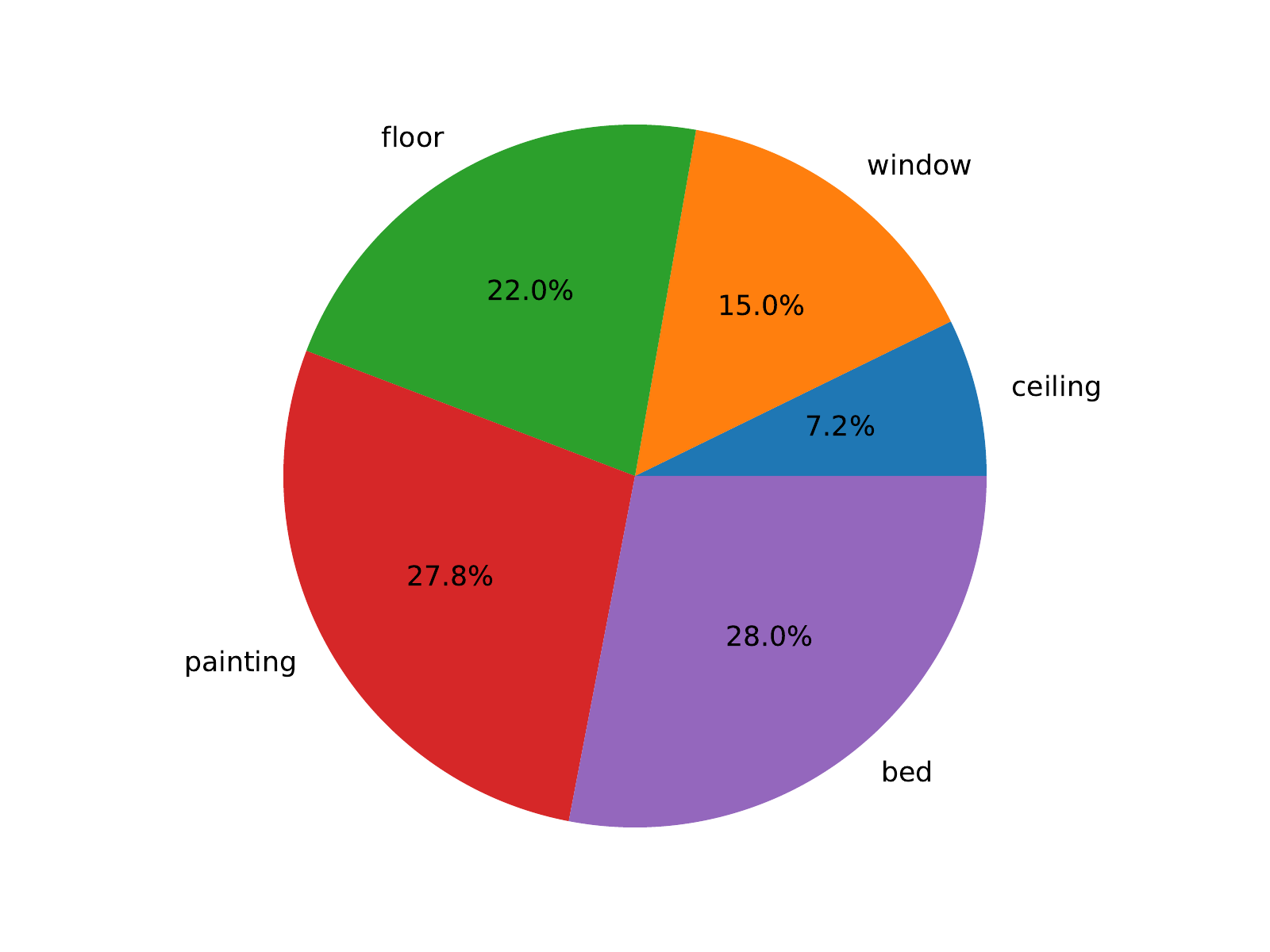}\hspace{3mm} &
\includegraphics[trim={ 2cm 0cm 2cm 0cm},scale=0.4]{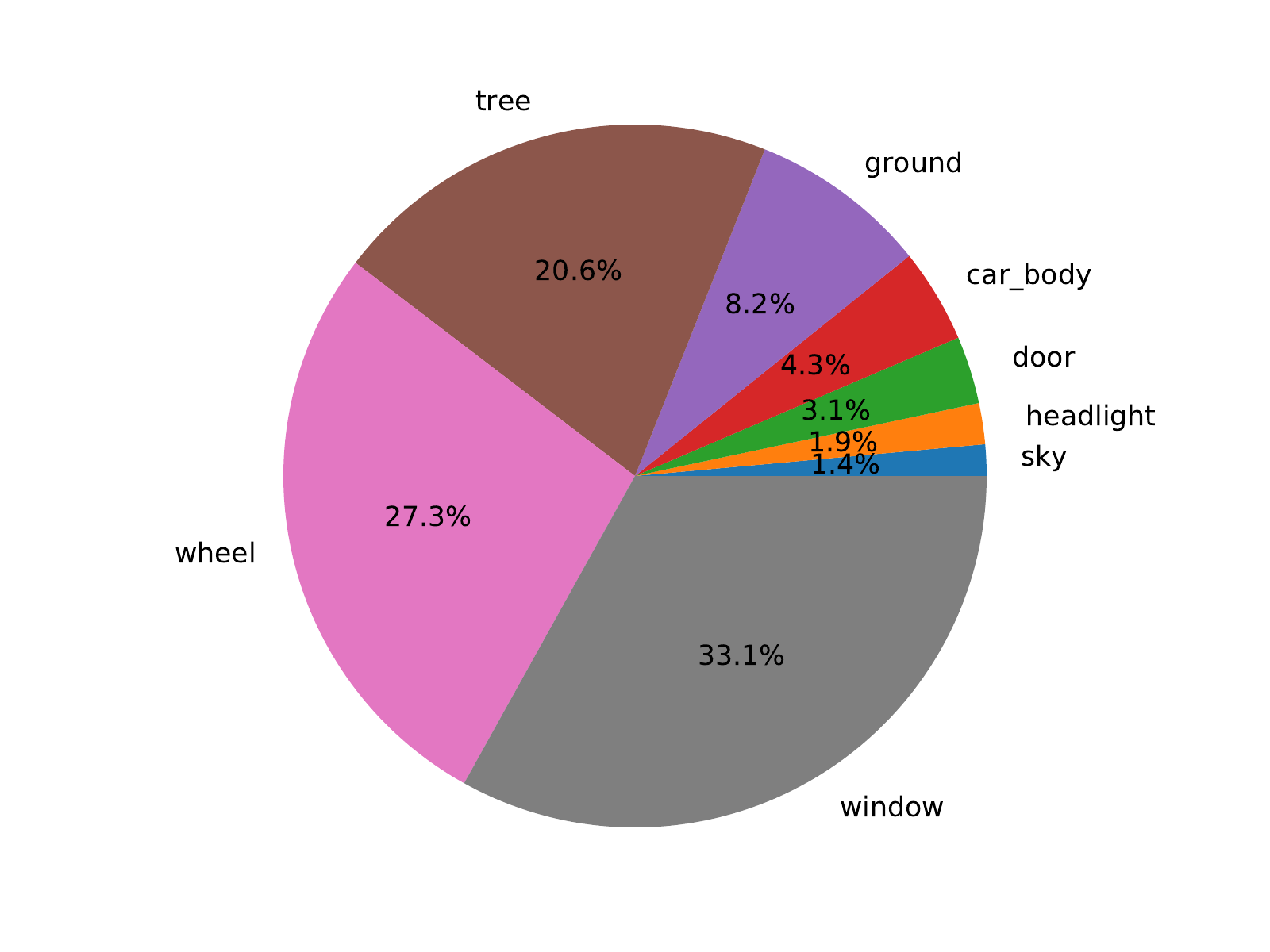}\\
{\small (a) FFHQ} & {\small (b) LSUN Bedroom} & (c) {\small LSUN Car}
\end{tabular}
\end{center}
   \caption{Breakdown of detected localized controls across different semantic regions. 
}
\label{fig:local_distribution}
\end{figure*}

The number of locally-active channels that we found using the method in Section~\ref{sec:localized} for each of the three models we experimented with is summarized in Table \ref{tab:Num_loc}. The breakdown of these localized controls across different semantic regions is plotted in Figure~\ref{fig:local_distribution}. 
Not all of the detected controls correspond to semantically meaningful manipulations. While there is no way to objectively determine which manipulations are meaningful, in Table~\ref{tab:meaningful} we report the number of manipulations that were (subjectively) determined as meaningful by the authors, among the most highly localized controls.

\begin{table}[h]
\begin{center}
\begin{tabular}{ l c c c } 
  & \textbf{FFHQ} & \textbf{LSUN Bedroom}  &\textbf{LSUN Car}\\ 
      \hline \\
Num. locally-active channels & 1871 & 421 & 913 \\ 
Total num. of feature map style channels & 6048 & 5376 & 5952\\ 
Percent of locally-active channels & $30.9\%$ & $7.8\%$ & 15.3\%\\ 
\end{tabular}
\end{center}
\caption{\label{tab:Num_loc} Number of locally-active channels detected in different StyleGAN2 models.} 
\end{table}

\begin{table}[h]
\begin{center}
\begin{tabular}{ l c c c } 
  & Top 5 & Top 10  & Top 20 \\ 
      \hline \\
Eyebrows & 5 & 10 & 19 \\ 
Hair & 5 & 9 & 17 \\ 
Nose & 4 & 7 & 13 \\ 
Mouth & 4 & 7 & 11 \\
Clothes & 5 & 6 & 9 \\ 
Neck & 2 & 4 & 7 \\ 
Eye & 4 & 5 & 6 \\ 
Ear & 3 & 4 & 6 \\ 
Background & 5 & 10 & 15 \\ 
\end{tabular}
\end{center}

\caption{\label{tab:meaningful}
Number of meaningful controls among the top $k=5,10,20$ most locally-active channels (those with the highest overlap coefficient, as defined by equation (\ref{OC}) in the main paper) in each semantic area (for the FFHQ model). Note that this count is subjective and may contain channels that control similar things (for example, size of lips).
} 
\end{table}

\begin{figure}[h]
	\begin{center}
		\begin{tabular}{l}
			\footnotesize \hspace{1.7cm} Smile (6\_501)
			\\
			\rotatebox{90}{\phantom{K}}
			\includegraphics[trim={ 0cm 0cm 0cm 0cm},scale=0.15]{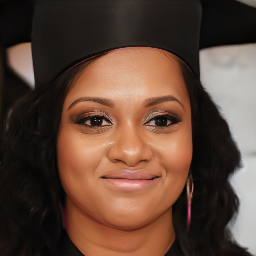} 
			\includegraphics[trim={ 0cm 0cm 0cm 0cm},scale=0.15]{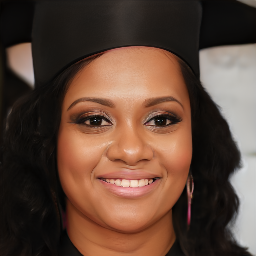} 
			\includegraphics[trim={ 0cm 0cm 0cm 0cm},scale=0.15]{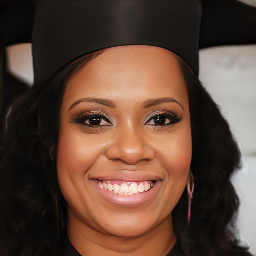} \\
			
			\rotatebox{90}{\footnotesize \hspace{-0.4cm} Lipstick (15\_45) \hspace{-0.4cm}}
			\includegraphics[trim={ 0cm 0cm 0cm 0cm},scale=0.15]{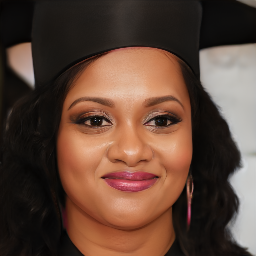} 
			\includegraphics[trim={ 0cm 0cm 0cm 0cm},scale=0.15]{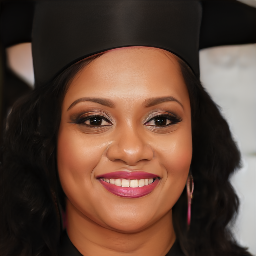} 
			\includegraphics[trim={ 0cm 0cm 0cm 0cm},scale=0.15]{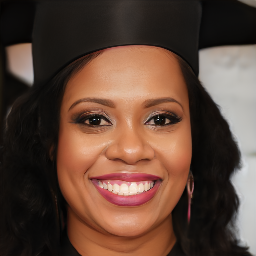}\\
			
			\rotatebox{90}{\phantom{K}}
			\includegraphics[trim={ 0cm 0cm 0cm 0cm},scale=0.15]{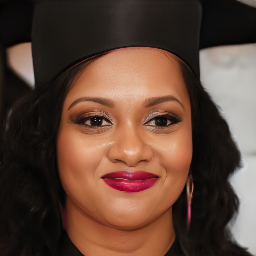} 
			\includegraphics[trim={ 0cm 0cm 0cm 0cm},scale=0.15]{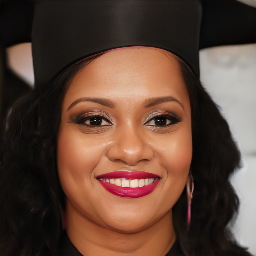} 
			\includegraphics[trim={ 0cm 0cm 0cm 0cm},scale=0.15]{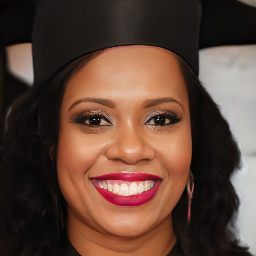}\\
		\end{tabular}
		\begin{tabular}{l}
			\footnotesize  \hspace{1.3cm} Wall Color (12\_91)
			\\
			\rotatebox{90}{\phantom{K}}
			\includegraphics[trim={ 0cm 0cm 0cm 0cm},scale=0.15]{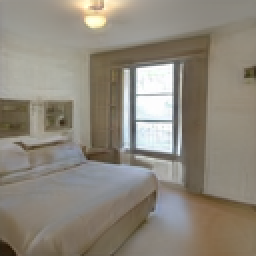} 
			\includegraphics[trim={ 0cm 0cm 0cm 0cm},scale=0.15]{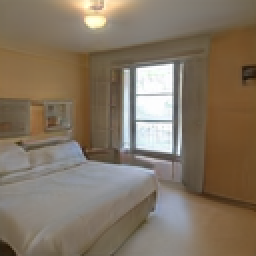} 
			\includegraphics[trim={ 0cm 0cm 0cm 0cm},scale=0.15]{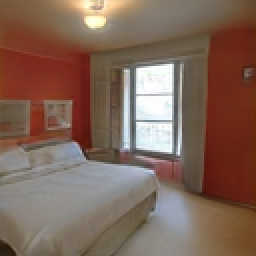} \\
			
			\rotatebox{90}{\footnotesize \hspace{-0.4cm} Floor Color (8\_358) \hspace{-0.7cm}}
			\includegraphics[trim={ 0cm 0cm 0cm 0cm},scale=0.15]{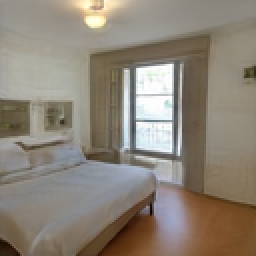} 
			\includegraphics[trim={ 0cm 0cm 0cm 0cm},scale=0.15]{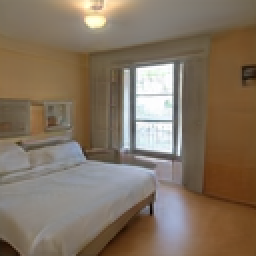} 
			\includegraphics[trim={ 0cm 0cm 0cm 0cm},scale=0.15]{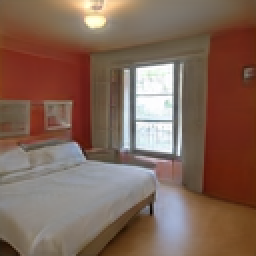}\\
			
			\rotatebox{90}{\phantom{K}}
			\includegraphics[trim={ 0cm 0cm 0cm 0cm},scale=0.15]{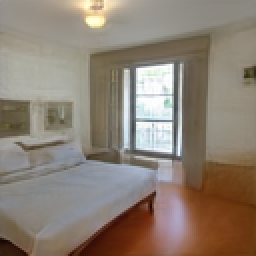} 
			\includegraphics[trim={ 0cm 0cm 0cm 0cm},scale=0.15]{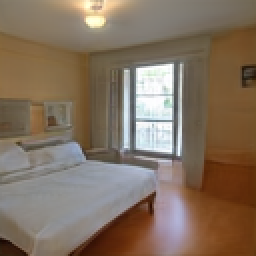} 
			\includegraphics[trim={ 0cm 0cm 0cm 0cm},scale=0.15]{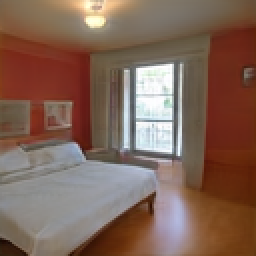}\\
		\end{tabular}
		\begin{tabular}{l}
			\footnotesize \hspace{1.5cm} Sunlight (12\_257)
			\\
			\rotatebox{90}{\phantom{K}}
			\includegraphics[trim={ 0cm 0cm 0cm 0cm},scale=0.15]{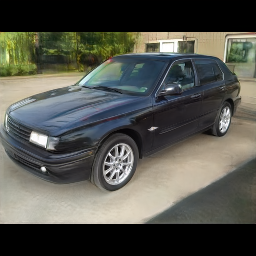} 
			\includegraphics[trim={ 0cm 0cm 0cm 0cm},scale=0.15]{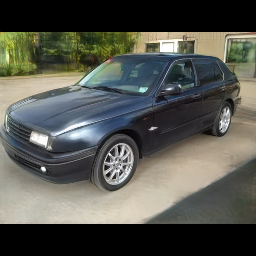} 
			\includegraphics[trim={ 0cm 0cm 0cm 0cm},scale=0.15]{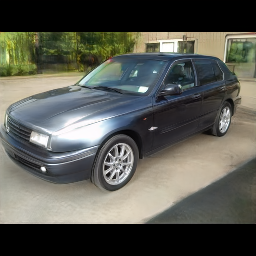} \\
			
			\rotatebox{90}{\footnotesize \hspace{-0.4cm} Headlight (8\_441) \hspace{-0.4cm}}
			\includegraphics[trim={ 0cm 0cm 0cm 0cm},scale=0.15]{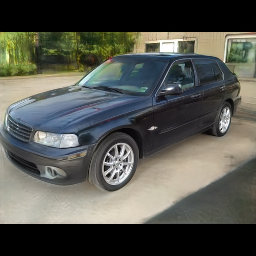} 
			\includegraphics[trim={ 0cm 0cm 0cm 0cm},scale=0.15]{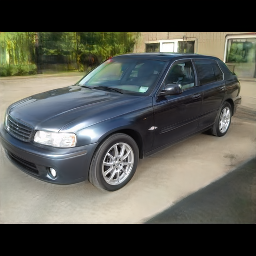} 
			\includegraphics[trim={ 0cm 0cm 0cm 0cm},scale=0.15]{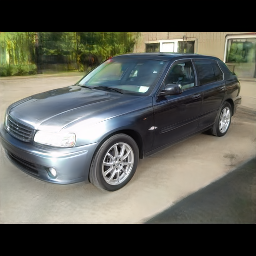}\\
			
			\rotatebox{90}{\phantom{K}}
			\includegraphics[trim={ 0cm 0cm 0cm 0cm},scale=0.15]{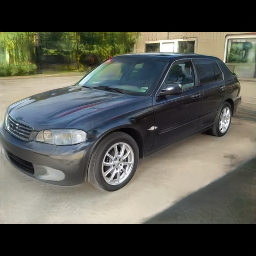} 
			\includegraphics[trim={ 0cm 0cm 0cm 0cm},scale=0.15]{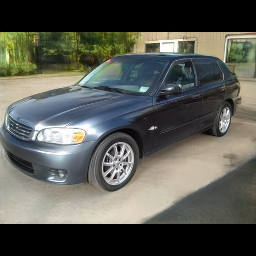} 
			\includegraphics[trim={ 0cm 0cm 0cm 0cm},scale=0.15]{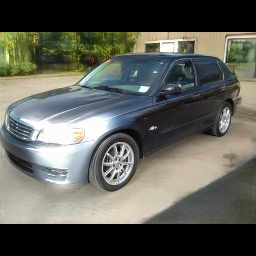}\\
		\end{tabular}

		\begin{tabular}{l}
			\footnotesize \hspace{1.5cm} Eye Size (12\_110)
			\\
			\rotatebox{90}{\phantom{K}}
			\includegraphics[trim={ 0cm 0cm 0cm 0cm},scale=0.15]{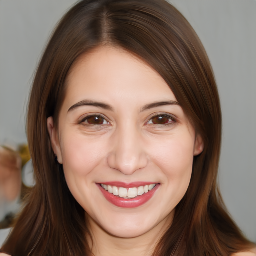} 
			\includegraphics[trim={ 0cm 0cm 0cm 0cm},scale=0.15]{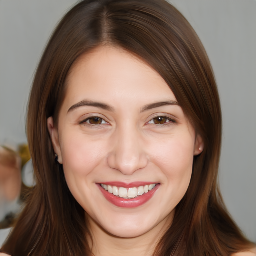} 
			\includegraphics[trim={ 0cm 0cm 0cm 0cm},scale=0.15]{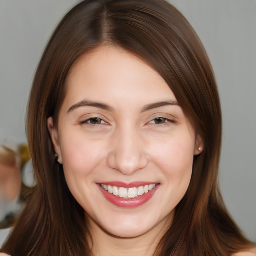} \\
			
			\rotatebox{90}{\footnotesize \hspace{-0.7cm}  Eye Lashes (12\_414) \hspace{-0.7cm}}
			\includegraphics[trim={ 0cm 0cm 0cm 0cm},scale=0.15]{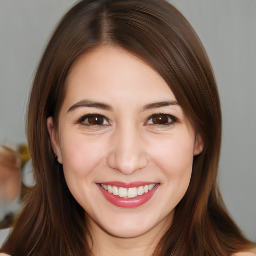} 
			\includegraphics[trim={ 0cm 0cm 0cm 0cm},scale=0.15]{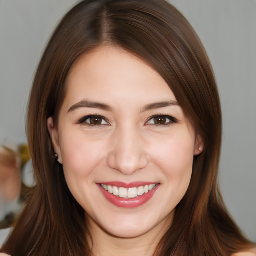} 
			\includegraphics[trim={ 0cm 0cm 0cm 0cm},scale=0.15]{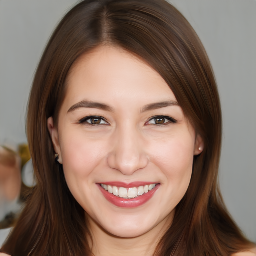}\\
			
			\rotatebox{90}{\phantom{K}}
			\includegraphics[trim={ 0cm 0cm 0cm 0cm},scale=0.15]{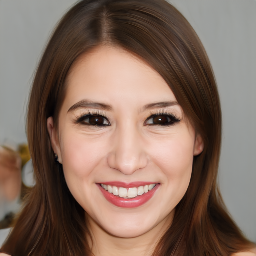} 
			\includegraphics[trim={ 0cm 0cm 0cm 0cm},scale=0.15]{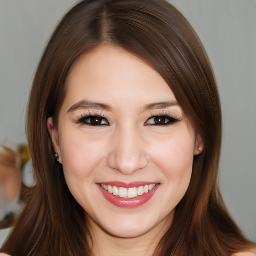} 
			\includegraphics[trim={ 0cm 0cm 0cm 0cm},scale=0.15]{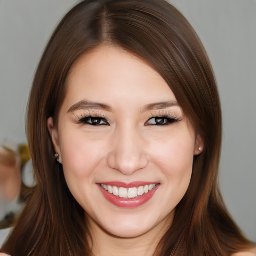}\\
		\end{tabular}
		\begin{tabular}{l}
			\footnotesize \hspace{1.5cm} Lamp on (11\_17)
			\\
			\rotatebox{90}{\phantom{K}}
			\includegraphics[trim={ 0cm 0cm 0cm 0cm},scale=0.15]{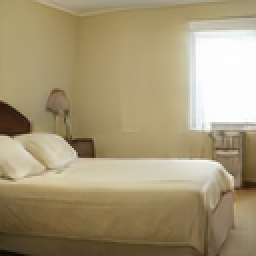} 
			\includegraphics[trim={ 0cm 0cm 0cm 0cm},scale=0.15]{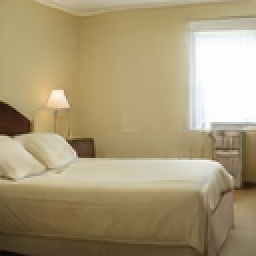} 
			\includegraphics[trim={ 0cm 0cm 0cm 0cm},scale=0.15]{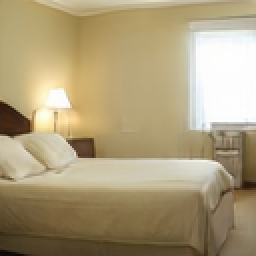} \\
			
			\rotatebox{90}{\footnotesize \hspace{-0.3cm}  Curtain (8\_297) \hspace{-0.3cm}}
			\includegraphics[trim={ 0cm 0cm 0cm 0cm},scale=0.15]{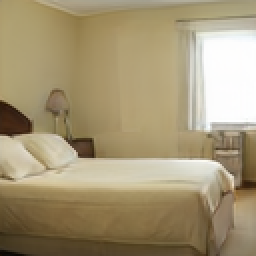} 
			\includegraphics[trim={ 0cm 0cm 0cm 0cm},scale=0.15]{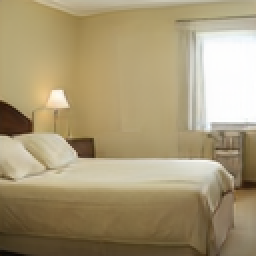} 
			\includegraphics[trim={ 0cm 0cm 0cm 0cm},scale=0.15]{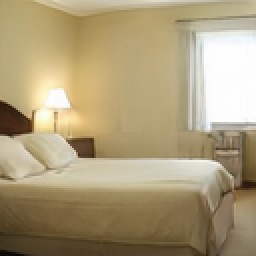}\\
			
			\rotatebox{90}{\phantom{K}}
			\includegraphics[trim={ 0cm 0cm 0cm 0cm},scale=0.15]{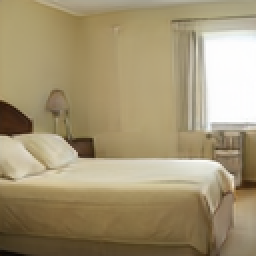} 
			\includegraphics[trim={ 0cm 0cm 0cm 0cm},scale=0.15]{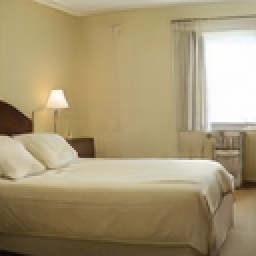} 
			\includegraphics[trim={ 0cm 0cm 0cm 0cm},scale=0.15]{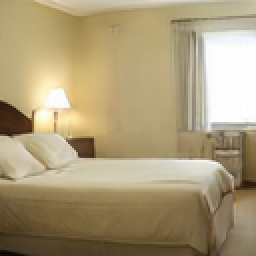}\\
		\end{tabular}
		\begin{tabular}{l}
			\footnotesize \hspace{1.5cm} Tree (9\_108)
			\\
			\rotatebox{90}{\phantom{K}}
			\includegraphics[trim={ 0cm 0cm 0cm 0cm},scale=0.15]{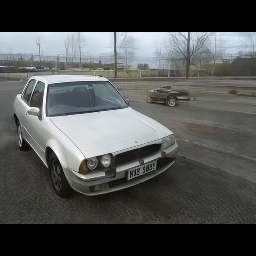} 
			\includegraphics[trim={ 0cm 0cm 0cm 0cm},scale=0.15]{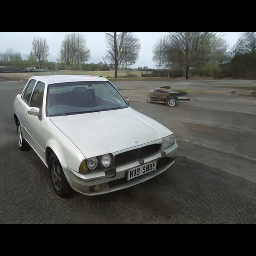} 
			\includegraphics[trim={ 0cm 0cm 0cm 0cm},scale=0.15]{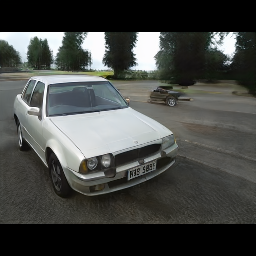} \\
			
			\rotatebox{90}{\footnotesize \hspace{-0.3cm} Grass (12\_107) \hspace{-0.4cm}}
			\includegraphics[trim={ 0cm 0cm 0cm 0cm},scale=0.15]{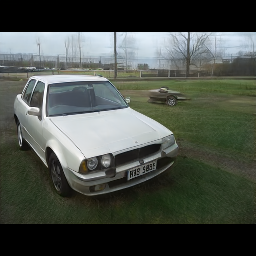} 
			\includegraphics[trim={ 0cm 0cm 0cm 0cm},scale=0.15]{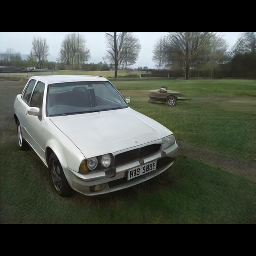} 
			\includegraphics[trim={ 0cm 0cm 0cm 0cm},scale=0.15]{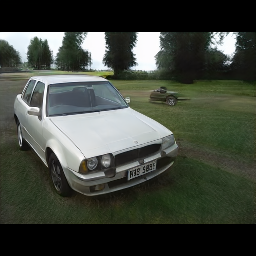}\\
			
			\rotatebox{90}{\phantom{K}}
			\includegraphics[trim={ 0cm 0cm 0cm 0cm},scale=0.15]{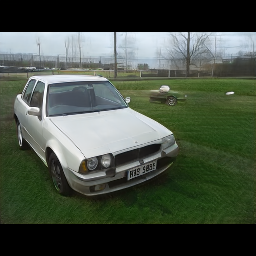} 
			\includegraphics[trim={ 0cm 0cm 0cm 0cm},scale=0.15]{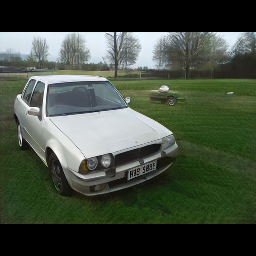} 
			\includegraphics[trim={ 0cm 0cm 0cm 0cm},scale=0.15]{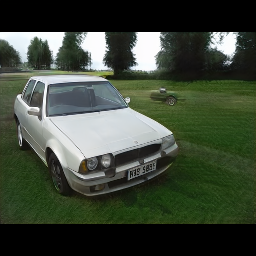}\\
		\end{tabular}
		
	\end{center}
	\caption{Disentanglement in style space, demonstrated using three different datasets (FFHQ, LSUN Bedroom, LSUN Car). Each of the six groups above shows two manipulations that occur independently inside the same image. The indices of the manipulated layer and channel are indicated in parentheses.}
	\label{fig:disentanglement2}
\end{figure}

\newpage
Finally, Figure~\ref{fig:disentanglement2} demonstrates (in addition to Figure~\ref{fig:disentanglement}) the high degree of disentanglement of the localized controls that our method detects. Even pairs of controls that affect the same semantic region, typically do so in an independent manner.

\newpage

\section{Attribute-specific channels} \label{ap:annotation_celebA}


Starting from the 40 attributes from CelebA \cite{karras2019style}, we first remove inactivated, ambiguous and neutral attributes. Inactivated attributes (defined in Section~\ref{sec:disent_level}) are those that are not well represented in the generated image distribution. Ambiguous attributes are highly subjective. Neutral attributes are those in between more extreme states, or attributes that are highly common across the dataset. For example, ``mouth slightly open'' is between an open mouth and a closed one. Another example of a neutral attribute is ``no beard'', since most of the faces in FFHQ don't have a beard.
The attributes that were found inactivated, ambiguous, or neutral, and were removed from further consideration are listed in Table~\ref{tab:remove_attributes}.


\begin{table}[h]
	\begin{center} 
		\begin{tabular}{llll} 
			\hline
		status	& type & \# attributes & list of attributes \\  
			\hline
			
			\multirow{4}{*}{removed} 
			&inactivated & 9 & \pbox{20cm}{\vspace{1mm}blurry, narrow eyes, necklace,\\ oval face, rosy cheeks, pointy noise, \\
				bald, mustache, pale skin} \\
			\cline{2-4}
			&ambiguous & 2 & \pbox{20cm}{\vspace{1mm}attractive, heavy make up $\quad\quad\quad\quad$}\\
			\cline{2-4}
			&neutral & 3 & \pbox{20cm}{\vspace{1mm}no beard, five-o-clock shadow,$\quad\quad$\\ mouth slightly open} \vspace{1mm} \\ 
			\hline
            \multirow{6}{*}{annotated}
			& \pbox{5cm}{one or more\\disentangled\\single-channel\\controls found}  & 16 & \pbox{20cm}{\vspace{1mm}gender, smiling, lipstick, \\ eyeglasses, bangs, wavy hair,\\ earrings, black hair, blond hair,\\ sideburns, goatee, receding hairline, \\ gray hair, suit (tie), double chin, hat } \\
			\cline{2-4}
			& \pbox{5cm}{no disentangled\\ single-channel\\controls found} & 10 & \pbox{20cm}{\vspace{1mm}bags under eyes, big nose, \\high cheekbones,\\ young, arched eyebrows, brown hair, \\ big lips, bushy eyebrows,\\ chubby, straight hair}\\
			\hline
		\end{tabular}
	\end{center}
	\caption{For 40 CelebA attributes, we first remove inactivated (9), ambiguous (2) or neutral (3) attributes. Our method is able to detect one or more disentangled single-channel controls for 16 out of the 26 remaining attributes.}
	\label{tab:remove_attributes}
\end{table}

\begin{table}[h]
\centering
 \begin{tabular}{llll} 
 \hline
region & attribute & (layer,channel,rank) & related attributes\\  
 \hline
\multirow{21}{*}{hair} 
& black hair & (12,479,1) & different hair color, lighting \\ \cline{2-4}
& blond hair  & \pbox{20cm}{(12,479,1) \\(12,266,3)} & gender, other hair color and style \\ \cline{2-4}
& gray hair   & \pbox{20cm}{(11,286,1)} & glasses, gender, wrinkle and beard \\ \cline{2-4}
& wavy hair   & \pbox{20cm}{(6,500,1)\\(8,128,2)\\(5,92,3)\\(6,394,7)\\(6,323,28)} & hair style, gender \\ \cline{2-4}
& bangs   & \pbox{20cm}{(3,259,1)\\(6,285,2)\\(5,414,3)\\(6,128,4)\\(9,295,8)\\(6,322,9)\\(6,487,11)\\(6,504,14)} & hair style \\ \cline{2-4}
& receding hairline   & \pbox{20cm}{(5,414,1)\\(6,322,2)\\(6,497,3)\\(6,504,8)} & hair style \\ \hline
\multirow{2}{*}{mouth} 
& smiling   & \pbox{20cm}{(6,501,1)} & size of face or eye \\ \cline{2-4}
& lipstick   & \pbox{20cm}{(15,45,1)} & gender, face expression \\ \hline
\multirow{2}{*}{beard} 
& sideburns   & \pbox{20cm}{(12,237,2)} & other type of beard, gender \\ \cline{2-4}
& goatee   & \pbox{20cm}{(9,421,1)} & other type of beard, gender \\ \hline
\multirow{1}{*}{chin} 
& double chin   & \pbox{20cm}{(9,132,1)} & size of neck, wrinkle \\ \hline
\multirow{1}{*}{ear} 
& \pbox{20cm}{earrings\\(entangled with gender)}   & \pbox{20cm}{(8,81,1) } & gender, face shape \\ \hline
\multirow{1}{*}{eye} 
& glasses   & \pbox{20cm}{(3,288,1)\\(2,175,3)\\(3,120,4)\\(2,97,6)} & gender, wrinkle and beard \\ \hline
\multirow{1}{*}{clothes} 
& suit (tie)   & \pbox{20cm}{(9,441,1)\\(8,292,2)\\(11,358,3)\\(6,223,11)} & cloth style \\ \hline
\multirow{1}{*}{hat} 
& hat size   & \pbox{20cm}{(5,200,7)} & nothing change \\ \hline
\multirow{1}{*}{overall} 
& gender   & \pbox{20cm}{(9,6,1)} & beard,hair style \\ \hline
\end{tabular}
\vspace{3mm}
\caption{
	List of attributes and the single-channel controls that were detected for them. The indices of layers and channels start from 0, while ranks start from 1.
}
\label{tab:attribute}
\end{table}

To find attribute-specific controls we apply our method described in Section 5 on the remaining 26 attributes. We found that 16 out of the 26 remaining attributes are controllable by one or more single style channels, in a disentangled manner. Our method was not able to identify any \emph{disentangled} single-channel controls for the other 10 attributes. All of the above attributes are listed in Table~\ref{tab:remove_attributes}. The attributes for which no disentangled single-channel controls were found indeed appear to be correlated with other visual attributes (in FFHQ). For example, ``bags under eyes'' is correlated with eye size, ``big lips'' is correlated with skin color, and ``high cheekbones'' is correlated with smiling. 

While our method could not find a disentangled single-channel control for the ``young'' attribute, it was able to find such controls for wrinkles, eyeglasses, and gray hair. Because all of these attributes are correlated with age, the ``young'' attribute can only be controlled by manipulating multiple style channels, rather than a single one. 



Note that although the attribute-specific detection method of Section~\ref{sec:attributes} could not detect a single-channel control for either ``arched eyebrows'' or ``bushy eyebrows'', our locally-active detection method in Section~\ref{sec:localized} was able to find disentangled controls for these attributes: (9,30) for arched eyebrows, and (12,325) for bushy eyebrows. Thus, these could be considered as failure cases for our attribute-specific detection method. 

Table~\ref{tab:attribute} lists the various attributes and the single-channel controls that were detected for them. For each control we list the layer and channel number, as well as its rank by the detection method of Section~\ref{sec:attributes}.

\newpage
\phantom{kk}
\newpage
\section{Attribute Dependency} \label{sec:insightAD}

To compare the disentanglement of different image manipulation methods, we propose a general disentanglement metric for real images, which we refer to as \textit{Attribute Dependency} (AD). Attribute Dependency measures the degree to which manipulation along a certain direction induces changes in other attributes, as measured by classifiers for those attributes. Below we share our insights regarding AD, and its implementation details.
Next, we use AD to show our image manipulation method is more disentangled that two other methods (GANSpace \cite{harkonen2020ganspace}, InterfaceGan \cite{shen2020interfacegan}) in Figure~\ref{fig:single-vs-all2} and Figure~\ref{fig:maxAD}. Additionally, we further show in Figure~\ref{fig:identity} that our method changes face identity less than GANSpace and InterFaceGAN.

For a given target attribute $t$, we measure AD as follows. First, we sample a set of images without the target attribute $t$ (e.g., without gray hair), and manipulate them towards the target attribute, by a certain amount measured by the change in the logit outcome $\Delta l_t$ of a classifier pretrained to detect attribute $t$.
Next, we measure the change of logit $\Delta l_i$ between the original images and the manipulated ones for other attributes $\forall i \in \mathcal{A}\backslash{t}$, where $\mathcal{A}$ is the set of all attributes. Each change is normalized by $\sigma(l_i)$, the standard deviation of the logit value for attribute $i$ over a large set of generated images.
We measure mean-AD, defined as $E({1 \over k}\sum_{i \in \mathcal{A}\backslash{t}} (\frac{\Delta l_i}{\sigma(l_i)}))$, where $k = |\mathcal{A}| - 1$.
Similarly, we measure max-AD, defined as $E(\max_{i \in \mathcal{A}\backslash{t}} (\frac{\Delta l_i}{\sigma(l_i)}))$.

\subsection{Insights}

\begin{figure*}[t]
\begin{center}
\includegraphics[trim={ 0cm 0cm 0cm 0cm},scale=0.33]{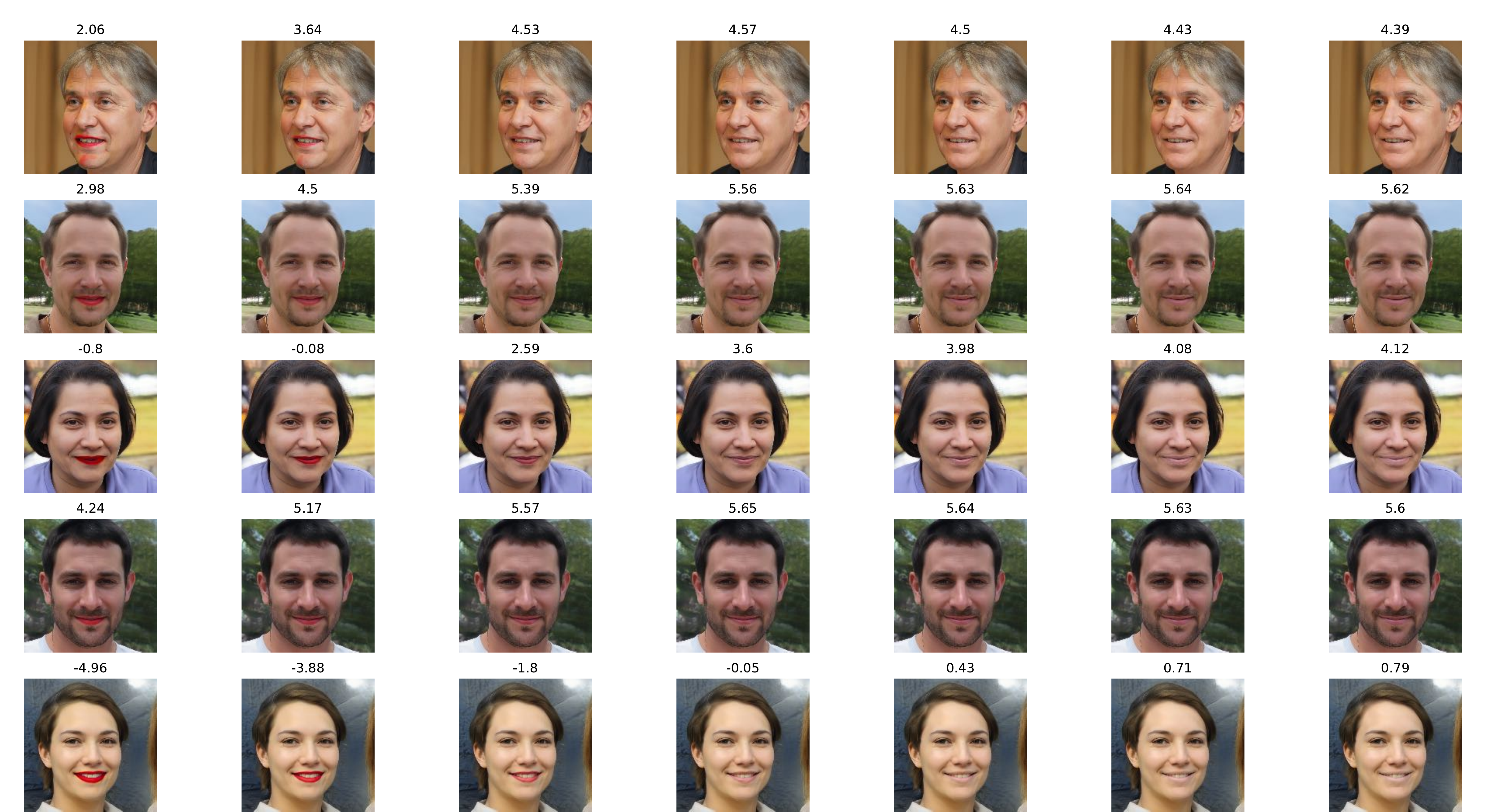}
\end{center}
   \caption{Logits of lipstick classifier. The strength of attribute is reduced from left to right. The classifier logit is on top of each image, increasing from left to right. Note that the logit sign is not aligned with the presence of the attribute: there are images with strong lipstick, but a positive logit. \vspace{3mm}}
\label{fig:logit_lipstick}
\end{figure*}
\begin{figure*}[tb]
	\begin{center}
		\includegraphics[trim={ 0cm 0cm 0cm 0cm},scale=0.33]{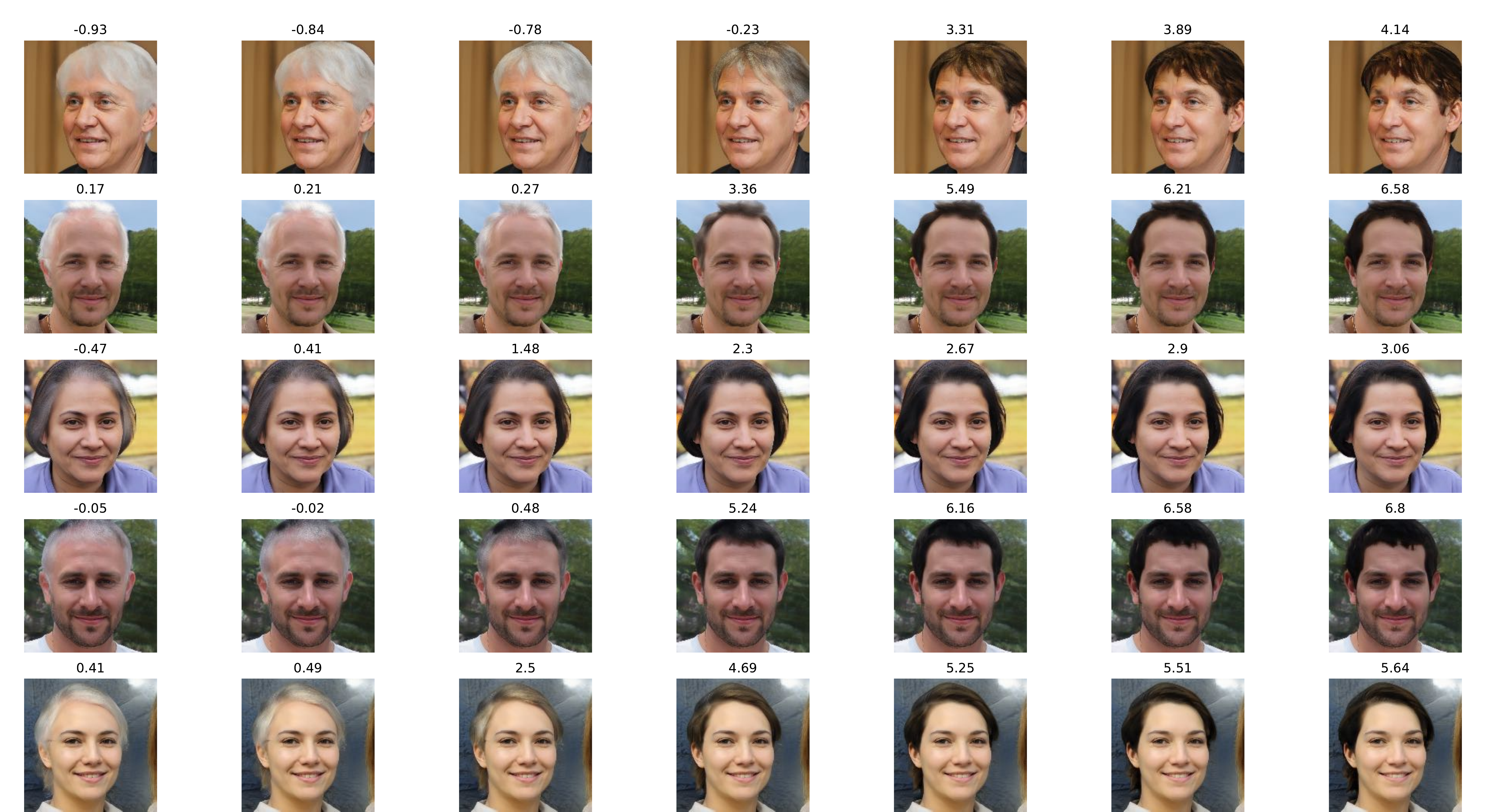}
	\end{center}
	\caption{Logits of hair greyness classifier. The strength of attribute is reduced from left to right. The classifier logit is on top of each image, increasing from left to right. Note that the logit sign is not aligned with the presence of the attribute: there are images with strong hair greyness, but a positive logit. }
	\label{fig:logit_grey}
\end{figure*}

To measure how much a specific attribute $i \in \mathcal{A}\backslash{t}$ has changed, we use a pretrained classifier for that attribute. Under normal operating mode, a binary classifier outputs a logit $l_i \in [-\infty,+\infty]$, which is then converted to a probability value in $[0,1]$, with positive logit values yielding probabilities higher than 0.5, and negative logit values yielding probabilities lower than 0.5.
However, classifiers trained on real data may be affected by entanglement present in the training data, and they are often unable to detect the presence or absence of an attribute in a disentangled manner. For example, a female face with lipstick will typically cause the classifier to output a negative logit value (indicating the presence of a lipstick), but the classifier might output a positive logit value given a face of a male with lipstick. Similarly, a gray hair classifier will output a negative logit value for a male with gray hair, but might output a positive logit value for a female with gray hair. This is demonstrated in Figures \ref{fig:logit_lipstick} and \ref{fig:logit_grey}.



Thus, when attempting to measure the magnitude of change of an attribute, we choose not to consider the classifier's logit sign or value; rather, we find that the change in the logit value, $\Delta l$, appears to be better correlated with an image space change in the attribute. We use the change in the logit, rather than the change in the probability because of the saturating effect of the sigmoid non-linearity that is used to convert logits to probabilities. For example, the probability produced by a lipstick classifier for a female wearing a lighter lipstick and a stronger lipstick is going to be nearly the same, while this is not the case for the logit values (see the last row of Figure \ref{fig:logit_lipstick}). 


Another insight is that if the manipulation strength is too high, the generated images will be unrealistic, and classifiers will give unexpected predictions. It is crucial not to use too high manipulation strengths to make sure the logits are meaningful. If the generated images are realistic, the logit is nearly a monotonic function of the strength of target attribute. And we consider the manipulation strength is a monotonic function of strength of target attribute. Therefore, we can control the amount of manipulation (measured by $\Delta l$) by searching for the corresponding manipulation strength through bisection method.

Our final insight is that the classifiers are not immune to noise. When provided with the same images with only slight texture changes in hair and background, the classifiers are supposed to output the same logits. In practice, however, the logits are slightly different. Thus, when comparing the effect of different manipulation methods on various attributes, it is necessary to ensure that the differences in the measurements are caused by the inherent differences between the methods, rather than by noise in the classifier outputs.

\subsection{Implementation}
We randomly generate 500K images, as our image bank, and annotate each image with 31 active attributes, same as was done in Section~\ref{sec:disent_level}, where a negative logit corresponds to presence of the target attribute in an image. Let $\sigma(l)$ denote the standard deviation of logits over the entire image bank. For each target attribute (for example, gray hair), we rank its logit from negative to positive, and take images with 50-75\% quantile as manipulation candidates, since they exhibit little to mild presence of the target attribute (without much gray hair). We don't take images with the most positive logit (75-100\% quantile) since they are less likely to result in a realistic manipulation. Candidates are manipulated toward strong attribute presence (adding gray hair, more negative logit). We set the $\Delta l_t= r\sigma(l_t)$, where $r\in \{0.25, 0.5, 0.75, 1\}$. $r$ should not be too large to make sure that most of the manipulated images are still realistic. Then we use the bisection method to find the manipulation strength $m$ that can generate an image with final logit $|(l_t^{final}-(l_t^{initial} -\Delta l_t))|< r_{tolerance} \sigma(l_t)$ with $r_{tolerance}=5\%$, and ignore images that don't converge after 20 iterations. 
We manually set the maximum manipulation strength $m_{max}$ such that almost all manipulated images with manipulation strength $m_{max}$ strongly exhibit the target attribute, but still look realistic. $m_{max}$ is used to initiate the bisection method.

Next, we add a control group with $r=0$ to the experiment, such that $\Delta l_t = 0\sigma(l_t)$. This group is used to represent the inherent noise of classifiers. The input images are copies of the original ones, obtained by keeping the latent code $s$ unchanged, but changing the noise inputs at different layers. They are essentially identical to the original images, with subtle differences in hair, skin, and background.

\begin{figure}[tb]
	\centering
	\begin{tabular}{cccc}
		{\phantom{k} Original} {\phantom{kk} GANSpace \phantom{}} {\phantom{} InterfaceGAN} {\phantom{kk} Ours \phantom{k}}\\
		
		\rotatebox{90}{\phantom{kk}Gender}
		\includegraphics[trim={ 0cm 0cm 0cm 0cm},scale=0.2]{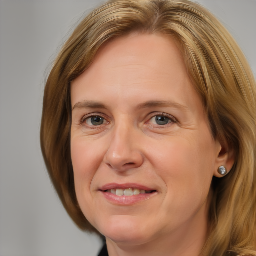} 
		\includegraphics[trim={ 0cm 0cm 0cm 0cm},scale=0.2]{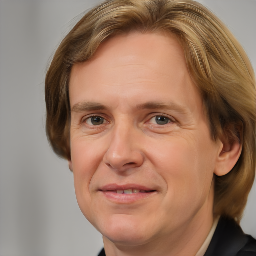} 
		\includegraphics[trim={ 0cm 0cm 0cm 0cm},scale=0.2]{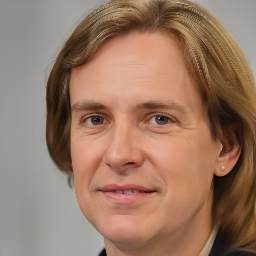} 
		\includegraphics[trim={ 0cm 0cm 0cm 0cm},scale=0.2]{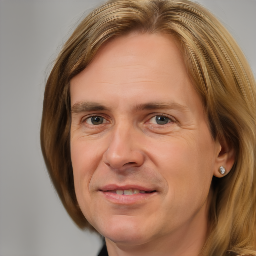} 
		\\
		
		\rotatebox{90}{\phantom{k}Gray hair}
		\includegraphics[trim={ 0cm 0cm 0cm 0cm},scale=0.2]{./fig5/original.png} 
		\includegraphics[trim={ 0cm 0cm 0cm 0cm},scale=0.2]{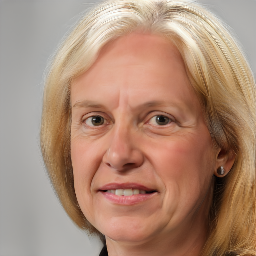} 
		\includegraphics[trim={ 0cm 0cm 0cm 0cm},scale=0.2]{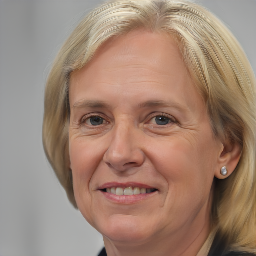} 
		\includegraphics[trim={ 0cm 0cm 0cm 0cm},scale=0.2]{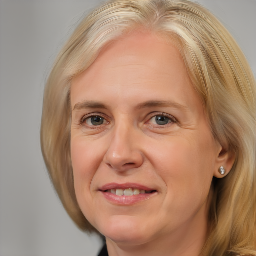} 
		\\
		
		\rotatebox{90}{\phantom{k} Lipstick}
		\includegraphics[trim={ 0cm 0cm 0cm 0cm},scale=0.2]{./fig5/original.png} 
		\includegraphics[trim={ 0cm 0cm 0cm 0cm},scale=0.2]{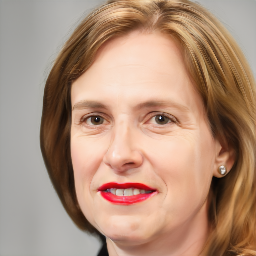} 
		\includegraphics[trim={ 0cm 0cm 0cm 0cm},scale=0.2]{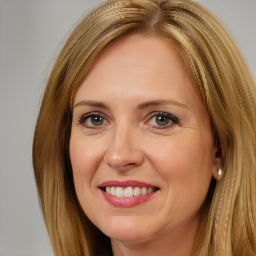} 
		\includegraphics[trim={ 0cm 0cm 0cm 0cm},scale=0.2]{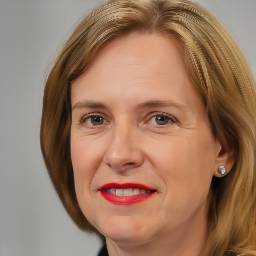} 
		\\	
	\end{tabular}
	
	\caption{Comparison with state-of-the-art methods with amount of manipulation $\Delta l_t = 1.5\sigma(l_t)$. We deliberately choose a strong manipulation (1.5 instead of 1) to emphasize the differences.
	}
	\label{fig:single-vs-all2}
\end{figure}
\begin{figure}[t]
	\begin{center}
		\includegraphics[trim={ 0cm 0cm 0cm 0cm},scale=0.335]{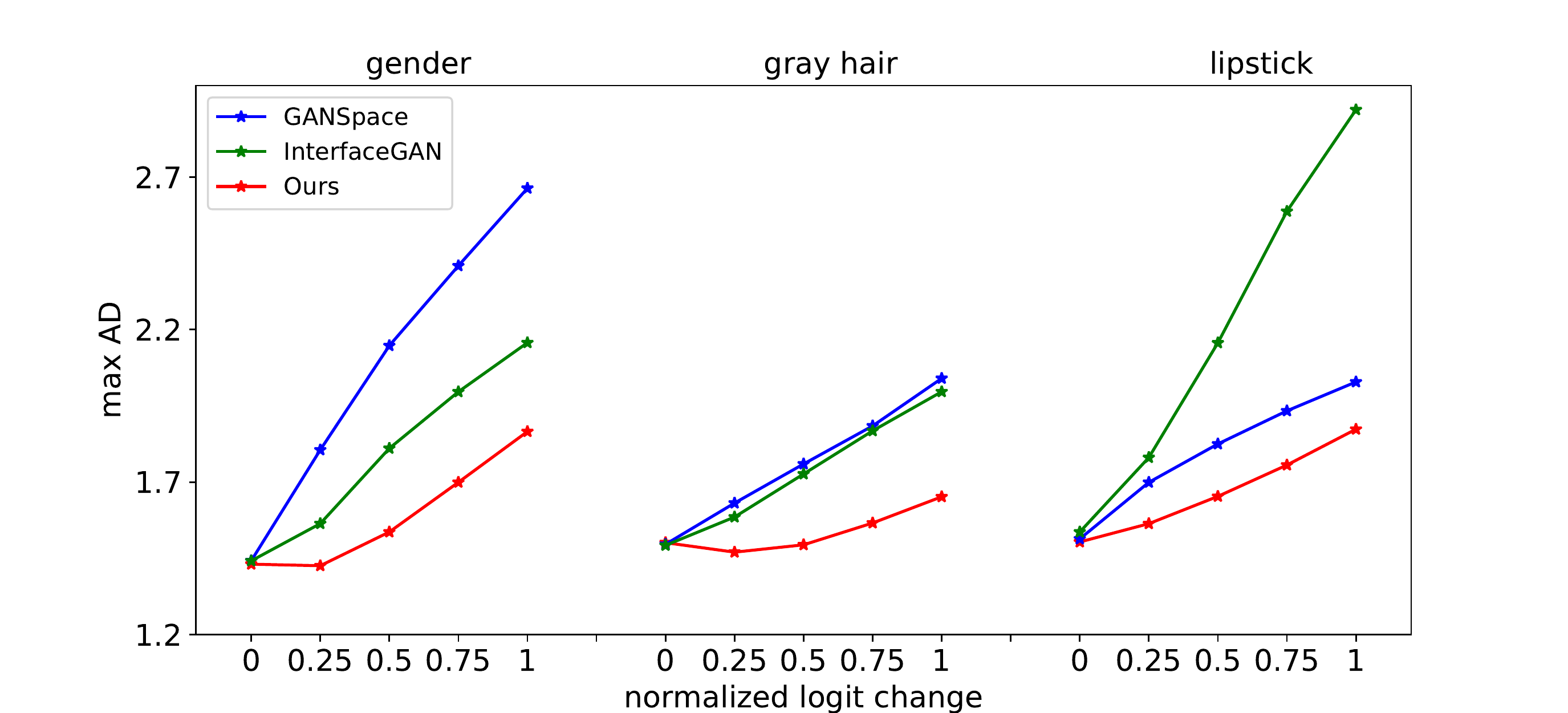} 
	\end{center}
	\caption{Max-AD vs.~the degree of target attribute manipulation ($\Delta l_t/\sigma(l_t)$). Lower max-AD indicates better disentanglement.}
	\label{fig:maxAD}
\end{figure}

Finally, we use the same 3K images with identifiable $m$ for all candidate manipulation methods to calculate mean-AD and max-AD. The mean-AD for the three methods (GANSpace, InterFaceGAN, and ours) for three attributes (gender, gray hair, and lipstick) are plotted in Figure~\ref{fig:meanAD}, and the max-AD in Figure~\ref{fig:maxAD}. Figures~\ref{fig:single-vs-all} and~\ref{fig:single-vs-all2} show a qualitative comparison between the manipulations produced by the three methods. Note that, like in Figure~\ref{fig:single-vs-all}, the \emph{Lipstick} manipulation by InterFaceGAN significantly changes the identity of the person, and the \emph{Gray hair} manipulation adds wrinkles. GANSpace manipulations also exhibit some entanglement (\textit{Lipstick} affects face lightness, \textit{Gray hair} ages the rest of the face). In contrast, our approach appears to affect only the target attribute. Our \textit{Gender} manipulation, for example, does not affect the hair style, and minimally changes the face, yet the gender unmistakably changes.

\subsection{Identity change}

\begin{figure}[t]
	\begin{center}
		\includegraphics[trim={ 0cm 0cm 0cm 0cm},scale=0.335]{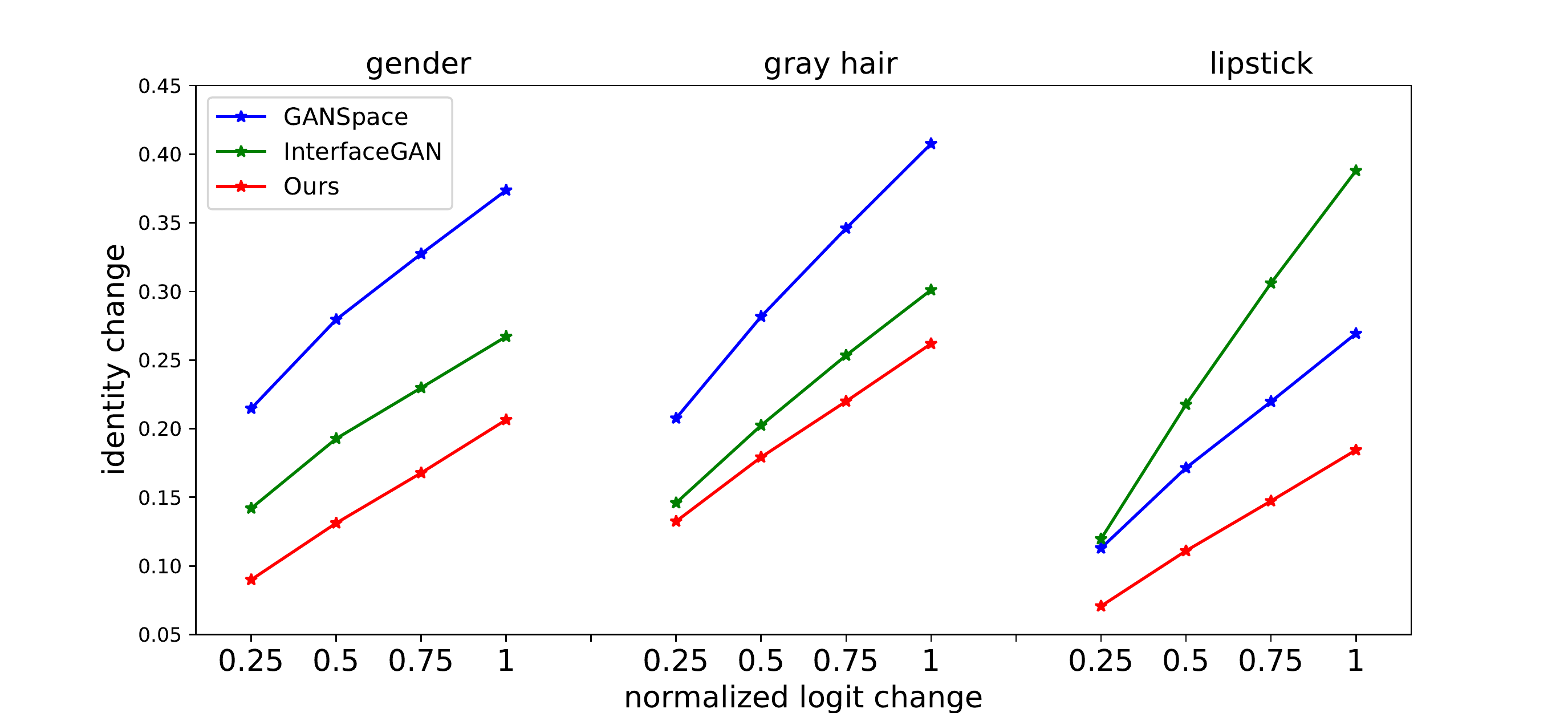} 
	\end{center}
	\caption{ Identity change vs.~the degree of target attribute manipulation ($\Delta l_t/\sigma(l_t)$). Lower identity changes indicates that the manipulation is better disentanglement from the identity.}
	\label{fig:identity}
\end{figure}

In addition to AD, we use another metric (identity change) to compare our method against GANSpace and InterFaceGAN. Specifically, we use FaceNet~\cite{schroff2015facenet}, which is a standard network for measuring identity change. We use the official implementation, which first detects faces, then crops faces out, obtains an embedding from the last layer, and calculates the Euclidean norm between the embedding of the original images and that of the manipulated ones. As shown in Figure~\ref{fig:identity}, manipulations done with our method change the identity less than manipulations of GANSpace or InterFaceGAN.

\newpage

\section{Manipulation of real images}

\begin{figure}[h]
\setlength{\tabcolsep}{1.8pt}
\centering
\begin{tabular}{cccc}

Original & $\mathcal{W}$ & $\mathcal{W+}$ & $\mathcal{S}$
\\
\includegraphics[trim={ 0cm 0cm 0cm 0cm},scale=0.2]{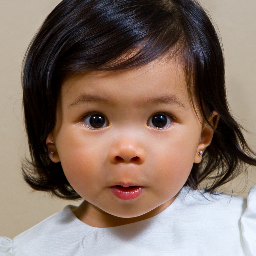} &
\includegraphics[trim={ 0cm 0cm 0cm 0cm},scale=0.2]{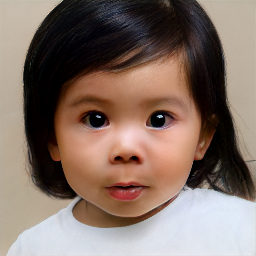} &
\includegraphics[trim={ 0cm 0cm 0cm 0cm},scale=0.2]{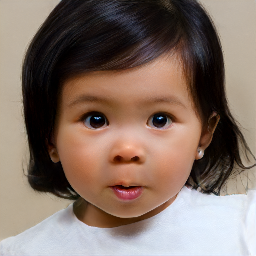} &
\includegraphics[trim={ 0cm 0cm 0cm 0cm},scale=0.2]{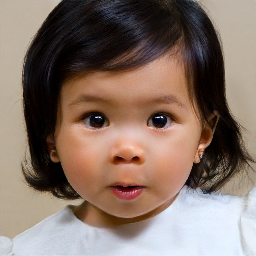} 
\\
\includegraphics[trim={ 0cm 0cm 0cm 0cm},scale=0.2]{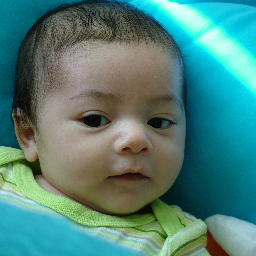} &
\includegraphics[trim={ 0cm 0cm 0cm 0cm},scale=0.2]{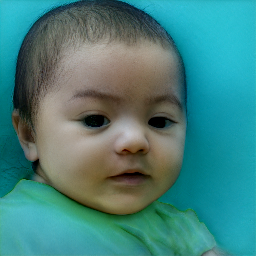} &
\includegraphics[trim={ 0cm 0cm 0cm 0cm},scale=0.2]{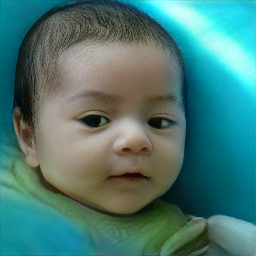} &
\includegraphics[trim={ 0cm 0cm 0cm 0cm},scale=0.2]{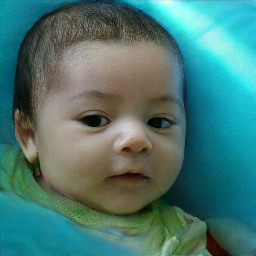} 
\\
\includegraphics[trim={ 0cm 0cm 0cm 0cm},scale=0.2]{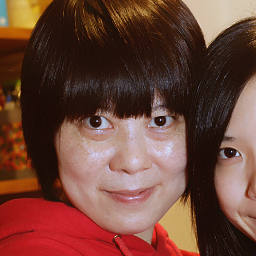} &
\includegraphics[trim={ 0cm 0cm 0cm 0cm},scale=0.2]{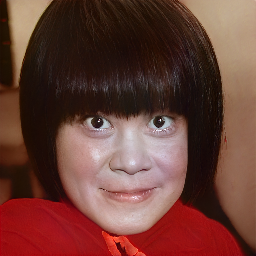} &
\includegraphics[trim={ 0cm 0cm 0cm 0cm},scale=0.2]{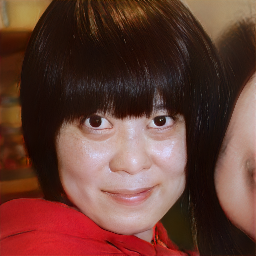} &
\includegraphics[trim={ 0cm 0cm 0cm 0cm},scale=0.2]{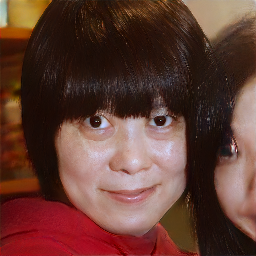} 
\\
\includegraphics[trim={ 0cm 0cm 0cm 0cm},scale=0.2]{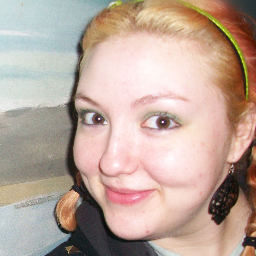} &
\includegraphics[trim={ 0cm 0cm 0cm 0cm},scale=0.2]{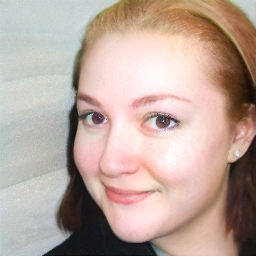} &
\includegraphics[trim={ 0cm 0cm 0cm 0cm},scale=0.2]{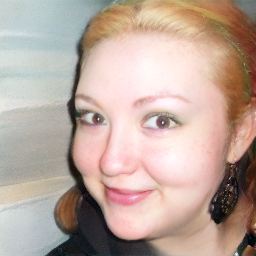} &
\includegraphics[trim={ 0cm 0cm 0cm 0cm},scale=0.2]{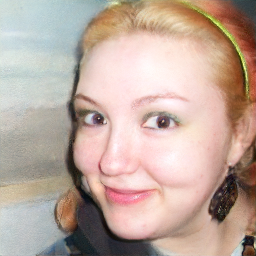} 
\\
\includegraphics[trim={ 0cm 0cm 0cm 0cm},scale=0.2]{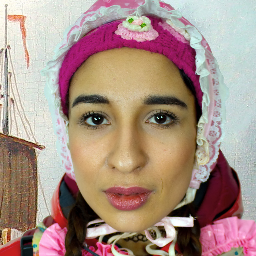} &
\includegraphics[trim={ 0cm 0cm 0cm 0cm},scale=0.2]{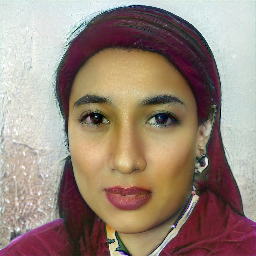} &
\includegraphics[trim={ 0cm 0cm 0cm 0cm},scale=0.2]{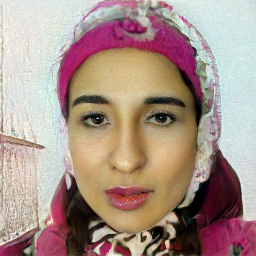} &
\includegraphics[trim={ 0cm 0cm 0cm 0cm},scale=0.2]{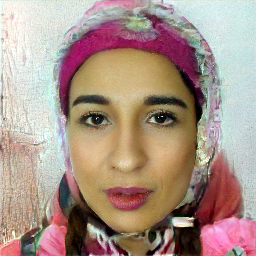} 
\\
\includegraphics[trim={ 0cm 0cm 0cm 0cm},scale=0.2]{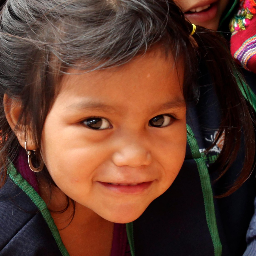} &
\includegraphics[trim={ 0cm 0cm 0cm 0cm},scale=0.2]{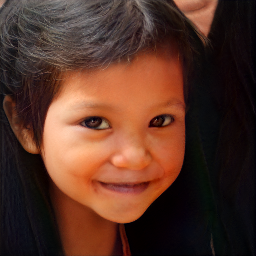} &
\includegraphics[trim={ 0cm 0cm 0cm 0cm},scale=0.2]{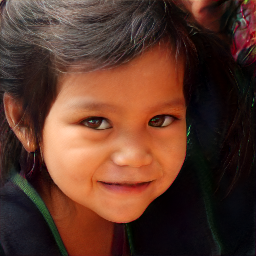} &
\includegraphics[trim={ 0cm 0cm 0cm 0cm},scale=0.2]{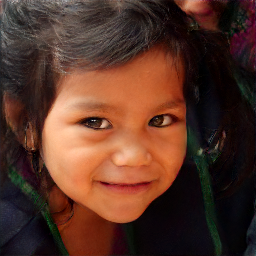} 
\\
\includegraphics[trim={ 0cm 0cm 0cm 0cm},scale=0.2]{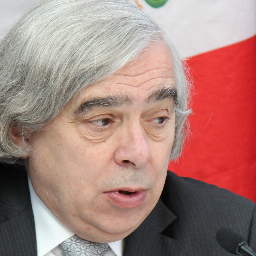} &
\includegraphics[trim={ 0cm 0cm 0cm 0cm},scale=0.2]{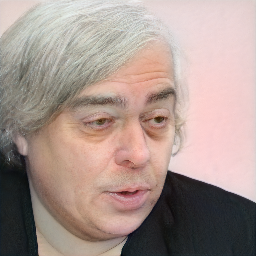} &
\includegraphics[trim={ 0cm 0cm 0cm 0cm},scale=0.2]{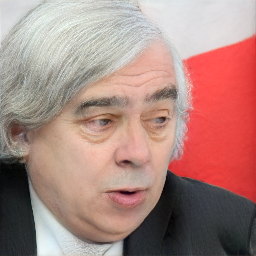} &
\includegraphics[trim={ 0cm 0cm 0cm 0cm},scale=0.2]{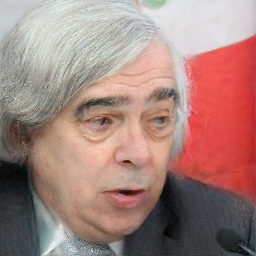} 
\\
\includegraphics[trim={ 0cm 0cm 0cm 0cm},scale=0.2]{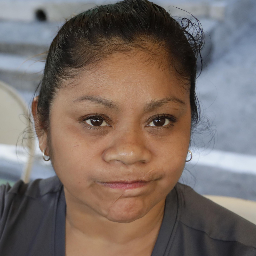} &
\includegraphics[trim={ 0cm 0cm 0cm 0cm},scale=0.2]{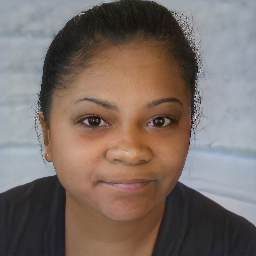} &
\includegraphics[trim={ 0cm 0cm 0cm 0cm},scale=0.2]{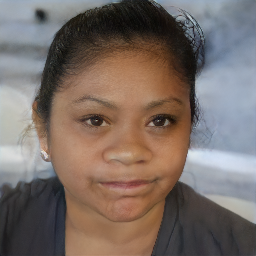} &
\includegraphics[trim={ 0cm 0cm 0cm 0cm},scale=0.2]{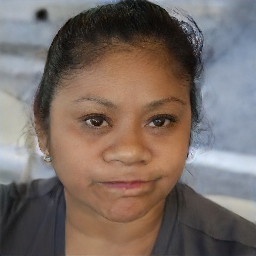} 
\\
\includegraphics[trim={ 0cm 0cm 0cm 0cm},scale=0.2]{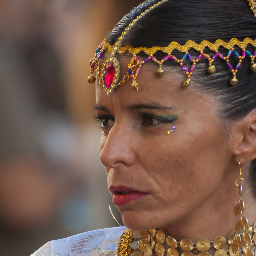} &
\includegraphics[trim={ 0cm 0cm 0cm 0cm},scale=0.2]{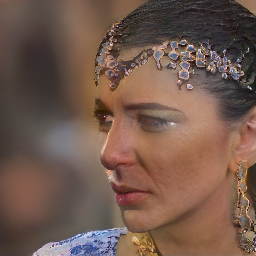} &
\includegraphics[trim={ 0cm 0cm 0cm 0cm},scale=0.2]{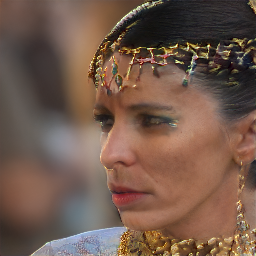} &
\includegraphics[trim={ 0cm 0cm 0cm 0cm},scale=0.2]{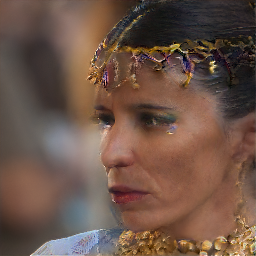} 
\\
\includegraphics[trim={ 0cm 0cm 0cm 0cm},scale=0.2]{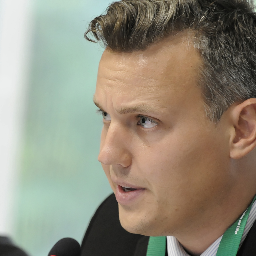} &
\includegraphics[trim={ 0cm 0cm 0cm 0cm},scale=0.2]{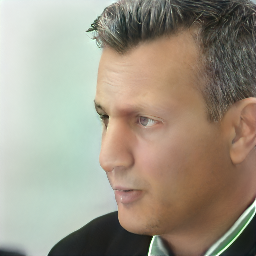} &
\includegraphics[trim={ 0cm 0cm 0cm 0cm},scale=0.2]{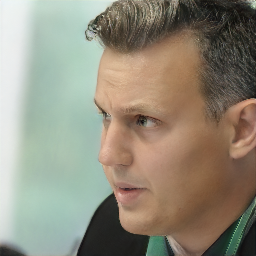} &
\includegraphics[trim={ 0cm 0cm 0cm 0cm},scale=0.2]{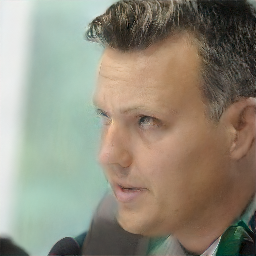} 
\\
\end{tabular}

   \caption{Inversion via latent optimization in $\mathcal{W}$, $\mathcal{W+}$, $\mathcal{S}$. It may be easily seen that the reconstruction is least accurate for $\mathcal{W}$, more accurate for $\mathcal{W+}$, and is best for $\mathcal{S}$}
\label{fig:invert}
\end{figure}

\begin{figure}[h]
\setlength{\tabcolsep}{1.8pt}
\centering
\begin{tabular}{cccc}
Original & $\mathcal{W}$ & $\mathcal{W+}$ & $\mathcal{S}$
\\
\rotatebox{90}{\phantom{kk}Smile}
\includegraphics[trim={ 0cm 0cm 0cm 0cm},scale=0.2]{./invert/0_original.png} &
\includegraphics[trim={ 0cm 0cm 0cm 0cm},scale=0.2]{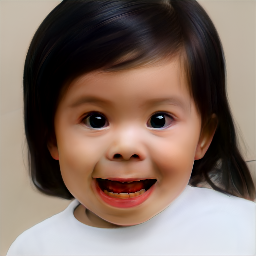} &
\includegraphics[trim={ 0cm 0cm 0cm 0cm},scale=0.2]{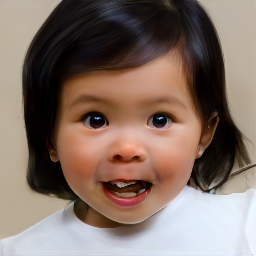} &
\includegraphics[trim={ 0cm 0cm 0cm 0cm},scale=0.2]{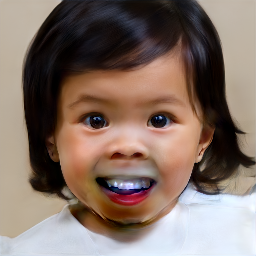} 
\\
\rotatebox{90}{\phantom{kk}Smile}
\includegraphics[trim={ 0cm 0cm 0cm 0cm},scale=0.2]{./invert/2_original.png} &
\includegraphics[trim={ 0cm 0cm 0cm 0cm},scale=0.2]{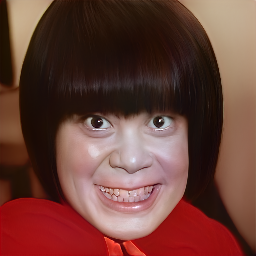} &
\includegraphics[trim={ 0cm 0cm 0cm 0cm},scale=0.2]{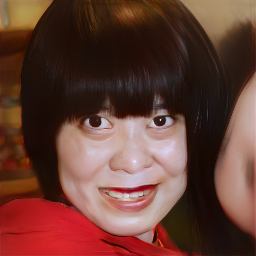} &
\includegraphics[trim={ 0cm 0cm 0cm 0cm},scale=0.2]{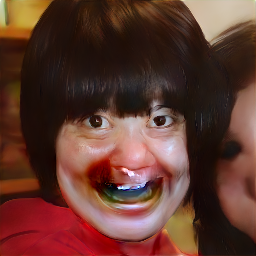} 
\\
\rotatebox{90}{\phantom{kk}Smile}
\includegraphics[trim={ 0cm 0cm 0cm 0cm},scale=0.2]{./invert/3_original.png} &
\includegraphics[trim={ 0cm 0cm 0cm 0cm},scale=0.2]{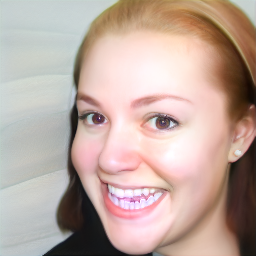} &
\includegraphics[trim={ 0cm 0cm 0cm 0cm},scale=0.2]{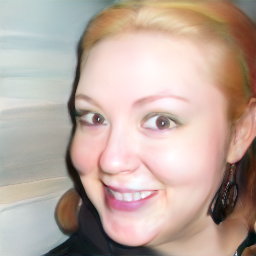} &
\includegraphics[trim={ 0cm 0cm 0cm 0cm},scale=0.2]{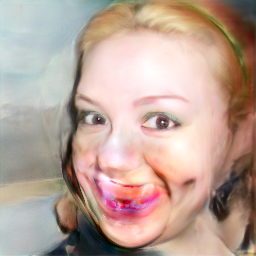} 
\vspace{2mm}
\\

\rotatebox{90}{\phantom{kk}Gaze}
\includegraphics[trim={ 0cm 0cm 0cm 0cm},scale=0.2]{./invert/0_original.png} &
\includegraphics[trim={ 0cm 0cm 0cm 0cm},scale=0.2]{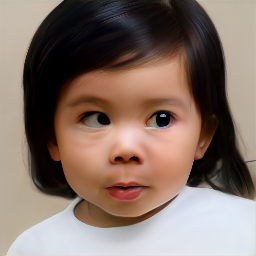} &
\includegraphics[trim={ 0cm 0cm 0cm 0cm},scale=0.2]{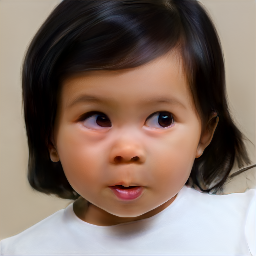} &
\includegraphics[trim={ 0cm 0cm 0cm 0cm},scale=0.2]{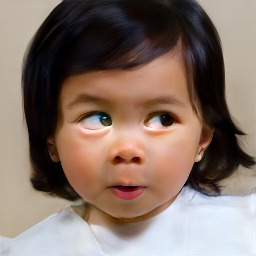} 
\\
\rotatebox{90}{\phantom{kk}Gaze}
\includegraphics[trim={ 0cm 0cm 0cm 0cm},scale=0.2]{./invert/2_original.png} &
\includegraphics[trim={ 0cm 0cm 0cm 0cm},scale=0.2]{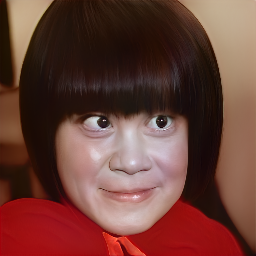} &
\includegraphics[trim={ 0cm 0cm 0cm 0cm},scale=0.2]{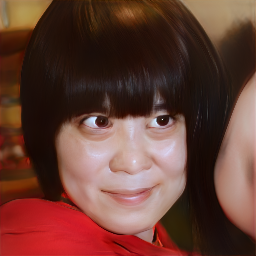} &
\includegraphics[trim={ 0cm 0cm 0cm 0cm},scale=0.2]{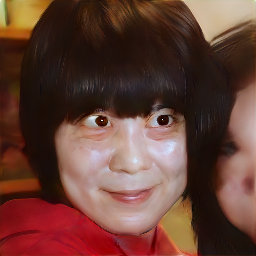} 
\\
\rotatebox{90}{\phantom{kk}Gaze}
\includegraphics[trim={ 0cm 0cm 0cm 0cm},scale=0.2]{./invert/3_original.png} &
\includegraphics[trim={ 0cm 0cm 0cm 0cm},scale=0.2]{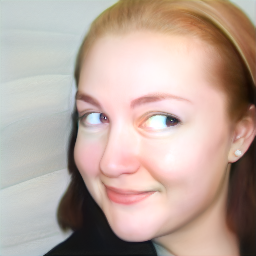} &
\includegraphics[trim={ 0cm 0cm 0cm 0cm},scale=0.2]{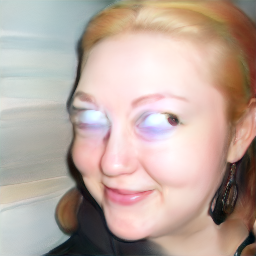} &
\includegraphics[trim={ 0cm 0cm 0cm 0cm},scale=0.2]{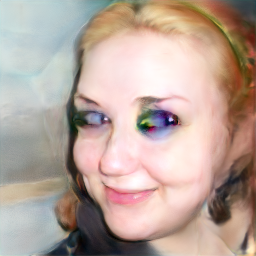}
\vspace{2mm}
\\

\rotatebox{90}{\phantom{kk}Lipstick}
\includegraphics[trim={ 0cm 0cm 0cm 0cm},scale=0.2]{./invert/0_original.png} &
\includegraphics[trim={ 0cm 0cm 0cm 0cm},scale=0.2]{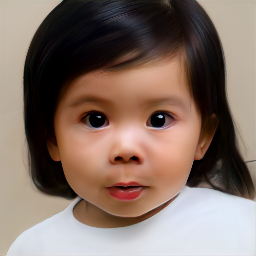} &
\includegraphics[trim={ 0cm 0cm 0cm 0cm},scale=0.2]{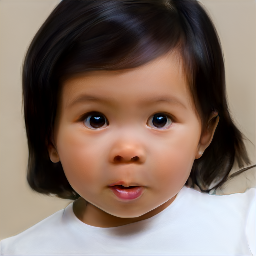} &
\includegraphics[trim={ 0cm 0cm 0cm 0cm},scale=0.2]{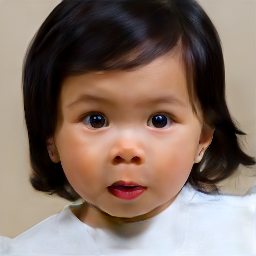} 
\\
\rotatebox{90}{\phantom{kk}Lipstick}
\includegraphics[trim={ 0cm 0cm 0cm 0cm},scale=0.2]{./invert/2_original.png} &
\includegraphics[trim={ 0cm 0cm 0cm 0cm},scale=0.2]{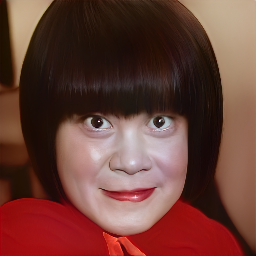} &
\includegraphics[trim={ 0cm 0cm 0cm 0cm},scale=0.2]{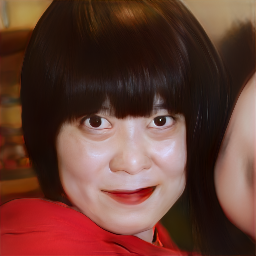} &
\includegraphics[trim={ 0cm 0cm 0cm 0cm},scale=0.2]{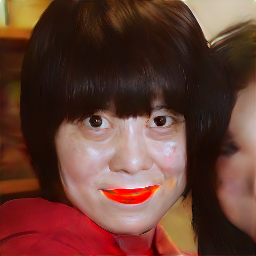} 
\\
\rotatebox{90}{\phantom{kk}Lipstick}
\includegraphics[trim={ 0cm 0cm 0cm 0cm},scale=0.2]{./invert/3_original.png} &
\includegraphics[trim={ 0cm 0cm 0cm 0cm},scale=0.2]{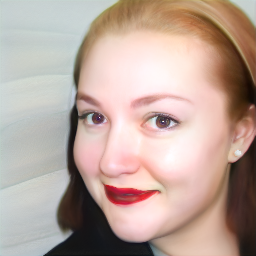} &
\includegraphics[trim={ 0cm 0cm 0cm 0cm},scale=0.2]{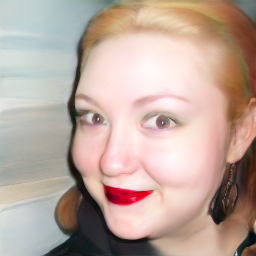} &
\includegraphics[trim={ 0cm 0cm 0cm 0cm},scale=0.2]{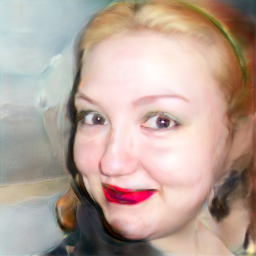}
\vspace{2mm}
\\

\end{tabular}

   \caption{Manipulation of style codes obtained by latent optimization in $\mathcal{W}$, $\mathcal{W+}$, and $\mathcal{S}$ spaces. The exact same manipulation is applied in each row.
   It may be seen that manipulation of codes optimized in $\mathcal{S}$ produces significant artifacts, while manipulation on codes optimized in $\mathcal{W}$ produce more realistic results.}
\label{fig:invert_m}
\end{figure}

\begin{figure}[h]
	\setlength{\tabcolsep}{1.8pt}
	
	{\footnotesize
		\begin{tabular}{ccccccccc}
			Original & Inverted & Smile & Lipstick &  Gaze & Eye Shape & Frown Eyebrows & Goatee & Bulbous Nose \\
			\includegraphics[trim={ 0cm 0cm 0cm 0cm},scale=0.2]{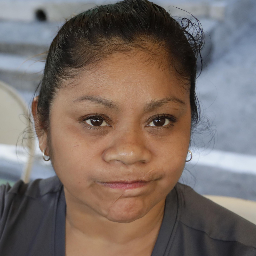} &
			\includegraphics[trim={ 0cm 0cm 0cm 0cm},scale=0.2]{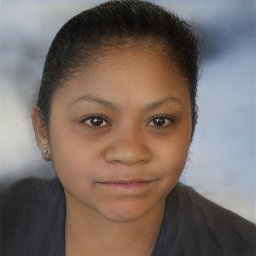} &
			\includegraphics[trim={ 0cm 0cm 0cm 0cm},scale=0.2]{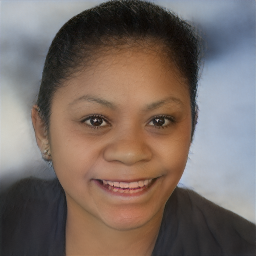} &
			\includegraphics[trim={ 0cm 0cm 0cm 0cm},scale=0.2]{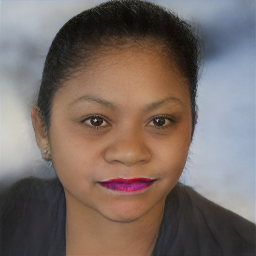} &
			\includegraphics[trim={ 0cm 0cm 0cm 0cm},scale=0.2]{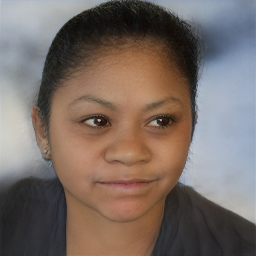} &
			\includegraphics[trim={ 0cm 0cm 0cm 0cm},scale=0.2]{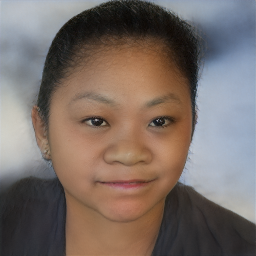} &
			\includegraphics[trim={ 0cm 0cm 0cm 0cm},scale=0.2]{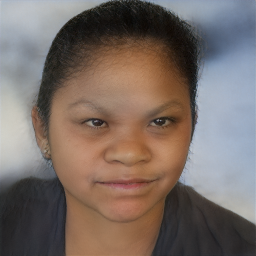}&
			\includegraphics[trim={ 0cm 0cm 0cm 0cm},scale=0.2]{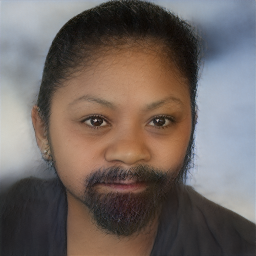}&
			\includegraphics[trim={ 0cm 0cm 0cm 0cm},scale=0.2]{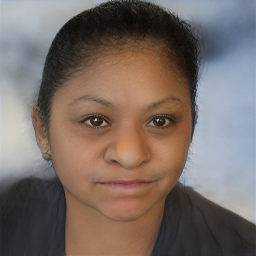}
			\\
			\includegraphics[trim={ 0cm 0cm 0cm 0cm},scale=0.2]{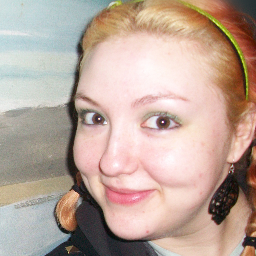} &
			\includegraphics[trim={ 0cm 0cm 0cm 0cm},scale=0.2]{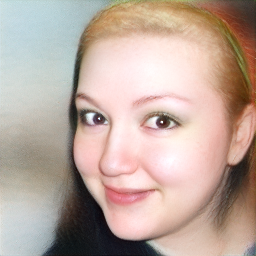} &
			\includegraphics[trim={ 0cm 0cm 0cm 0cm},scale=0.2]{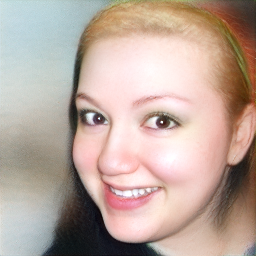} &
			\includegraphics[trim={ 0cm 0cm 0cm 0cm},scale=0.2]{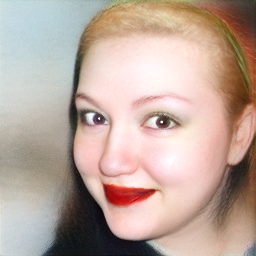} &
			\includegraphics[trim={ 0cm 0cm 0cm 0cm},scale=0.2]{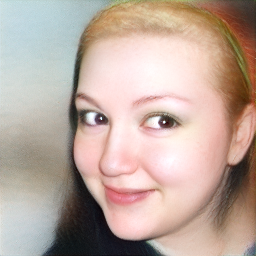} &
			\includegraphics[trim={ 0cm 0cm 0cm 0cm},scale=0.2]{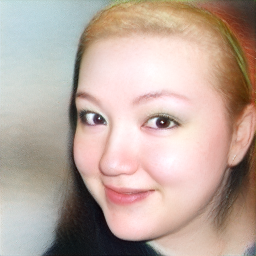} &
			\includegraphics[trim={ 0cm 0cm 0cm 0cm},scale=0.2]{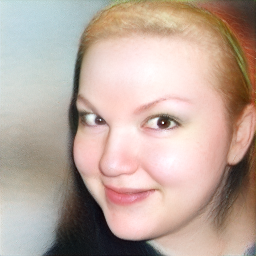}&
			\includegraphics[trim={ 0cm 0cm 0cm 0cm},scale=0.2]{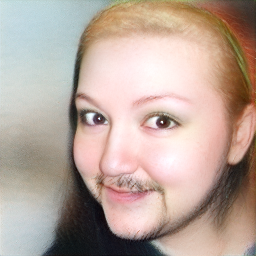}&
			\includegraphics[trim={ 0cm 0cm 0cm 0cm},scale=0.2]{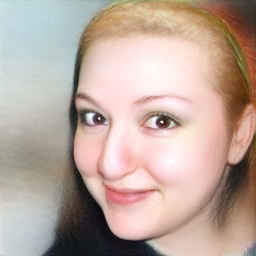}
			\\
			\includegraphics[trim={ 0cm 0cm 0cm 0cm},scale=0.2]{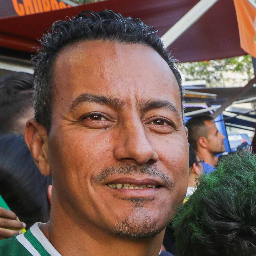} &
			\includegraphics[trim={ 0cm 0cm 0cm 0cm},scale=0.2]{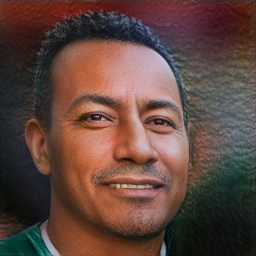} &
			\includegraphics[trim={ 0cm 0cm 0cm 0cm},scale=0.2]{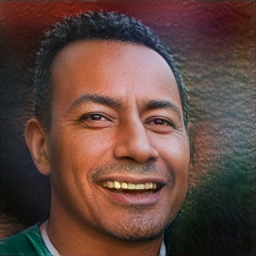} &
			\includegraphics[trim={ 0cm 0cm 0cm 0cm},scale=0.2]{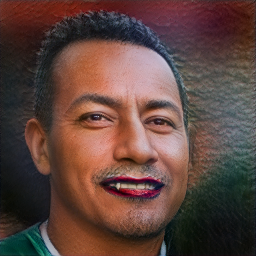} &
			\includegraphics[trim={ 0cm 0cm 0cm 0cm},scale=0.2]{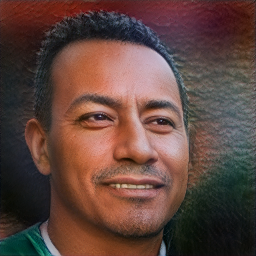} &
			\includegraphics[trim={ 0cm 0cm 0cm 0cm},scale=0.2]{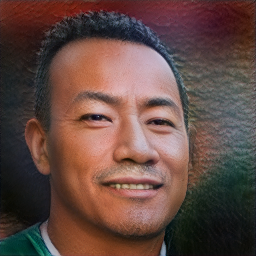} &
			\includegraphics[trim={ 0cm 0cm 0cm 0cm},scale=0.2]{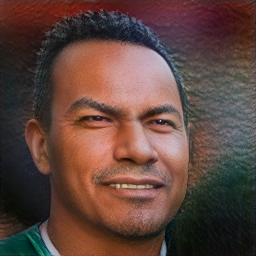}&
			\includegraphics[trim={ 0cm 0cm 0cm 0cm},scale=0.2]{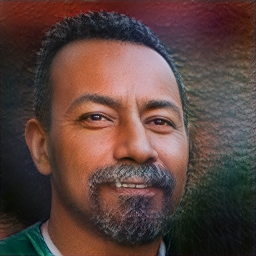}&
			\includegraphics[trim={ 0cm 0cm 0cm 0cm},scale=0.2]{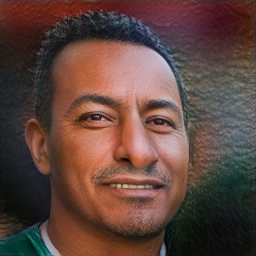}
			\\
			\includegraphics[trim={ 0cm 0cm 0cm 0cm},scale=0.2]{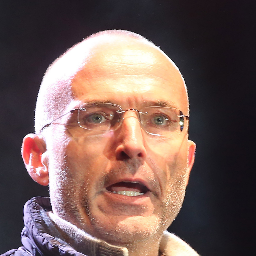} &
			\includegraphics[trim={ 0cm 0cm 0cm 0cm},scale=0.2]{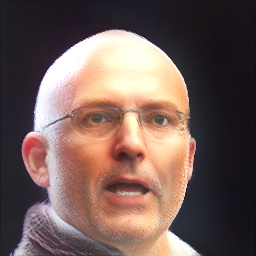} &
			\includegraphics[trim={ 0cm 0cm 0cm 0cm},scale=0.2]{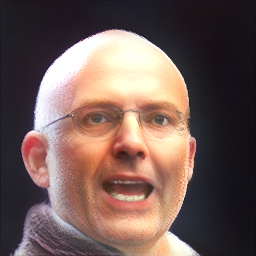} &
			\includegraphics[trim={ 0cm 0cm 0cm 0cm},scale=0.2]{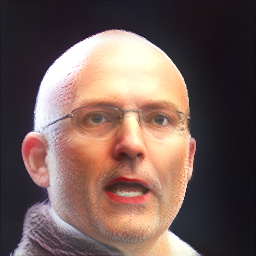} &
			\includegraphics[trim={ 0cm 0cm 0cm 0cm},scale=0.2]{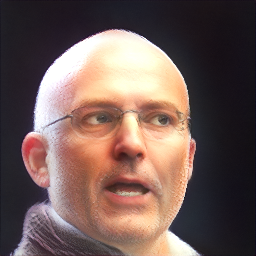} &
			\includegraphics[trim={ 0cm 0cm 0cm 0cm},scale=0.2]{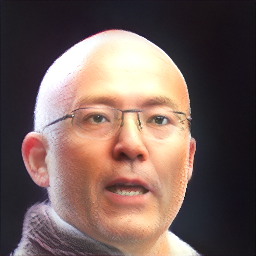} &
			\includegraphics[trim={ 0cm 0cm 0cm 0cm},scale=0.2]{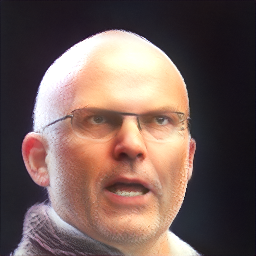}&
			\includegraphics[trim={ 0cm 0cm 0cm 0cm},scale=0.2]{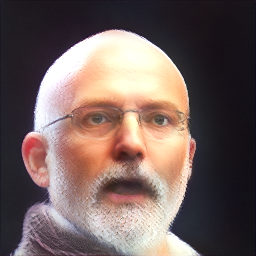}&
			\includegraphics[trim={ 0cm 0cm 0cm 0cm},scale=0.2]{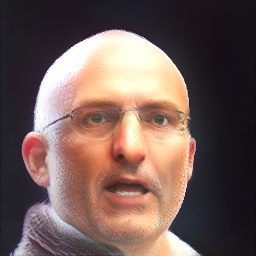}
			\\
			\includegraphics[trim={ 0cm 0cm 0cm 0cm},scale=0.2]{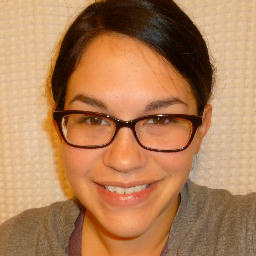} &
			\includegraphics[trim={ 0cm 0cm 0cm 0cm},scale=0.2]{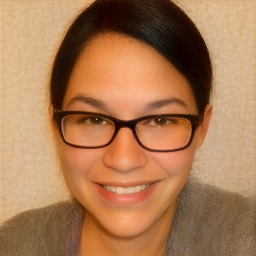} &
			\includegraphics[trim={ 0cm 0cm 0cm 0cm},scale=0.2]{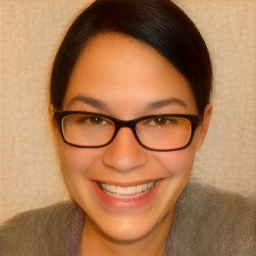} &
			\includegraphics[trim={ 0cm 0cm 0cm 0cm},scale=0.2]{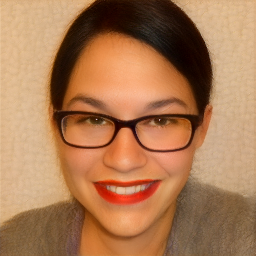} &
			\includegraphics[trim={ 0cm 0cm 0cm 0cm},scale=0.2]{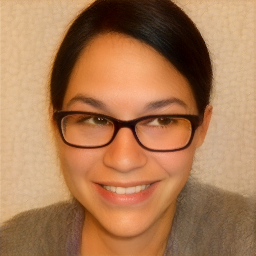} &
			\includegraphics[trim={ 0cm 0cm 0cm 0cm},scale=0.2]{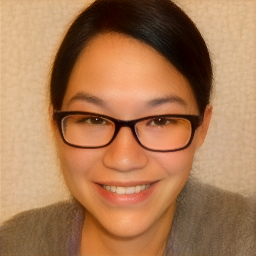} &
			\includegraphics[trim={ 0cm 0cm 0cm 0cm},scale=0.2]{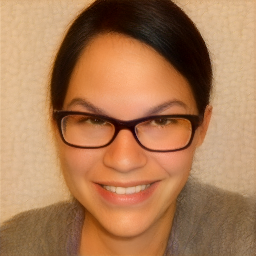}&
			\includegraphics[trim={ 0cm 0cm 0cm 0cm},scale=0.2]{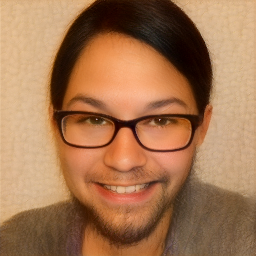}&
			\includegraphics[trim={ 0cm 0cm 0cm 0cm},scale=0.2]{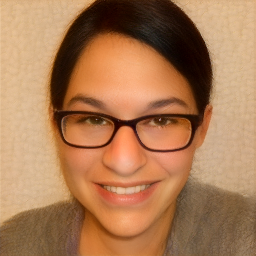}
			\\
			\includegraphics[trim={ 0cm 0cm 0cm 0cm},scale=0.2]{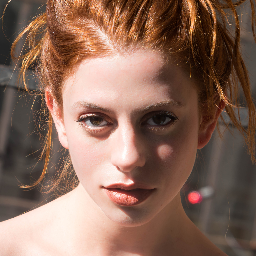} &
			\includegraphics[trim={ 0cm 0cm 0cm 0cm},scale=0.2]{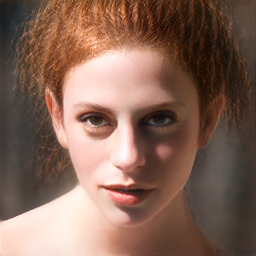} &
			\includegraphics[trim={ 0cm 0cm 0cm 0cm},scale=0.2]{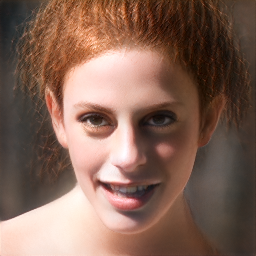} &
			\includegraphics[trim={ 0cm 0cm 0cm 0cm},scale=0.2]{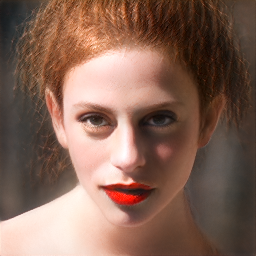} &
			\includegraphics[trim={ 0cm 0cm 0cm 0cm},scale=0.2]{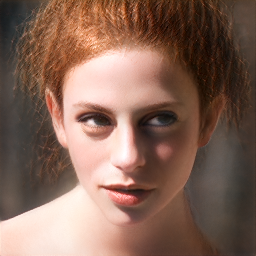} &
			\includegraphics[trim={ 0cm 0cm 0cm 0cm},scale=0.2]{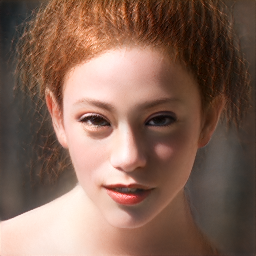} &
			\includegraphics[trim={ 0cm 0cm 0cm 0cm},scale=0.2]{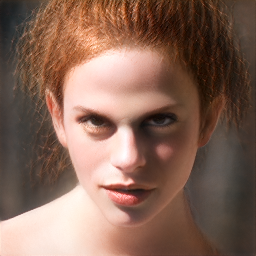}&
			\includegraphics[trim={ 0cm 0cm 0cm 0cm},scale=0.2]{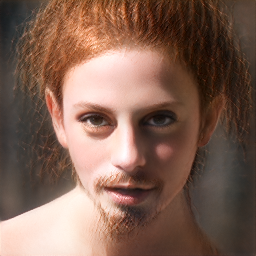}&
			\includegraphics[trim={ 0cm 0cm 0cm 0cm},scale=0.2]{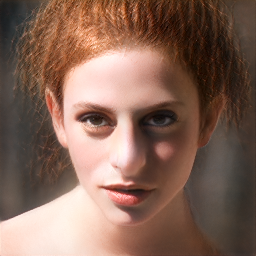}
			\\
			\includegraphics[trim={ 0cm 0cm 0cm 0cm},scale=0.2]{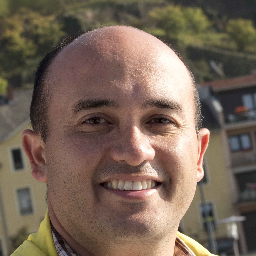} &
			\includegraphics[trim={ 0cm 0cm 0cm 0cm},scale=0.2]{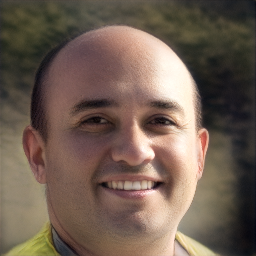} &
			\includegraphics[trim={ 0cm 0cm 0cm 0cm},scale=0.2]{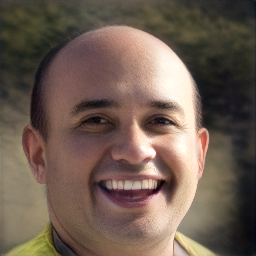} &
			\includegraphics[trim={ 0cm 0cm 0cm 0cm},scale=0.2]{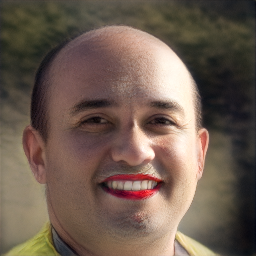} &
			\includegraphics[trim={ 0cm 0cm 0cm 0cm},scale=0.2]{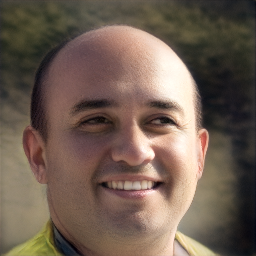} &
			\includegraphics[trim={ 0cm 0cm 0cm 0cm},scale=0.2]{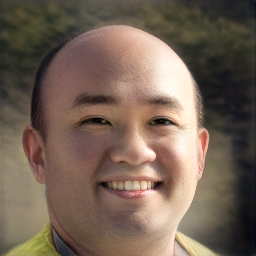} &
			\includegraphics[trim={ 0cm 0cm 0cm 0cm},scale=0.2]{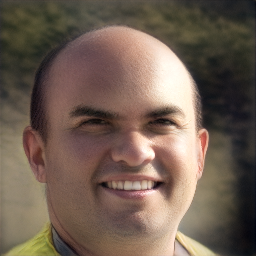}&
			\includegraphics[trim={ 0cm 0cm 0cm 0cm},scale=0.2]{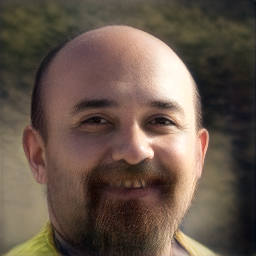}&
			\includegraphics[trim={ 0cm 0cm 0cm 0cm},scale=0.2]{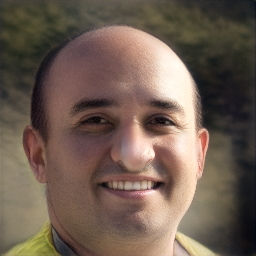}
			\\
		\end{tabular}
	}
	\caption{Manipulation of real images using encoder-based inversion.
		Original images are from FFHQ, and were not part of the encoder's training set.}
	\label{fig:real2}
\end{figure}

\begin{figure}[h]
	\setlength{\tabcolsep}{1.8pt}
	
	{\footnotesize
		\begin{tabular}{ccccccccc}
			Original & Inverted & Smile & Lipstick &  Gaze & Eye Shape & Frown Eyebrows & Goatee & Bulbous Nose \\
			\includegraphics[trim={ 0cm 0cm 0cm 0cm},scale=0.2]{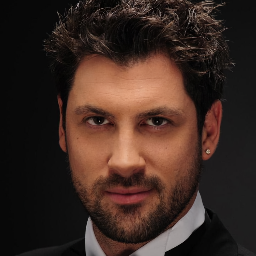} &
			\includegraphics[trim={ 0cm 0cm 0cm 0cm},scale=0.2]{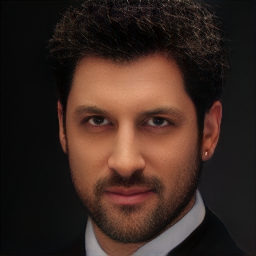} &
			\includegraphics[trim={ 0cm 0cm 0cm 0cm},scale=0.2]{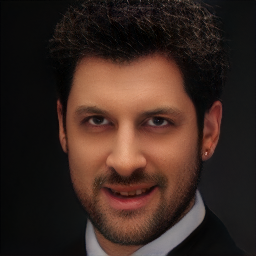} &
			\includegraphics[trim={ 0cm 0cm 0cm 0cm},scale=0.2]{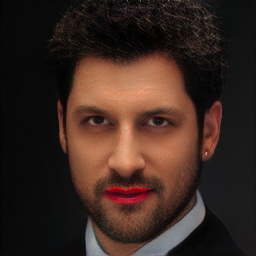} &
			\includegraphics[trim={ 0cm 0cm 0cm 0cm},scale=0.2]{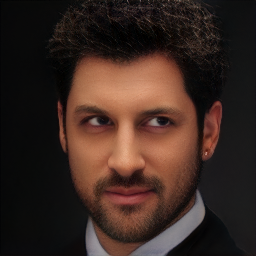} &
			\includegraphics[trim={ 0cm 0cm 0cm 0cm},scale=0.2]{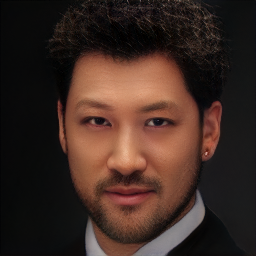} &
			\includegraphics[trim={ 0cm 0cm 0cm 0cm},scale=0.2]{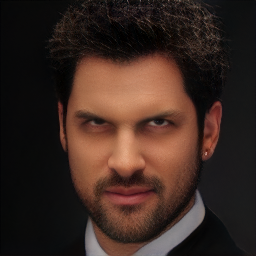}&
			\includegraphics[trim={ 0cm 0cm 0cm 0cm},scale=0.2]{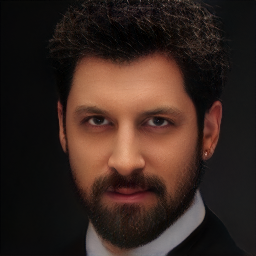}&
			\includegraphics[trim={ 0cm 0cm 0cm 0cm},scale=0.2]{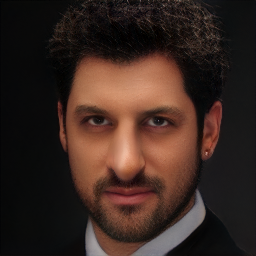}
			\\
			\includegraphics[trim={ 0cm 0cm 0cm 0cm},scale=0.2]{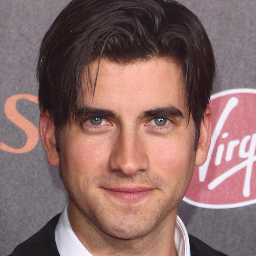} &
			\includegraphics[trim={ 0cm 0cm 0cm 0cm},scale=0.2]{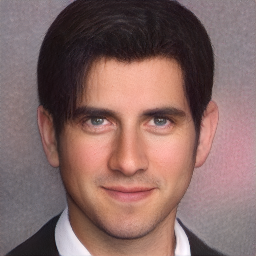} &
			\includegraphics[trim={ 0cm 0cm 0cm 0cm},scale=0.2]{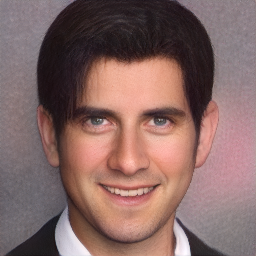} &
			\includegraphics[trim={ 0cm 0cm 0cm 0cm},scale=0.2]{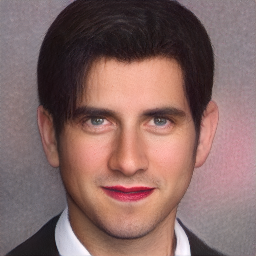} &
			\includegraphics[trim={ 0cm 0cm 0cm 0cm},scale=0.2]{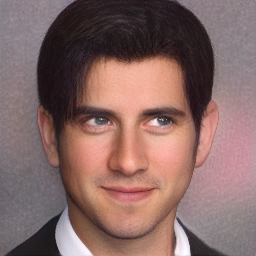} &
			\includegraphics[trim={ 0cm 0cm 0cm 0cm},scale=0.2]{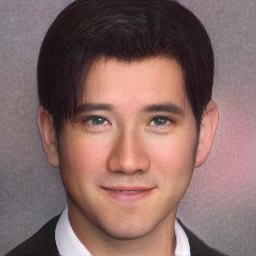} &
			\includegraphics[trim={ 0cm 0cm 0cm 0cm},scale=0.2]{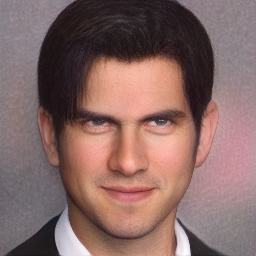}&
			\includegraphics[trim={ 0cm 0cm 0cm 0cm},scale=0.2]{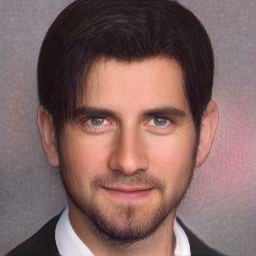}&
			\includegraphics[trim={ 0cm 0cm 0cm 0cm},scale=0.2]{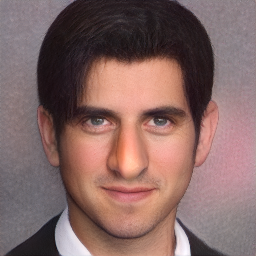}
			\\
			\includegraphics[trim={ 0cm 0cm 0cm 0cm},scale=0.2]{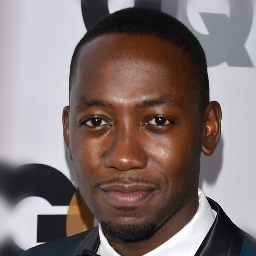} &
			\includegraphics[trim={ 0cm 0cm 0cm 0cm},scale=0.2]{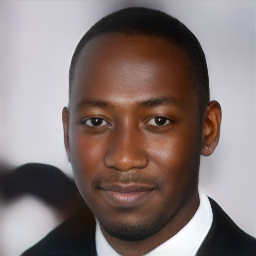} &
			\includegraphics[trim={ 0cm 0cm 0cm 0cm},scale=0.2]{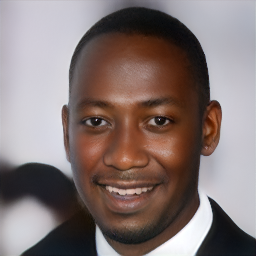} &
			\includegraphics[trim={ 0cm 0cm 0cm 0cm},scale=0.2]{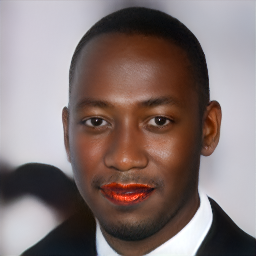} &
			\includegraphics[trim={ 0cm 0cm 0cm 0cm},scale=0.2]{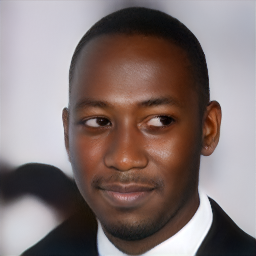} &
			\includegraphics[trim={ 0cm 0cm 0cm 0cm},scale=0.2]{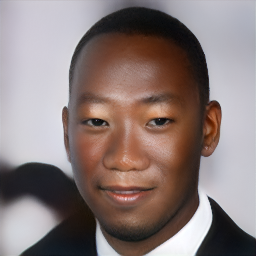} &
			\includegraphics[trim={ 0cm 0cm 0cm 0cm},scale=0.2]{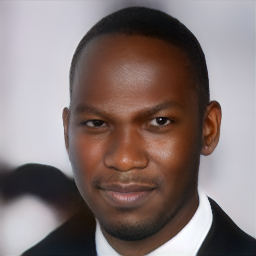}&
			\includegraphics[trim={ 0cm 0cm 0cm 0cm},scale=0.2]{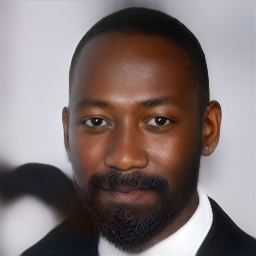}&
			\includegraphics[trim={ 0cm 0cm 0cm 0cm},scale=0.2]{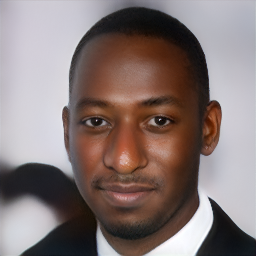}
			\\
			\includegraphics[trim={ 0cm 0cm 0cm 0cm},scale=0.2]{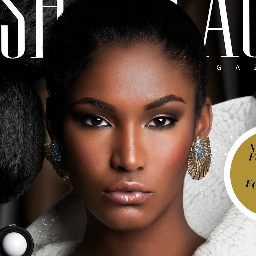} &
			\includegraphics[trim={ 0cm 0cm 0cm 0cm},scale=0.2]{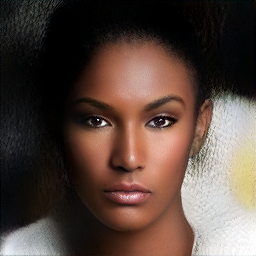} &
			\includegraphics[trim={ 0cm 0cm 0cm 0cm},scale=0.2]{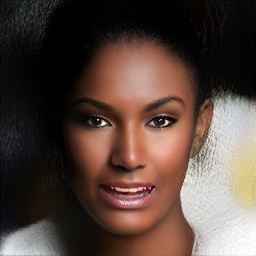} &
			\includegraphics[trim={ 0cm 0cm 0cm 0cm},scale=0.2]{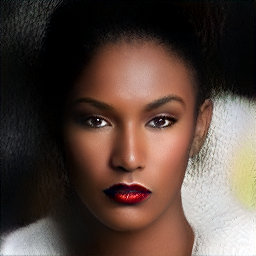} &
			\includegraphics[trim={ 0cm 0cm 0cm 0cm},scale=0.2]{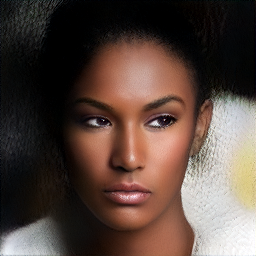} &
			\includegraphics[trim={ 0cm 0cm 0cm 0cm},scale=0.2]{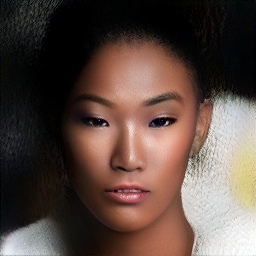} &
			\includegraphics[trim={ 0cm 0cm 0cm 0cm},scale=0.2]{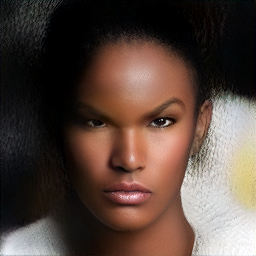}&
			\includegraphics[trim={ 0cm 0cm 0cm 0cm},scale=0.2]{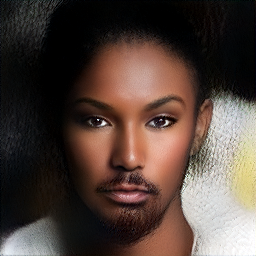}&
			\includegraphics[trim={ 0cm 0cm 0cm 0cm},scale=0.2]{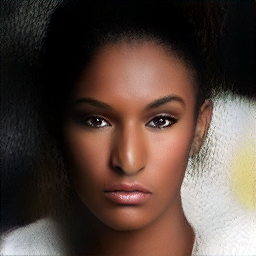}
			\\
			\includegraphics[trim={ 0cm 0cm 0cm 0cm},scale=0.2]{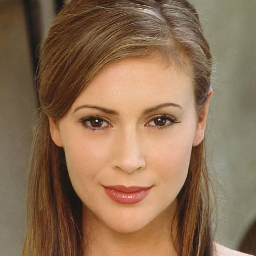} &
			\includegraphics[trim={ 0cm 0cm 0cm 0cm},scale=0.2]{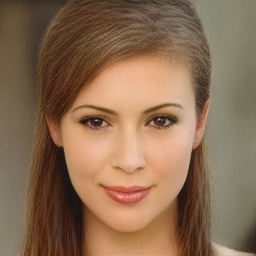} &
			\includegraphics[trim={ 0cm 0cm 0cm 0cm},scale=0.2]{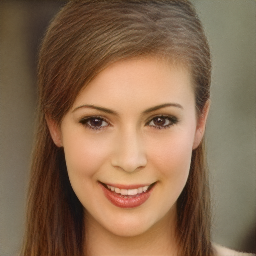} &
			\includegraphics[trim={ 0cm 0cm 0cm 0cm},scale=0.2]{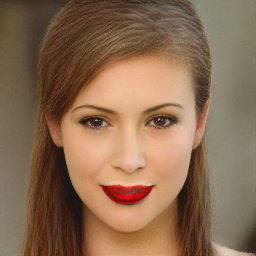} &
			\includegraphics[trim={ 0cm 0cm 0cm 0cm},scale=0.2]{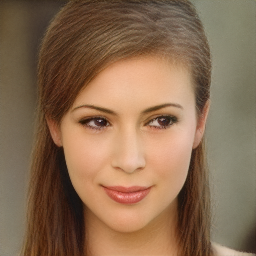} &
			\includegraphics[trim={ 0cm 0cm 0cm 0cm},scale=0.2]{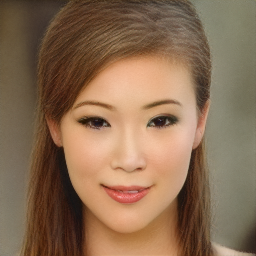} &
			\includegraphics[trim={ 0cm 0cm 0cm 0cm},scale=0.2]{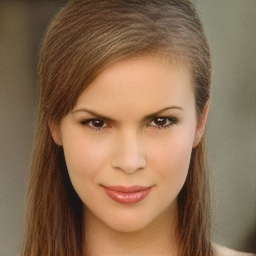}&
			\includegraphics[trim={ 0cm 0cm 0cm 0cm},scale=0.2]{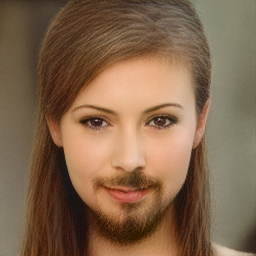}&
			\includegraphics[trim={ 0cm 0cm 0cm 0cm},scale=0.2]{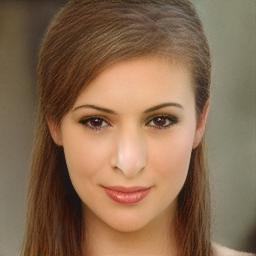}
			\\
			\includegraphics[trim={ 0cm 0cm 0cm 0cm},scale=0.2]{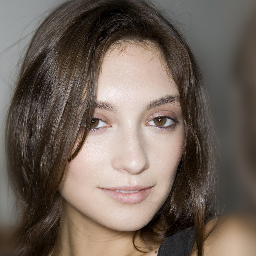} &
			\includegraphics[trim={ 0cm 0cm 0cm 0cm},scale=0.2]{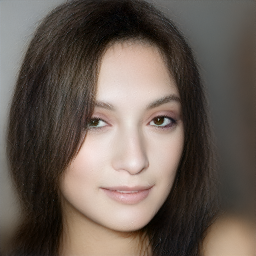} &
			\includegraphics[trim={ 0cm 0cm 0cm 0cm},scale=0.2]{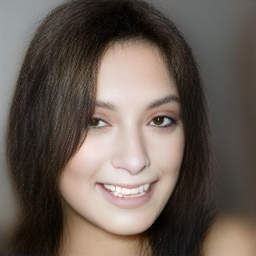} &
			\includegraphics[trim={ 0cm 0cm 0cm 0cm},scale=0.2]{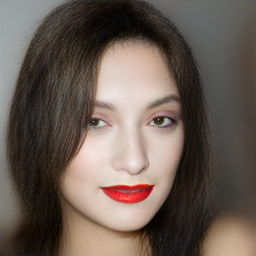} &
			\includegraphics[trim={ 0cm 0cm 0cm 0cm},scale=0.2]{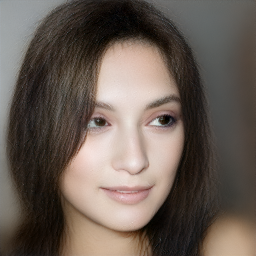} &
			\includegraphics[trim={ 0cm 0cm 0cm 0cm},scale=0.2]{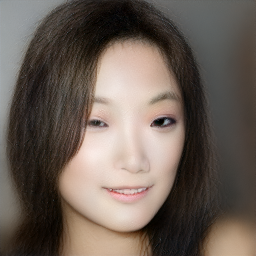} &
			\includegraphics[trim={ 0cm 0cm 0cm 0cm},scale=0.2]{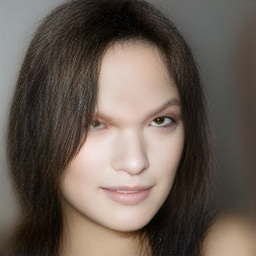}&
			\includegraphics[trim={ 0cm 0cm 0cm 0cm},scale=0.2]{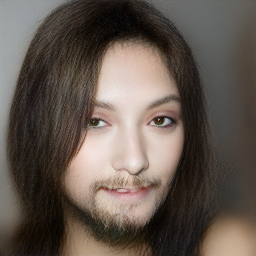}&
			\includegraphics[trim={ 0cm 0cm 0cm 0cm},scale=0.2]{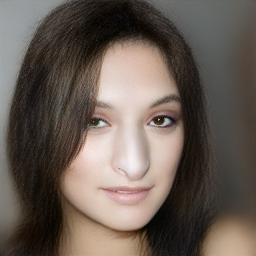}
			\\
			\includegraphics[trim={ 0cm 0cm 0cm 0cm},scale=0.2]{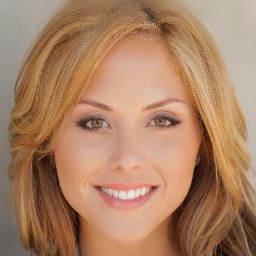} &
			\includegraphics[trim={ 0cm 0cm 0cm 0cm},scale=0.2]{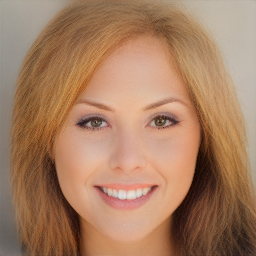} &
			\includegraphics[trim={ 0cm 0cm 0cm 0cm},scale=0.2]{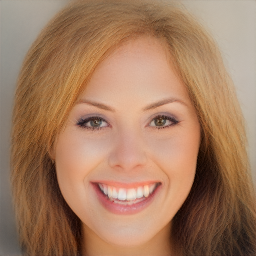} &
			\includegraphics[trim={ 0cm 0cm 0cm 0cm},scale=0.2]{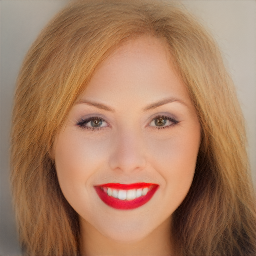} &
			\includegraphics[trim={ 0cm 0cm 0cm 0cm},scale=0.2]{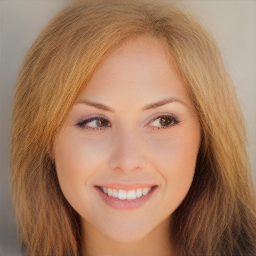} &
			\includegraphics[trim={ 0cm 0cm 0cm 0cm},scale=0.2]{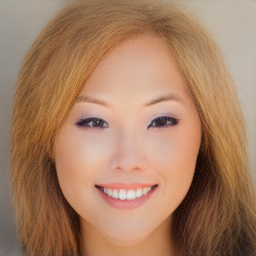} &
			\includegraphics[trim={ 0cm 0cm 0cm 0cm},scale=0.2]{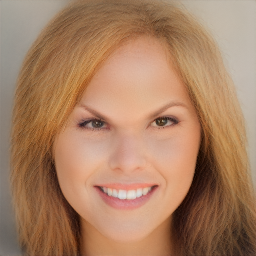}&
			\includegraphics[trim={ 0cm 0cm 0cm 0cm},scale=0.2]{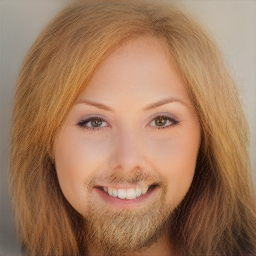}&
			\includegraphics[trim={ 0cm 0cm 0cm 0cm},scale=0.2]{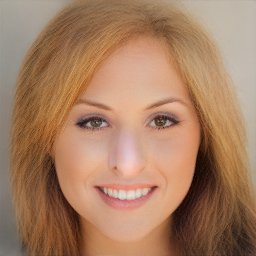}
			\\
			\includegraphics[trim={ 0cm 0cm 0cm 0cm},scale=0.2]{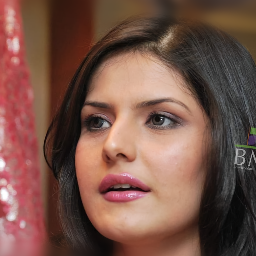} &
			\includegraphics[trim={ 0cm 0cm 0cm 0cm},scale=0.2]{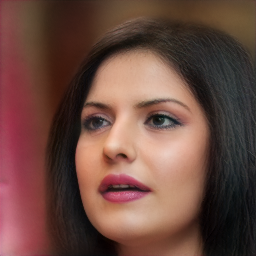} &
			\includegraphics[trim={ 0cm 0cm 0cm 0cm},scale=0.2]{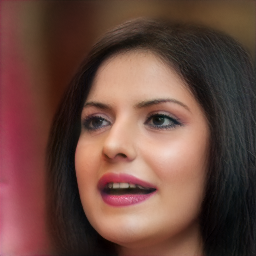} &
			\includegraphics[trim={ 0cm 0cm 0cm 0cm},scale=0.2]{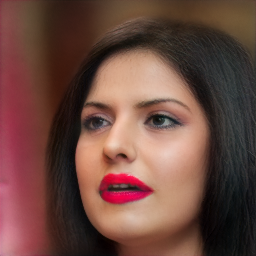} &
			\includegraphics[trim={ 0cm 0cm 0cm 0cm},scale=0.2]{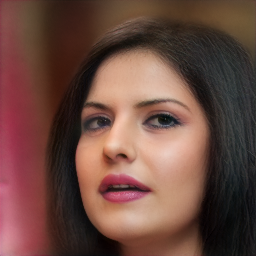} &
			\includegraphics[trim={ 0cm 0cm 0cm 0cm},scale=0.2]{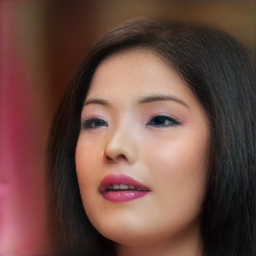} &
			\includegraphics[trim={ 0cm 0cm 0cm 0cm},scale=0.2]{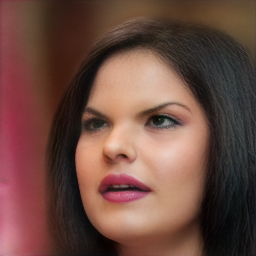}&
			\includegraphics[trim={ 0cm 0cm 0cm 0cm},scale=0.2]{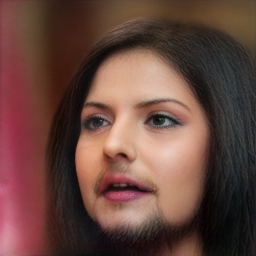}&
			\includegraphics[trim={ 0cm 0cm 0cm 0cm},scale=0.2]{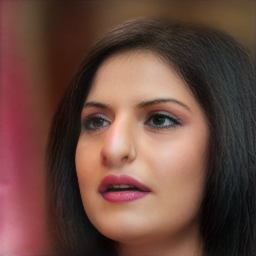}
		\end{tabular}
	}
	\caption{Manipulation of real images using encoder-based inversion.
		Original images are from CelebA-HQ, which were not part of the encoder's training set, and not part of the GAN training set (the StyleGAN2 model was trained on the FFHQ dataset).}
	\label{fig:real3}
\end{figure}

To manipulate real images, it is necessary to first invert them into latent codes. Through latent optimization \cite{karras2020analyzing}, we observe that the reconstruction quality is the highest when optimizing in $\mathcal{S}$, followed by $\mathcal{W+}$, and is the lowest for $\mathcal{W}$, as demonstrated in Figure~\ref{fig:invert}. However, the naturalness of subsequent manipulation is the best when the latent optimization is done in $\mathcal{W}$, followed by $\mathcal{W+}$, and the worst for $\mathcal{S}$, as shown in  Figure~\ref{fig:invert_m}. Through training a latent embedding encoder and using the embedding produced by the encoder as the initial point for a few iterations of latent optimization, we obtain both good reconstruction and natural manipulation. We demonstrate manipulation of real images in Figure~\ref{fig:real2} (for real images from the FFHQ dataset) and in Figure~\ref{fig:real3} for images from the CelebA-HQ \cite{liu2015deep} dataset.

\end{document}